%% file: IEEETSP-main-paper.tex
\documentclass[11pt,twoside]{article}

\usepackage{geometry}
\input{command-template}

\begin{document}

\begin{center}

  {\bf{\LARGE{Do algorithms and barriers for sparse principal \\
  \mbox{component analysis extend to other structured settings?}}}}
  
\vspace*{.2in}

{\large{
\begin{tabular}{ccc}
Guanyi Wang$^{\ddagger}$, Mengqi Lou$^{\star}$, Ashwin Pananjady$^{\star, \dagger}$
\end{tabular}
}}

\vspace*{.2in}

\begin{tabular}{c}
Department of Industrial Systems Engineering and Management$^\ddagger$, National University of Singapore \\
Schools of Industrial and Systems Engineering$^\star$ and Electrical and Computer Engineering$^\dagger$ \\
Georgia Institute of Technology
\end{tabular}

\vspace*{.2in}
\today
\vspace*{.2in}

\begin{abstract}
We study a principal component analysis problem under the spiked Wishart model in which the structure in the signal is captured by a class of union-of-subspace models. This general class includes vanilla sparse PCA as well as its variants with graph sparsity. With the goal of studying these problems under a unified statistical and computational lens, we establish fundamental limits that depend on the geometry of the problem instance, and show that a natural projected power method exhibits local convergence to the statistically near-optimal neighborhood of the solution. We complement these results with end-to-end analyses of two important special cases given by path and tree sparsity in a general basis, showing initialization methods and matching evidence of computational hardness. Overall, our results indicate that several of the phenomena observed for vanilla sparse PCA extend in a natural fashion to its structured counterparts.
\end{abstract}
\end{center}



\input{intro}

\input{problem-setting}

\input{stat-optest}

\input{PPM-EP}

\input{example-path-SPCA}
\input{example-tree-SPCA}

\input{discussion}


\subsubsection*{Acknowledgements}
GW was supported by the National University of Singapore under AcRF Tier-1 grant (A-8000607-00-00) 22-5539-A0001. ML and AP were supported in part by the NSF under grants CCF-2107455 and DMS-2210734 and by research awards/gifts from Adobe, Amazon, and Mathworks. We are grateful to the Simons Institute for the Theory of Computing for their hospitality, where part of this work was performed.

\bibliographystyle{abbrvnat}
\bibliography{reference}

 

\appendix  
\input{appendix}

\end{document}

%% file: command-template.tex
\usepackage{comment}
\usepackage{epsf}
\usepackage{graphicx}
\usepackage{subfigure}
\usepackage{bbm}
\usepackage{color}
\usepackage{nicefrac}
\usepackage{mathtools}
\usepackage{amsfonts}
\usepackage{amsmath, bm}
\usepackage{amssymb}
\usepackage{textcomp}
\usepackage{xcolor}

\usepackage{algorithm}
\usepackage{algpseudocode} 

\newenvironment{fminipage}%
  {\begin{Sbox}\begin{minipage}}%
  {\end{minipage}\end{Sbox}\fbox{\TheSbox}}

\newcommand{\BlackBox}{\rule{1.5ex}{1.5ex}}  
\ifdefined\proof
    \renewenvironment{proof}{\par\noindent{\bf Proof\ }}{\hfill\BlackBox\\[2mm]}
\else
    \newenvironment{proof}{\par\noindent{\bf Proof\ }}{\hfill\BlackBox\\[2mm]}
\fi
\newtheorem{theorem}{Theorem}
\newtheorem{example}{Example}
\newtheorem{assumption}{Assumption}
\newtheorem{proposition}{Proposition}
\newtheorem{lemma}{Lemma}
\newtheorem{corollary}{Corollary}
\newtheorem{definition}{Definition}
\newtheorem{remark}{Remark}
\newtheorem{conjecture}{Conjecture}

\usepackage{natbib}
\setcitestyle{round}

\newlength{\widebarargwidth}
\newlength{\widebarargheight}
\newlength{\widebarargdepth}

\makeatletter
\long\def\@makecaption#1#2{
       \vskip 0.8ex
       \setbox\@tempboxa\hbox{\small {\bf #1:} #2}
       \parindent 1.5em 
       \dimen0=\hsize
       \advance\dimen0 by -3em
       \ifdim \wd\@tempboxa >\dimen0
               \hbox to \hsize{
                       \parindent 0em
                       \hfil 
                       \parbox{\dimen0}{\def\baselinestretch{0.96}\small
                               {\bf #1.} #2
                               } 
                       \hfil}
       \else \hbox to \hsize{\hfil \box\@tempboxa \hfil}
       \fi
       }
\makeatother

\long\def\comment#1{}

\newcommand{\blue}[1]{\textcolor{blue}{#1}}
\newcommand{\apcomment}[1]{{\bf{{\blue{{AP --- #1}}}}}}

\newcommand{\red}[1]{\textcolor{red}{#1}}

\newcommand{\gwcomment}[1]{{\bf{{\red{{GW --- #1}}}}}}

\DeclareMathOperator*{\argmin}{argmin}
\DeclareMathOperator*{\argmax}{argmax}

\newcommand{\1}{\ensuremath{{\sf (i)}}}
\newcommand{\2}{\ensuremath{{\sf (ii)}}}

\newcommand{\ESest}{\widehat{\bm{v}}_{\mathsf{ES}}}
\newcommand{\EST}{{\tt Est}}
\newcommand{\DET}{{\tt Det}}
\newcommand{\subindex}{j}

%% file: intro.tex


\section{Introduction} \label{sec:intro}

Principal component analysis (PCA) is a preponderant tool for dimensionality reduction and feature extraction. PCA and its generalizations have been used for numerous applications including wavelet decomposition~\citep{mallat1999wavelet, baraniuk2010model}, representative stock selection from
business sectors~\citep{asteris2015stay}, human face recognition \citep{hancock1996face, tran2020tensor}, eigen-gene selection and shaving \citep{alter2000singular,hastie2000gene,feng2019supervised}, handwriting classification \citep{hastie2009elements}, clustering of functional connectivity \citep{frusque2019sparse}, and single-cell RNA sequencing analysis \citep{wang2021manifold}, to name but a few. 

Given a set of $n$ samples $\bm{x}_1, \ldots, \bm{x}_n \in \mathbb{R}^d$, PCA is traditionally phrased as the problem of recovering the direction of maximal variance.
However, in high dimensions when $d \gg n$, it is well-known that the vanilla estimator given by the maximal eigenvector of the sample covariance matrix of the data is inconsistent (see, e.g.,~\citet{johnstone2009consistency} and references therein).
This inconsistency motivates imposing sparsity assumptions on the "ground-truth" principal component and studying the resulting problem under a generative model for the data. Indeed, sparsity has emerged as a key structural assumption inspired by the diverse applications mentioned earlier, and a wealth of literature now exists on the sparse PCA problem. In practice, additional structure exists on the ground truth principle component. For instance, in applications involving wavelet decompositions, the signal is well-modeled by structured sparsity defined on a binary tree \citep{baraniuk2010model}. Similarly, path sparsity on the principal component is a reasonable assumption when dealing with data representing stocks across distinct business sectors \citep{asteris2015stay}. 

In this paper, we study a class of union-of-linearly-structured models (see Section~\ref{sec:setting-background}), which includes vanilla sparse PCA and path/tree sparse PCA as special cases. Our goal is to understand, through a statistical and computational lens, if and to what extent the theoretical results and insights developed for vanilla sparsity extend to these structured settings. In particular, given that the vanilla sparse PCA has a delicate statistical-computational gap, a conceptual question that motivates our research is 
	\begin{center}
		\textit{Does such a statistical-computational gap persist when additional structure is imposed in PCA?}
	\end{center}
	Generally speaking, statistical-computational gaps in related problems are delicate\footnote{For one such example, note that gaps disappear in the sparse stochastic block model in the presence of a "monotone adversary"~\citep{moitra2016robust}.}, and so understanding the influence of additional structure in such problems is an important goal. 
	In making progress toward this goal, we carry out a detailed statistical and computational study of a broad family of structured PCA problems.

\subsection{Contributions and organization}
In Section~\ref{sec:setting-background}, we formally introduce a family of 
\emph{union of linearly structured} PCA problems under the spiked Wishart model. Section~\ref{sec:general-results} presents our main results:
We begin by studying the fundamental limits of estimation under this model, providing both upper and lower bounds on the $\ell_2$ error of estimation that depend on the geometry of the problem.
Our upper bound is achieved by an exhaustive search algorithm, and we analyze a natural projected power method to approximately compute its solution. We show that this iterative method enjoys local geometric convergence to within a neighborhood of the ground truth solution that attains the optimal statistical rate as a function of the sample size and geometry of the problem instance. We also present a general initialization algorithm for this method.
%
%
In Section~\ref{sec:specific-examples}, we study two prototypical examples of structured PCA---those given by path and tree sparsity---in an end-to-end fashion, additionally providing explicit initialization methods and evidence of computational hardness (see in particular Propositions~\ref{prop:reduction-PathPCA} and~\ref{prop:TS-SDP-hard}). Detailed proofs of our results can be found in the supplementary material.

Through our statistical, algorithmic, and reduction-based results, we find that several features of vanilla sparse PCA---on both the statistical and computational fronts---persist and extend in natural ways to its structured counterparts. In particular, while the imposition of structure can help mildly, it does not seem to make the problem significantly easier to solve in a computationally efficient manner.

\subsection{Related work} \label{sec:related-works}
Structured PCA has been studied extensively over the past two decades, and we cannot hope to cover this vast literature here. We discuss the papers most relevant to our results.

\paragraph{Optimization algorithms for sparse PCA} The most commonly used and studied structural assumption in PCA is (vanilla) sparsity, in which the true principal component is assumed to be $k$-sparse. Letting $\widehat{\bm{\Sigma}} := \frac{1}{n} \sum_{i = 1}^n \bm{x}_i \bm{x}_i^{\top}$ denote the sample covariance matrix, such a sparse principal component can be found by solving the following optimization problem: 
\begin{align}
    \max_{\bm{v} \in \mathbb{R}^d} ~ \bm{v}^{\top} \widehat{\bm{\Sigma}} \bm{v} ~~\text{s.t.}~~ \|\bm{v}\|_2 = 1, \|\bm{v}\|_0 \leq k, \label{eq:sparse-PCA}
\end{align} 
where $\| \cdot \|_0$ denotes the $\ell_0$ norm or number of nonzeros. This program was first proposed by \cite{cadima1995loading}. In contrast to classical PCA (which is akin to program~\eqref{eq:sparse-PCA} but without the $\ell_0$ norm constraint), solving the sparse PCA problem \eqref{eq:sparse-PCA} is NP-hard. Many computationally efficient reformulations of sparse PCA have been proposed over the years. \cite{jolliffe2003modified} give the first computational tractable method---termed SCoTLASS---which reformulates the program~\eqref{eq:sparse-PCA} using an $\ell_1$-norm regularization akin to the LASSO \citep{tibshirani1996regression}. \cite{zou2005regularization} and \cite{ zou2006sparse} propose an ElasticNet version of SPCA, and \cite{witten2009penalized} study connections between SCoTLASS and ElasticNet SPCA. \cite{erichson2020sparse, chen2020alternating} propose alternative formulations and show the convergence of their alternating gradient methods to stationary points. Another approach focuses on convex relaxations of sparse PCA. For example, \cite{d2004direct, zhang2012sparse, vu2013fantope, d2014approximation, kim2019convexification} consider a convex relaxation by lifting the variable space $\bm{v} \in \mathbb{R}^d$ to its product space, and relax to a semidefinite programming problem. More recently, \cite{dey2021using, dey2020solving, li2020exact} provide a more computationally scalable type of convex relaxation for problem~\eqref{eq:sparse-PCA} using mixed-integer programming with theoretical worst-case guarantees. Other than methods based on convex relaxation, there is also a substantial literature on specialized iterative algorithms for finding good feasible solutions. Examples include the deflation method~\citep{mackey2008deflation}, generalized power method~\citep{journee2010generalized}, truncated power method~\citep{yuan2013truncated}, and iterative thresholding~\citep{ma2013sparse}. 

\paragraph{Statistical and computational limits of sparse PCA} 


	Several papers have established (by now classical) minimax lower bounds for sparse PCA in a purely statistical sense, i.e., without computational considerations. Examples for vector recovery in $\ell_2$ norm include \cite{birnbaum2013minimax} and \cite{cai2013sparse}; the latter is phrased in terms of estimating the principal subspace and considers a more general model than the rank-1 model. \cite{vu2013minimax} present nonasymptotic lower and upper bounds for the minimax risk considering both row-sparse and column-sparse principal subspaces. \cite{amini2008high} study the rank-1 spiked covariance model considered here, but establish minimax lower bounds for support recovery.

Sparse PCA has also been a key cog in the study of \emph{computational} lower bounds in high dimensional statistics problems, and has received a lot of attention from the perspective of reductions, sum-of-squares and low-degree lower bounds, as well as approaches rooted in statistical physics; let us cover a non-exhaustive list of examples here. 
Assuming the planted clique conjecture,~\cite{berthet2013complexity} show that a sub-Gaussian variant of sparse PCA is hard, in that the optimal rate of estimation is not achievable in polynomial time. 
\citet{ma2015sum} show degree-4 sum of squares lower bounds for $k$-sparse PCA 
(see Section~\ref{sec:TS-PCA}). 
\cite{zdeborova2016statistical} study the fundamental statistical-computational barriers of inference and estimation problems as phase transitions and develop new algorithms using techniques from statistical physics.
\cite{wang2016statistical} show computational lower bounds for estimation for a distributionally-robust variant of sparse PCA. \cite{gao2017sparse} show computational lower bounds for sparse PCA in the spiked covariance model, and \cite{brennan2018reducibility} provide an alternative reduction based on random rotations to strengthen these lower bounds. 
\citet{ding2019subexponential} explore subexponential-time algorithms for sparse PCA, and give rigorous evidence that their proposed algorithm is optimal by analyzing the low-degree likelihood ratio.
\cite{brennan2019optimal} give a reduction from planted clique that yields the first complete characterization of the computational barrier in the spiked covariance model, providing tight lower bounds at all sparsities $k$. 


\paragraph{Structured PCA and related problems} While vanilla sparsity (and the resulting sparse PCA problem) is by far the most well-studied, there also exist other examples of structure one could impose. Examples from the literature on sparse linear regression include graph sparsity~\citep{hegde2015nearly}, group sparsity structure~\citep{yuan2006model}, block and tree sparsity \citep{baraniuk2010model}, and subspace constraints \citep{bie2004learning}. For PCA in particular, several structural constraints have been studied, such as non-negative orthant cone structure \citep{montanari2015non}, and general cone structure \citep{deshpande2014cone, yi2020non}. \citet{asteris2015stay} study path-sparse structure in the PCA problem. 
Some of these papers study fundamental limits of estimation for their specific forms of structure, and \citet{asteris2015stay} and \cite{yi2020non} propose specialized projected power methods.
\cite{cai2021optimal} present a unified framework for the statistical analysis of structured principal subspace estimation and lower and upper bounds on the minimax risk. In recent work,~\cite{liu2021generative} study structured PCA under the assumption that the true principal component is generated from an $L$-Lipschitz continuous generative model, showing that the projected power method enjoys local geometric convergence. While their result is related in spirit to a subset of our results on the projected power method, our structural assumptions are different (see Definition~\ref{cond:linear-structure} and discussions following Theorem~\ref{thm:convergence}) for a detailed comparison. 
There are also papers that study computational hardness in structured settings, both from the perspective of low-degree polynomials~\citep{bandeira2019computational} and reductions from the so-called ``secret-leakage'' variant of the planted clique conjecture~\citep{brennan2020reducibility}. Our work adds to this literature for a particular family of structured PCA problems.


%% file: problem-setting.tex

\section{Problem setting, background, and examples}\label{sec:setting-background}

Throughout this paper, we operate under the spiked Wishart model. Assume that our data set consists of $n$ i.i.d. samples $\{\bm{x}_i\}_{i = 1}^n$ drawn from a $d$-dimensional Gaussian distribution with zero-mean and covariance $\bm{\Sigma} := \lambda \bm{v}_* \bm{v}_*^{\top} + \bm{I}_{d \times d}$. For brevity, we use $\mathcal{D}(\lambda; \bm{v}_*) := \mathcal{N}(\bm{0}_d, \lambda \bm{v}_* \bm{v}_*^{\top} + \bm{I}_{d \times d})$ to denote the distribution of each $\bm{x}_i$. Here $\lambda > 0$ represents the \emph{strength of the signal}, and $\bm{v}_*$ is a $d$-dimensional, unit-norm \emph{ground truth} vector that we wish to estimate. In addition to the unit norm condition, we also assume the inclusion $\bm{v}^* \in \mathcal{M}$, where $\mathcal{M}$ is a known union of subspaces satisfying a certain \emph{union of linear structures} assumption defined below.

\begin{definition} \label{cond:linear-structure}
\textbf{Union of linear structures condition.} Let $\mathcal{B} := \{\phi_1, \ldots, \phi_d\}$ be an orthonormal basis of $\mathbb{R}^d$ and $\mathcal{L} := \left\{ L_1, \ldots, L_M \right\}$ be a collection of $M$ distinct linear subspaces such that for each $m \in [M]$, we have $L_m = \text{span}(\mathcal{B}_m)$ for some $\mathcal{B}_m \subseteq \mathcal{B}$. We say set $\mathcal{M}$ obeys the union of linear structures condition if $\mathcal{M} := \bigcup_{m = 1}^M L_m$, i.e., $\mathcal{M}$ is the union of all linear subspaces in $\mathcal{L}$.
\end{definition}

\begin{remark} \label{rem:eq-structured}
It is worth noting that the union of linear structures condition in Definition~\ref{cond:linear-structure} resembles a structured sparsity condition. Indeed, using the rotation invariance of the Gaussian distribution, the problem of estimating $\bm{v}_*$ from observations $\{\bm{x}_i\}_{i = 1}^n$ is \emph{statistically} equivalent to estimating the structured-sparse vector $\bm{\Phi}^\top \bm{v}_*$ from $\{ \bm{\Phi}^\top \bm{x}_{i} \}_{i = 1}^n$, where $\bm{\Phi} \in \mathbb{R}^{d \times d}$ is an orthonormal matrix with columns $\phi_1, \ldots, \phi_d$. However, the two problems may not be \emph{computationally} equivalent when $\bm{\Phi}$ is unknown. We provide an example in Appendix~\ref{app:time-consuming-case} to illustrate that if an efficient projection oracle onto the union of subspaces $\mathcal{M}$ is accessible, then it is more computationally efficient to estimate the vector $\bm{v}_*$ directly, rather than to estimate $\bm{\Phi}^\top \bm{v}_*$ from $\{ \bm{\Phi}^\top \bm{x}_{i} \}_{i = 1}^n$ by first computing $\bm{\Phi}$. Accordingly, the rest of the paper assumes that $\bm{\Phi}$ is unknown, and that we have access to a projection oracle onto the union of subspaces~$\mathcal{M}$.
\end{remark}

\subsection{Examples of union of linearly structure in Section~\ref{sec:setting-background}} \label{app:examples-definition}

Clearly, vanilla sparse PCA is covered by our formulation. We instantiate the union of linear structures assumption with two other canonical examples.

\subsubsection{Example 1: Tree-Sparse PCA} \label{sec:TS-PCA-intro}

Motivated by applications in signal and image processing and computer graphics \citep{baraniuk2010model}, a particular model for the underlying signal is \emph{tree sparsity} in an underlying basis. In particular, consider the following simplified model for tree-sparsity with one-dimensional signals and binary wavelet trees as a typical such instance. We require some notation to introduce it formally.

Given a natural number $h$, a \textit{complete binary tree} or $\mathsf{CBT}$ of size $d = 2^h - 1$ is given by the following construction. 
Create $h$ levels $\{1, \ldots, h\}$, with $2^{\ell - 1}$ nodes in $\ell$-th level. Index each node from $1$ to $d$, top to bottom and left to right in the following way. The root node $r_{\mathsf{CBT}}$ of $\mathsf{CBT}$ has index $1$, and for any node with index $i \in \{2, \ldots, 2^{h - 1} - 1\}$, its parent is the node with index $\lfloor \frac{i}{2} \rfloor$ and its children are the nodes with indices $2i, 2i + 1$. Define the collection of vertex sets
\begin{align*}
	\mathcal{T}^k := \big\{T: \;  &|T| = k, ~ \text{root node }1 \in T, \text{the subgraph of $\mathsf{CBT}$ induced by $T$ is connected} \big\}. 
\end{align*}

Abusing notation slightly, consider a bijection between the coordinates of any $d$-dimensional vector and the vertices of a $\mathsf{CBT}$.
The vector $\bm{v}_*$ is said to be $k$-\emph{tree-sparse} if $\mathsf{supp}(\bm{v}_*) \in \mathcal{T}^k$.


Therefore, tree-sparse PCA is a specific example of union of linear structures in our formulation. To see this, let $\bm{e}_i \in \mathcal{S}^{d - 1}$ denote the $i$-th standard basis vector in $\mathbb{R}^d$, and set
\begin{align*}
    & ~ \mathcal{B} := \{\bm{e}_1, \ldots, \bm{e}_d\}, \text{ and } ~ \mathcal{L} := \left\{L = \mathsf{span}(\{\bm{e}_i\}_{i \in T}) ~|~ T \in \mathcal{T}^k \right\}
\end{align*}
in Definition~\ref{cond:linear-structure}. 

\subsubsection{Example 2: Path-Sparse PCA} \label{sec:PS-PCA-intro}

Another commonly used variant of union-of-linearly structured PCA is path-sparse PCA~\citep{asteris2015stay}, in which the support set of $\bm{v}_*$ forms a path on an underlying directed acyclic graph $G = (V,E)$. For a vertex $v$ in this graph, let $\delta_{\text{out}}(v)$ denote the out-neighborhood of $v$.

\begin{definition}
$(d,k)$-\textit{Layered Graph}. 
A directed acyclic graph $G = (V,E)$ is a $(d,k)$-layered graph if
\begin{itemize}
    \item
     $V = \{v_s, v_t\} \cup \widetilde{V}$ such that $|\widetilde{V}| = d-2$ and $v_s,v_t \notin \widetilde{V}$.
    \item $\widetilde{V} = \cup_{i=1}^{k} V_i$ where $V_i \cap V_j = \emptyset$ for all $i \neq j \in [k]$ and $|V_1| =  \cdots =  |V_k| = \frac{d - 2}{k}$. 
    \item $\delta_{\text{out}}(v) = V_{i + 1}$ for all $v \in V_{i}$ and $i = 1, \ldots, k - 1$, and 
        \item $\delta_{\text{out}}(v_s) = V_1$ and $\delta_{\text{out}}(v) = \{v_t\}$ for all $v \in V_k$. 
\end{itemize}
\end{definition}

Let $G = (V,E)$ be a $(d,k)$-layered graph and we define the collection of vertex sets
\begin{align*}
 \mathcal{P}^k := \big\{ P \subseteq V ~|~ v_s,v_t \in P \text{ and } |P\cap V_i|=1 ~ \forall ~ i\in [k]\big\}. 
\end{align*}
Once again, we consider the natural bijection between the coordinates of any $d$-dimensional vector and the vertices of a $(d, k)$-layered graph, and a vector $\bm{v}_*$ is said to be $k$-\emph{path-sparse} if $\mathsf{supp}(\bm{v}_*) \in \mathcal{P}^k$. It is straightforward to see that the set of all $k$-\emph{path-sparse} vectors satisfies the union of linear structures condition in Definition~\ref{cond:linear-structure} with 
\begin{align*}
    & ~ \mathcal{B} := \{\bm{e}_1, \ldots, \bm{e}_d\}, \text{ and } ~ \mathcal{L} := \left\{L = \mathsf{span}(\{\bm{e}_i\}_{i \in P}) ~|~ P \in \mathcal{P}^k \right\}.
\end{align*}

\subsection{Notation} \label{sec:notations}
We use $\bm{I}_{d \times d}$ to denote the $d$-by-$d$ identity matrix, and $\lambda_i(\bm{M})$ to denote the $i$-th largest eigenvalue of a symmetric matrix $\bm{M}$. 
We use $\bm{X} := [\bm{x}_1 ~|~ \cdots ~|~ \bm{x}_n]^{\top} \in \mathbb{R}^{n \times d}$ to denote the sample matrix where the $i$-th row of $\bm{X}$ is the $i$-th sample $\bm{x}_i$. The sample covariance matrix is given by $\widehat{\bm{\Sigma}} := \frac{1}{n} \bm{X}^{\top} \bm{X}$, and we let 
\begin{align} \label{eq:noise-def}
\bm{W} := \widehat{\bm{\Sigma}} - \bm{\Sigma}
\end{align}
denote the $d \times d$ matrix of noise. For any linear subspace $L \subseteq \mathbb{R}^d$ and its projection matrix $\bm{P}_L \in \mathbb{R}^{d \times d}$, we use $\widehat{\bm{\Sigma}}_L := \bm{P}_L^{\top} \widehat{\bm{\Sigma}} \bm{P}_L$ to denote the sample covariance matrix restricted to the subspace $L$. We also use the analogous notation $\bm{\Sigma}_L := \bm{P}_L^{\top} \bm{\Sigma} \bm{P}_L$ and $\bm{W}_L := \bm{P}_L^{\top} \bm{W} \bm{P}_L$. 
We index the subspaces $L_1, \ldots, L_M$ in some consistent lexicographic order.
We reserve the notation $\mathcal{M} := \bigcup_{m = 1}^M L_m$ to denote the set containing $\bm{v}_*$, and the notation $L_* \in \{L_{1},\dots,L_{M}\}$ to denote the specific linear subspace that contains $\bm{v}_*$, with ties broken lexicographically.
We let $\mathcal{S}^{d - 1}:= \{\bm{v} \in \mathbb{R}^{d}: \|\bm{v}\|_2 = 1\}$ denote the unit $\ell_2$-sphere in $d$-dimensional Euclidean space. For any subspace $L$, let $\widehat{\bm{v}}_L := \argmax_{\bm{v} \in \mathcal{S}^{d-1}} \bm{v}^{\top} \widehat{\bm{\Sigma}}_L \bm{v} = \argmax_{\bm{v} \in \mathcal{S}^{d-1} \cap L} \bm{v}^{\top} \widehat{\bm{\Sigma}} \bm{v}$ be the leading eigenvector of the restricted sample covariance $\widehat{\bm{\Sigma}}_L$. For an arbitrary symmetric matrix $\bm{M} \in \mathbb{R}^{d \times d}$ and set $S \subseteq \mathbb{R}^d$, define the scalar 
\begin{align}\label{eqs:rho}
    \rho(\bm{M}, S) := \max_{\|\bm{v}\|_2 = 1,\bm{v} \in S} \big| \bm{v}^{\top} \bm{M} \bm{v} \big|.
\end{align}
For two sequences of non-negative reals $\{f_n\}_{n \geq 1}$ and $\{g_n\}_{n \geq 1}$, we use $f_n \gtrsim g_n$ to indicate that there is a universal positive constant $C$ such that $f_n \leq C g_n$ for all $n \geq 1$. We also use standard order notation $f_n = O(g_n)$ to indicate that $f_n \lesssim g_n$ and $f_n = \tilde{O}(g_n)$ to indicate that $f_n \lesssim g_n \ln^c n$ for some universal constant $c$. We say that $f_n = \Omega(g_n)$ (resp. $f_n = \tilde{\Omega}(g_n)$) if
$g_n = \Omega(f_n)$ (resp. $g_n = \Omega(f_n)$). We use $f_n = \Theta(g_n)$ (resp. $f_n = \tilde{\Theta}(g_n)$) if $f_n = O(g_n)$ and $f_n = \Omega(g_n)$ (resp. $f_n = \tilde{O}(g_n)$ and $f_n = \tilde{\Omega}(g_n)$). We say that $f_n = o(g_n)$ (resp. $f_n = \tilde{o}(g_n)$) when $\lim_{n \rightarrow \infty} f_n / g_n = 0$ (resp. $\lim_{n \rightarrow \infty} f_n / (g_n \ln^c n) = 0$ for some universal constant $c$). We also use $f_n = \omega(g_n)$ to indicate that $\lim_{n \rightarrow \infty} f_n / g_n = \infty$. Throughout, we use $c, c_1, c_2, \ldots$ and $C, C_1, C_2, \ldots$ to denote universal positive constants, and their values may change from line to line.

%% file: stat-optest.tex
\section{General results} \label{sec:general-results}

In this section, we present our general results for union of linearly structured PCA, covering both fundamental limits of estimation and local convergence properties of a projected power method. Recall the notation $\rho(\bm{M}, S)$ from Eq.~\eqref{eqs:rho} for any symmetric matrix $\bm{M} \in \mathbb{R}^{d \times d}$ and set $S \subseteq \mathbb{R}^d$. We let
\begin{align}\label{exhaustive-search-estimator} 
    \ESest := \argmax_{\bm{v} \in \mathcal{M} \cap \mathcal{S}^{d-1}} \bm{v}^{\top} \widehat{\bm{\Sigma}} \bm{v}
\end{align} 
denote the general exhaustive search estimator.

\subsection{Fundamental limits of estimation} \label{sec:fund-limits}

We begin by studying the fundamental limits of estimation for linearly structured PCA, without computational considerations. These serve as baselines for the results to follow. We first introduce some notation before presenting main results. Recall $\mathcal{L} = \{L_1, \ldots, L_M\}$, the collection of $M$ linear subspaces, and subsets of bases $\mathcal{B}_m \subseteq \mathcal{B} = \{\phi_1, \ldots, \phi_d\}$ such that $L_m = \text{span}(\mathcal{B}_m)$. For each $m \in [M]$, define the characteristic vector $\bm{z}_{m} \in \{0,1\}^d$ of each subset $\mathcal{B}_m$ as follows
\begin{align}\label{definition-z-m}
    \bm{z}_{m}(i) := \left\{
    \begin{array}{lll}
        1 & \text{if } \quad \phi_i \in \mathcal{B}_m \\
        0 & \text{if } \quad \phi_i \notin \mathcal{B}_m
    \end{array}
    \right., \quad \text{for all} \; i \in [d],
\end{align}
where $\bm{z}_{m}(i)$ is the $i$-th entry of $\bm{z}_{m}$. We further define 
\begin{align}\label{definition-i-star}
    i_* := \argmax_{i \in [d]} \sum_{m=1}^{M}\bm{z}_{m}(i) 
\end{align}
as the index with the most ones among $\{\bm{z}_m\}_{m = 1}^M$, breaking ties lexicographically. In words, this is the index of the basis vector that appears in the most subspaces. Now let 
\begin{align}\label{definition-Z-star}
    \mathcal{Z}_{*} := \left\{ \bm{z}_{m} \in \{\bm{z}_1, \ldots, \bm{z}_M\} ~ | ~ \bm{z}_{m}(i_*) = 1 \right\}.
\end{align}
be the set of characteristic vectors with $\bm{z}_{m}(i_*) = 1$. For any fixed integer $r \geq 0$ and characteristic vector $\bm{z} \in \{\bm{z}_{m}\}_{m=1}^{M}$, we use $\mathcal{N}_H(\bm{z}; r) := \left\{ \bm{z}' \in \mathcal{Z}_{*} \;|\; \delta_H(\bm{z}, \bm{z}') \leq r \right\}$ to denote the neighborhood of $\bm{z}$ in $\mathcal{Z}_{*}$ with Hamming ball distance $\delta_H(\bm{z}, \bm{z}') := |\{i: \bm{z}(i) \neq \bm{z}'(i) \}|$ at most $r$. We further state Assumption~\ref{assump:minimax-assumption} for the minimax lower bound.

\begin{assumption} \label{assump:minimax-assumption}
This assumption has two parts:
\begin{description}
    \item[(a)] For all $m \in [M]$, $|\mathcal{B}_m| = k$ for some $k\leq d$. 
    \item[(b)] There exists $\xi \in [3/4,1)$ such that
    \begin{align} 
        \frac{|\mathcal{Z}_{*}|}{\max_{\bm{z} \in \mathcal{Z}_{*}}|\mathcal{N}_H(\bm{z}; 2(1-\xi)k)|} \geq 16. \label{eq:minimax-assump}
    \end{align}
\end{description}
\end{assumption}

Assumption~\ref{assump:minimax-assumption}(a) is clearly satisfied by vanilla sparse PCA, tree-sparse PCA, and path-sparse PCA. For a general $\mathcal{L}$, one can always set $k = \max_{m \in [M]} |\mathcal{B}_m|$. Assumption~\ref{assump:minimax-assumption}(b), on the other hand, controls the ratio of the sizes between the largest neighborhood $\mathcal{N}_H(\bm{z};2(1-\xi) k)$ (among $\bm{z} \in \mathcal{Z}_*$) and $\mathcal{Z}_*$. Geometric intuition for this assumption will be provided shortly.
It is worth noting that the specific constant $16$ in Ineq.~\eqref{eq:minimax-assump} is arbitrary, and any constant greater than $2$ can be used. We choose $16$ for simplicity and convenience in presenting the subsequent theoretical results (Theorem~\ref{thm:fund-limits}(b)). 

We are now poised to state the main result of this subsection. Recall that $L_*$ denotes the subspace containing the vector $\bm{v}_*$. Let $\widehat{L} \in \mathcal{L}$ be the linear subspace such that $\ESest \in \widehat{L}$ (once again breaking ties lexicographically) and let $\widehat{F} := \mathsf{conv}(\widehat{L} \cup L_*)$. 

\begin{theorem}\label{thm:fund-limits}
Suppose the union-of-linear structures condition in Definition~\ref{cond:linear-structure} holds.

\noindent (a) Let $\ESest$ be defined in equation~\eqref{exhaustive-search-estimator}. Without loss of generality, suppose $\langle \bm{v}_*, \widehat{\bm{v}}_{\mathsf{ES}} \rangle \geq 0$.  Then for all $\bm{v}_* \in \mathcal{S}^{d - 1} \cap \mathcal{M}$, we have
\begin{subequations}
    \begin{align} \label{eq:fund-limit-a}
        \|\widehat{\bm{v}}_{\mathsf{ES}} - \bm{v}_*\|_2 \leq \frac{2\sqrt{2}}{\lambda} \rho \big(\bm{W}, \widehat{F} \big),
    \end{align}
where the function $\rho$ is defined in Eq.~\eqref{eqs:rho}.

\noindent (b) Let $\xi \in [3/4,1)$ such that Assumption~\ref{assump:minimax-assumption} holds. 
We have the minimax lower bound
    \begin{align}
        &\inf_{\widehat{\bm{v}}} \; \sup_{\bm{v}_* \in \mathcal{S}^{d - 1} \cap \mathcal{M}} \mathbb{E} \left[ \left\| \widehat{\bm{v}} \widehat{\bm{v}}^{\top} - \bm{v}_* \bm{v}_*^{\top}  \right\|_F \right]   \geq \frac{\sqrt{2(1 - \xi)}}{4} \; \cdot  \nonumber
        \\&\min\Bigg\{1, \sqrt{\frac{1 + \lambda}{8 \lambda^2}} \cdot
        \sqrt{ \log \bigg( \frac{\big|\mathcal{Z}_* \big|}{ \max_{\bm{z} \in \mathcal{Z}_{*}}\big|\mathcal{N}_H(\bm{z}; 2(1 - \xi) k) \big| } \bigg)} \Bigg\}. \label{eq:fund-limit-b}
    \end{align}
    \end{subequations}
    Here, the infimum is taken over all measurable functions of the observations $\{ \bm{x}_i \}_{i = 1}^n$, which are drawn i.i.d. from the distribution $\mathcal{D}(\lambda; \bm{v}_*)$.
\end{theorem}

Theorem~\ref{thm:fund-limits}(a) provides a deterministic upper bound on the $\ell_2$ error between the estimate $\widehat{\bm{v}}_{\mathsf{ES}}$ and the ground truth $\bm{v}_*$, showing that this error can be bounded on the order $\rho\big(\bm{W}, \widehat{F}\big)$ for any fixed $\lambda$. 
We provide the proof of this result in Section~\ref{app:fund-limits-up} of the supplementary material. While the result is deterministic, we will see that Eq.~\eqref{eq:fund-limit-a} nearly matches the minimax lower bound Eq.~\eqref{eq:fund-limit-b} for our special cases of interest. Consequently, we use Theorem~\ref{thm:fund-limits}(a) as a heuristic baseline to assess the performance of efficient algorithms. 
 


On its own, Theorem~\ref{thm:fund-limits}(b) provides a minimax lower bound that depends on the local structure of $\mathcal{M}$ around any choice of ground truth $\bm{v}_*$. 
The proof uses the generalized Fano inequality \citep{verdu1994generalizing}, and we construct a rich packing set $\mathcal{V}_{\epsilon}$ in $\mathcal{S}^{d - 1} \cap \mathcal{M}$ (i.e., $\mathcal{V}$ in Proposition~\ref{prop:g-fano-method} of Section~\ref{app:fund-limits-lb} in the supplementary material) such that the points in $\mathcal{V}_{\epsilon}$ are $\mathcal{O}(\epsilon)$ separated in some appropriate distance measure. In contrast to existing proofs for sparse PCA \citep{vu2012minimax} and path PCA \citep{asteris2015stay}, the set $\mathcal{V}_{\epsilon}$ here is constructed so that there exists a common support index (i.e., the index $i_*$, defined in Eq~\eqref{definition-i-star}) for every point $\bm{v} \in \mathcal{V}_{\epsilon}$ that one can use to construct the packing. 
%
On a related note, a paper by \citet{cai2021optimal} studies the minimax risk of a general structured principal subspace estimation problem, including vanilla sparse PCA as a special case. These bounds are phrased in terms of critical inequalities that arise from local packing numbers (see~\cite{yang1999information,wainwright2019high}). Our lower bound instead takes a more global approach, which we show suffices for union-of-linear structure.
In particular, the minimax lower bound~\eqref{eq:fund-limit-b} is controlled by the relative ratio between $|\mathcal{N}_H(\bm{z};2(1-\xi) k)|$ and $|\mathcal{Z}_*|$: Our assumption in Ineq.~\eqref{eq:minimax-assump} avoids the scenario that many linear subspaces heavily overlap on a few bases.  

Finally, it is instructive to note that Theorem~\ref{thm:fund-limits}(b) recovers the existing minimax lower bound for vanilla sparse PCA [Theorem 2.1, \cite{vu2012minimax}]. 
 Indeed, supposing that $d \gg k$ and applying Theorem~\ref{thm:fund-limits}(b) for sparse PCA, we obtain the known lower bound
\begin{align} \label{eq:SPCA-known}
    &\inf_{ \widehat{\bm{v}} } \; \sup_{ \bm{v_*} \in \mathcal{S}^{d - 1} ,  \| \bm{v_*} \|_0 \leq k }  \mathbb{E} \;\left[  \big\| \widehat{\bm{v}} \widehat{\bm{v}}^{\top} - \bm{v}_* \bm{v}_*^{\top}   \big\|_F \right]  \gtrsim  \min\bigg\{1,  \sqrt{\frac{1 + \lambda}{8 \lambda^2}} \sqrt{\frac{k \log d}{n}} \bigg\}. 
\end{align}
The proof of inequality~\eqref{eq:SPCA-known} is provided in Section~\ref{min-max-pca-proof} for completeness. In Section~\ref{sec:specific-examples}, we provide novel corollaries for tree-sparse PCA and path-sparse PCA.


%% file: PPM-EP.tex

\subsection{A locally convergent projected power method} \label{sec:PPM-EM} 

In Section~\ref{sec:fund-limits}, we studied the fundamental limits of the problem, where our upper bounds were achieved by the exhaustive search estimator $\ESest$.
Given the computational challenge of searching over every linear subspace $L \in \mathcal{L}$, we propose the following iterative projected power method (Algorithm~\ref{alg:PPM}) and show that with access to a suitable exact projection oracle, it locally converges to a statistical neighborhood of the ground truth.

\begin{definition}[Exact projection] \label{assump:exact-proj}
For all $\bm{v} \in \mathbb{R}^d$, let 
\[ \Pi_{\mathcal{M}} (\bm{v}) := \argmin_{\bm{v}' \in \mathcal{M}}\; \|\bm{v}' - \bm{v}\|_2 = \argmin_{\bm{v}' \in L_{m},m \in [M]} \; \| \bm{v}' - \bm{v}\|_2,
\]
where ties between subspaces are broken lexicographically.
\end{definition}
Owing to the tie-breaking rule, this projection is always unique. As we will see in Section~\ref{sec:specific-examples}, an exact projection oracle $\Pi_{\mathcal{M}}$ can be constructed efficiently (in time nearly logarithmic in $M$) in some specific examples of union-of-linearly structured PCA. 
%
%
We are now in a position to present the projected power method, described formally in Algorithm~\ref{alg:PPM}.

\begin{algorithm}
\caption{Projected Power Method}
\label{alg:PPM}
\textbf{Input:} Sample covariance matrix $\widehat{\bm{\Sigma}}$.
\begin{algorithmic}[1]
\State \textbf{Initialize} with a vector $\bm{v}_0 \in \mathcal{M} \cap \mathcal{S}^{d - 1}$. 
\For{$t = 0, 1, \ldots, T - 1$}
\State Compute $\tilde{\bm{v}}_{t + 1} = \widehat{\bm{\Sigma}} \bm{v}_t / \big\|\widehat{\bm{\Sigma}} \bm{v}_t \big\|_2$. \label{alg:PPM-compute}
\State Project $\bm{v}_{t + 1}^{\mathcal{M}} = \Pi_{\mathcal{M} }(\tilde{\bm{v}}_{t + 1})$. 
\label{alg:PPM-project}
\State Normalize to unit sphere $\bm{v}_{t + 1} = \frac{ \bm{v}_{t + 1}^{\mathcal{M}} }{ \|\bm{v}_{t + 1}^{\mathcal{M}} \|_2} \in \mathcal{M} \cap \mathcal{S}^{d - 1}$. 
\EndFor
\end{algorithmic}
\textbf{Output:} $\bm{v}_T$.  
\end{algorithm}

Using the notation $\rho(\bm{M}, S)$ from Eq.~\eqref{eqs:rho}, we define 
$
F^{*} := \argmax_{F} \rho\big(\bm{W}, F \big)\;\text{s.t.}\; F = \text{conv}\big(L_{m_1} \cup  L_{m_2} \cup L_{m_3} \big),\; \forall \;  m_1,m_2,m_3 \in [M].
$
We now state the definition of a ``good region"; Theorem~\ref{thm:convergence} to follow shows that once Algorithm~\ref{alg:PPM} is initialized in this region, it will converge geometrically to a neighborhood of the ground truth $\bm{v}_*$.
\begin{definition}[Good region]\label{good-region}
For eigen gap $\lambda > 2 \rho(\bm{W}, F^{*})$, we define the good region 
\begin{align*} 
&\mathbb{G}(\lambda) = \big\{ \bm{v} \in \mathcal{M} \cap \mathcal{S}^{d - 1}: \langle \bm{v}, \bm{v}_{*} \rangle \geq t_1(\lambda) \big\},\quad \text{where}
\\& \quad t_1(\lambda):= \frac{4}{\lambda + 1 - \rho(\bm{W}, F^{*})} + \frac{5\rho(\bm{W}, F^{*})}{\lambda - 2\rho(\bm{W}, F^{*})}.
\end{align*}
\end{definition}
\noindent Note that $\mathbb{G}(\lambda)$ becomes a larger set as $\lambda$ increases. To ensure that such a good region is non-empty, it is necessary to have $t_1(\lambda) < 1$. In our proof of convergence (see Appendix~\ref{app:PPM-EM}), we require that the good region is not just non-empty but large enough. In particular, we require the eigengap $\lambda$ to be large enough so that
\begin{align}\label{eigen-gap-condition}
     t_2(\lambda) := \frac{4}{\lambda + 1 - \rho(\bm{W}, F^{*})} + \frac{10\rho(\bm{W}, F^{*})}{\lambda - 2\rho(\bm{W}, F^{*})} < 1. 
\end{align}
Note that this automatically ensures that $t_1(\lambda) < 1$ since $t_{2}(\lambda) > t_1(\lambda)$. 
We are now poised to state our main result for this subsection.
\begin{theorem} \label{thm:convergence}
Suppose the eigengap satisfies $\lambda > 2 \rho(\bm{W}, F^{*})$, and condition~\eqref{eigen-gap-condition} holds. Suppose in Algorithm~\ref{alg:PPM-project} the initialization satisfies $\bm{v}_{0} \in \mathbb{G}(\lambda)$. Then for all $t\geq 1$, we have
\begin{align}\label{ineq-thm-convergence}
    \| \bm{v}_{t+1} - \bm{v}_{*} \|_2 \leq \frac{1}{2^{t}} \cdot \| \bm{v}_{0} - \bm{v}_{*} \|_2 + \frac{6\rho(\bm{W}, F^{*})}{\lambda - 2\rho(\bm{W}, F^{*})}.
\end{align}

\end{theorem}

The proof of Theorem~\ref{thm:convergence} can be found in Section~\ref{app:PPM-EM}. Once (a) an exact projection oracle $\Pi_{\mathcal{M}}$ is accessible; and (b) an initial vector $\bm{v}_0$ is in the good region $\mathbb{G}(\lambda)$, Theorem~\ref{thm:convergence} ensures a deterministic convergence result. 
Note that the result parallels that of~\cite{yuan2013truncated} for vanilla sparse PCA, where $\rho(\bm{W}, F^{*}) = O(\sqrt{k\log d / n})$. The key additional technique that we use to control the error accumulated at each iteration is based on an ``equivalent replacement'' step; see Section~\ref{proof-lemma-equivalence-of-v}. 

The projected power method was also recently analyzed by~\citet{liu2021generative} for PCA with generative models when given access to an exact projection oracle.
While they also proved local geometric convergence results given access to a sufficiently correlated initialization, there are significant differences in the assumptions of that paper and our own. First, our work imposes the union-of-linear structure assumption on the principal component, which is an altogether different structural assumption from a generative model. Given this, our proof techniques differ significantly from those of~\citet{liu2021generative}. Second, 
we present a computationally efficient initialization method and matching evidence of computational hardness for two prototypical examples; see below. 

\subsection{Initialization method} \label{sec:initial-method-general}

Recall that Theorem~\ref{thm:convergence} requires an initialization $\bm{v}_0$ in the good region $\mathbb{G}(\lambda)$. In this subsection, we provide such an initialization method (see Algorithm~\ref{alg:initialization}) that works when given a projection oracle, provided the following assumption holds. 



\begin{assumption}\label{assump:M-set-initial}
The set $\mathcal{M}$ in  satisfies
\[
    \mathcal{M} \subseteq \{\bm{v} \in \mathbb{R}^{d} : \|\bm{v}\|_{0} = k\},\quad \text{where} \quad k \in \mathbb{N}.
\]
\end{assumption}

Assumption~\ref{assump:M-set-initial} is not guaranteed by Definition~\ref{cond:linear-structure}, but includes many typical examples. For instance, the sets $\mathcal{T}^k$ and $\mathcal{P}^k$ for tree-sparse or path-sparse PCA, respectively, satisfy Assumption~\ref{assump:M-set-initial} in addition to union-of-linear structure. Moreover, if the orthonormal matrix $\bm{\Phi}$ is known, one can reformulate the problem as  estimating the structured-sparse vector $\bm{\Phi}^{\top} \bm{v}_{*}$ from observations $\{\bm{\Phi}^{\top} \bm{x}_i\}_{i = 1}^n$ (see Remark~\ref{rem:eq-structured}).



\begin{algorithm}
\caption{Initialization Method -- Covariance Thresholding with Projection Oracle}
\label{alg:initialization}
\textbf{Input.} $\{\bm{x}_i\}_{i = 1}^{n}$, parameter $k \in \mathbb{N}$, thresholding parameter $\tau$ and exact projection $\Pi_{\mathcal{M}}$. 
\begin{algorithmic}[1]
\State Compute covariance matrix $\widehat{\bm{\Sigma}} = \sum_{i = 1}^n \bm{x}_i \bm{x}_i^{\top} / n$. 
\State Set the soft-thresholding matrix $\widehat{\bm{G}}(\tau)$ as:
\begin{align*}
&\text{If} \quad \quad~~~ \widehat{\bm{\Sigma}}_{ij} - [\bm{I}_d]_{ij} \geq \tau/\sqrt{n}, \quad ~~ \text{then} \quad [\widehat{\bm{G}}(\tau)]_{ij} = \widehat{\bm{\Sigma}}_{ij} - [\bm{I}_d]_{ij} - \tau/\sqrt{n};\\
&\text{else if} \quad \widehat{\bm{\Sigma}}_{ij} - [\bm{I}_d]_{ij} \leq -\tau/\sqrt{n}, \quad \text{then} \quad [\widehat{\bm{G}}(\tau)]_{ij} = \widehat{\bm{\Sigma}}_{ij} - [\bm{I}_d]_{ij} + \tau/\sqrt{n};
\\&\text{else} \quad \quad [\widehat{\bm{G}}(\tau)]_{ij} = 0.
\end{align*}
\State Compute  $\widehat{\bm{v}}_{\textup{soft}} := \max_{\|\bm{v}\|_2 = 1} \bm{v}^{\top} \widehat{\bm{G}}(\tau) \bm{v}$ as the leading eigenvector of $\widehat{\bm{G}}(\tau)$.
\State Project $\bm{v}_0 :=  \Pi_{\mathcal{M}}(\widehat{\bm{v}}_{\textup{soft}}) / \|\Pi_{\mathcal{M}}(\widehat{\bm{v}}_{\textup{soft}})\|_2$.
\end{algorithmic}
\textbf{Return} $\bm{v}_0 \in \mathcal{S}^{d - 1} \cap \mathcal{M}$. 
\end{algorithm}  
 

\begin{theorem} \label{thm:initialization-method}
Suppose Assumption~\ref{assump:M-set-initial} holds and $k^2 \leq d/e$. There exists a tuple of universal, positive constants $(C_1,C_2,C_3,C)$ such that the following holds. Suppose $n\geq \max\{C\log d,k^{2}\}$ and let $\tau_{*} := C_1 \max\{\lambda, 1\} \sqrt{\log (d/k^2)}$. Set the thresholding level according to 
\begin{align}\label{initia-threshold-tau}
    \tau := \left\{
    \begin{array}{lll}
        \tau_* & \textup{when  } \tau_* \leq \sqrt{\log d} / 2, \\
        C_2 \tau_* & \textup{when  } \tau_* \geq \sqrt{\log d} / 2, \\
        0 & \textup{otherwise.}
    \end{array}
    \right. 
\end{align}
Then for any $0<c_{0} <1$, if
\begin{align*}
    n \geq  n_0(c_0) := \frac{18 C_3 \max\{\lambda^2, ~ 1\} k^2}{2(1 - c_0)^2 \lambda^2} \log(d/k^2),
\end{align*}
then the initial vector $\bm{v}_0 \in \mathcal{S}^{d - 1} \cap \mathcal{M}$ obtained from Algorithm~\ref{alg:initialization} satisfies $\langle \bm{v}_0, \bm{v}_* \rangle \geq c_0$ with probability $1 - C'\exp(- \min\{\sqrt{d}, n\}/C')$ for some positive constant $C'$.

%
\end{theorem}


The proof of Theorem~\ref{thm:initialization-method}, which builds on existing results in \cite{deshpande2016sparse}, can be found in Appendix~\ref{app:initialization}. Let us now show that the output of this algorithm serves as a valid initialization for the projected power method, since this is not immediate given that the event $\mathcal{E}_1 = \{\langle \bm{v}_0, \bm{v}_{*} \rangle \geq c_0\}$  depends on the samples $\{\bm{x}_i\}_{i = 1}^n$.
%
%
Recall the quantities $t_1(\lambda)$ and $t_2(\lambda)$ in Definition~\ref{good-region} and Eq.~\eqref{eigen-gap-condition}, respectively. 
In Theorem~\ref{thm:initialization-method}, set $c_0 := t_2(\lambda)$ and recall that $t_2(\lambda) > t_2(\lambda)$ by definition. Suppose $\lambda \geq 5$ for convenience. Then it can be shown that the event $\mathcal{E}_2 = \{ t_1(\lambda) < t_2(\lambda) = c_0 < 7/8\} \subseteq \{\rho(\bm{W}, F^*) < 9/400\}$ occurs with probability at least $1 - C' \exp(- \min\{\sqrt{d}, n\} / C')$. 
%
%
Consequently, on the high probability event $\mathcal{E}_1 \cap \mathcal{E}_2$, we have that the initialization $\bm{v}_{0}$ obtained by Algorithm~\ref{alg:initialization} satisfies $\bm{v}_0 \in \mathbb{G}(\lambda)$. The projected power method can thus be employed after this initialization to guarantee convergence to a small neighborhood of $\bm{v}_*$.


A key feature of Theorem~\ref{thm:initialization-method} is the lower bound $n_0 = \Theta(k^2 \log(d / k^2) )$ on the number of samples required for the Algorithm~\ref{thm:initialization-method} to succeed. 
Note that this is of a strictly 
larger order than the number of samples required information-theoretically even for vanilla sparse PCA---this is a well-known phenomenon. In the next section, we show that even with the additional structure afforded by tree and path sparsity, this larger sample size is in some sense necessary for computationally efficient algorithms.

%% file: example-path-SPCA.tex
\section{End-to-end analysis for specific examples} \label{sec:specific-examples}

In this section, we provide end-to-end analyses for path-sparse and tree-sparse PCA, including results on their information-theoretic limits of estimation as well as the performance of the projected power method when initialized using covariance thresholding. We complement these with what may be considered as the main results of this section: matching suggestions of computational hardness. 

\subsection{Path-Sparse PCA} \label{sec:PS-PCA}

\subsubsection{Fundamental limits for Path-Sparse PCA} \label{sec:limits-PS-PCA}

Recall the notation $\mathcal{P}^k$ as the structure set of path-sparse PCA from Section~\ref{sec:PS-PCA-intro}. We write $\bm{v} \in \mathcal{P}^k$ if the support set satisfies $\mathsf{supp}(\bm{v}) \in \mathcal{P}^k$. We use 
\begin{align}
    \widehat{\bm{v}}_{\textsf{PS}} := \argmax_{\bm{v}} ~ \bm{v}^{\top} \widehat{\bm{\Sigma}} \bm{v} ~~\text{s.t.}~~ \bm{v} \in \mathcal{S}^{d - 1} \cap \mathcal{P}^k \label{eq:PS-PCA-est}
\end{align}  
to denote the corresponding estimate from exhaustive search. 

\begin{corollary}\label{coro:PS-PCA-fund-limits}
There exists a pair of positive constants $(c, C)$ such that the following holds.
\noindent (a) Without loss of generality, assume $\langle \bm{v}_*, \widehat{\bm{v}}_{\mathsf{PS}} \rangle \geq 0$. Then for any $c_1>0$ and $\bm{v}_* \in \mathcal{S}^{d - 1} \cap \mathcal{P}^k$, we have 
\[
 \big\|\widehat{\bm{v}}_{\mathsf{PS}} - \bm{v}_*\big\|_2 \leq C \left( \frac{1 + \lambda}{\lambda} \right) \sqrt{\frac{3 (\ln d - \ln k)k + c_1k}{n}}
 \] 
 with probability at least $1 - 2\exp(- c_1k)$.

\noindent (b) Suppose that $d\geq 16 k^2$ and $k\geq 4$. Then we have the minimax lower bound 
\begin{align*}
&\inf_{\widehat{\bm{v}}} \; \sup_{\bm{v}_* \in \mathcal{S}^{d - 1} \cap \mathcal{P}^k} \mathbb{E}\left[ \left\| \widehat{\bm{v}} \widehat{\bm{v}}^{\top} - \bm{v}_* \bm{v}_*^{\top}  \right\|_F \right] \geq  c \cdot \min\bigg\{1,  \sqrt{\frac{1 + \lambda}{8 \lambda^2}} \sqrt{\frac{k \cdot \big( \frac{\ln d}{2} - \ln k \big)}{n}} \bigg\}.
\end{align*}
Here, the infimum is taken over all measurable functions of the observations $\{ \bm{x}_i \}_{i = 1}^n$ drawn i.i.d. from the distribution $\mathcal{D}(\lambda; \bm{v}_*)$.
\end{corollary}

Corollary~\ref{coro:PS-PCA-fund-limits} is proved in Section~\ref{proof-coro-PS-PCA-fund-limits} of the supplementary material, and is based on Theorem~\ref{thm:fund-limits}. In particular, Corollary~\ref{coro:PS-PCA-fund-limits}(a) gives an upper bound on the estimation error of $\widehat{\bm{v}}_{\textsf{PS}}$ by showing that the statistical noise term\footnote{As expected, this term does not differ significantly from the corresponding term for vanilla sparse PCA, since the number of sparsity patterns for path sparse PCA $|\mathcal{P}^k|$ is on the order $(d/k)^k$.} $\rho(\bm{W}, P^*)$ is of the order $(\lambda + 1) \sqrt{k \cdot ( \ln d - \ln k)/n}$. 
The minimax lower bound obtained in Corollary~\ref{coro:PS-PCA-fund-limits}(b) is of the same order as the minimax lower bound given in [Theorem 1, \cite{asteris2015stay}] with the outer degree parameter $|\Gamma_{\text{out}}(v)| = (d - 2) / k$.

\subsubsection{Local convergence and initialization} \label{sec:initialization-PS-PCA}
\paragraph{Exact projection oracle} We build the exact projection oracle for path-sparse PCA $\Pi_{\mathcal{P}^k}$ by picking the component with the largest absolute value in each partition (layer) for a given $(d,k)$-layered graph $G$. The formal procedure is given in Algorithm~\ref{alg:projection-PSPCA} (see Section~\ref{app:initial-examples}), and has running time $O(d)$.

\begin{corollary} \label{coro:PS-PCA-PPM}
 Suppose the initiatialization $\bm{v}_0$ in Algorithm~\ref{alg:PPM} satisfies $\bm{v}_{0} \in \mathcal{P}^{k} \cap \mathcal{S}^{d-1}$ and $\langle \bm{v}_0,\bm{v}_* \rangle \geq 1/2$. There exists a tuple of universal positive constants $(c,C_1,C_2,C_3)$ such that for $\lambda \geq C_1$, $n\geq C_2k\ln(d)$,  and all $t \geq 1$, the iterate $\bm{v}_t$ from Algorithm~\ref{alg:PPM} satisfies 
 \[
 \| \bm{v}_{t} - \bm{v}_{*} \|_2 \leq \frac{1}{2^t} \cdot \| \bm{v}_{0} - \bm{v}_{*} \|_2 +  C_3 \sqrt{\frac{k(2\ln d - \ln k)}{n}},
 \] 
 with probability at least $1-\exp(-ck)$. 
\end{corollary}
Corollary~\ref{coro:PS-PCA-PPM} is proved in Section~\ref{proof-coro-PS-PCA-PPM}
by applying Theorem~\ref{thm:convergence}. 
%

The final problem is to obtain an initialization $\bm{v}_0$.
To do so, note that the set $\mathcal{P}^k$ satisfies Assumption~\ref{assump:M-set-initial}, leading to the following corollary of Theorem~\ref{thm:initialization-method}.

\begin{corollary} \label{coro:initial-PS-PCA}
Assume $k^2 \leq d / e$. There exists a pair of universal positive constants $(C, C')$ such that if $n\geq \max\{C\log d,k^{2}\}$ and $n \geq C'\max \big\{1,\lambda^{-2} \big\} \log(d/k^2) k^{2}, $ then the initial vector $\bm{v}_0 \in \mathcal{S}^{d - 1} \cap \mathcal{P}^k$ obtained from Algorithm~\ref{alg:initialization} satisfies $\langle \bm{v}_0, \bm{v}_* \rangle \geq 7/8$ with probability $1 - C'\exp(- \min\{\sqrt{d}, n\}/C')$. 
\end{corollary}

In words, Corollary~\ref{coro:initial-PS-PCA} provides an initialization method whose outputs can be used for the general projected power method (Algorithm~\ref{alg:PPM}) for path-sparse PCA when the number of samples\footnote{The constant $7/8$ in $\langle \bm{v}_0, \bm{v}_* \rangle \geq 7/8$ can be replaced by any positive constant within $(0, 1)$  provided it ensures the good region condition $\langle \bm{v}_0, \bm{v}_* \rangle > t_2(\lambda)$.} $n \gtrsim k^2 \log(d/k^2)$. 

As previously mentioned, there is a gap between the condition $n \gtrsim k$ required for Corollary~\ref{coro:PS-PCA-PPM} and the stronger condition above. We will now show evidence that $k^{2}$ samples are necessary. In particular, 
we will show that no randomized polynomial-time algorithm can ``solve" (i.e. produce a consistent estimate for) path-sparse when $n \ll k^2$, provided we assume the  average-case hardness of the secret-leakage planted clique problem. This can be regarded as the main takeaway for path-sparse PCA: The additional structure has minimal effect on its statistical and computational limits.


\subsubsection{Average-Case Hardness of Path Sparse PCA}\label{sec:examples-average-hard}


This section focuses on the average-case hardness of the path sparse PCA, which is obtained via a reduction from the $K$-partite planted clique (PC) detection problem, which is in turn conjectured to be hard. 

\begin{definition} \label{defn:SL-PC}
\textbf{Secret Leakage $\text{PC}_{\mathcal{D}}$ Detection Problem,} \cite{brennan2020reducibility}. Given a distribution $\mathcal{D}$ on $K$-subsets of $[N]$, let $\mathcal{G}_{\mathcal{D}}(N, K, 1/2)$ be the distribution on $N$-vertex graphs sampled by first sampling $G \sim \mathcal{G}(N, 1/2)$ and $S \sim \mathcal{D}$ independently and then planting a $K$-clique on the vertex set $S$ in $G$. The secret leakage $\text{PC}_{\mathcal{D}}$ detect problem $\text{PC}_{\mathcal{D}}(N, K, 1/2)$ is defined as the resulting hypothesis testing problem between 
\[
	H_0: ~ G \sim \mathcal{G}(N, 1/2) \quad \text{and} \quad H_1: ~ G \sim \mathcal{G}_{\mathcal{D}}(N, K, 1/2).
\] 
\end{definition}

Now consider the following $K$-partite PC as a special case of the secret leakage $\text{PC}_{\mathcal{D}}$ detection problem. 

\begin{definition} \label{def:KPC}
\textbf{$K$-Partite Planted Clique Detection Problem (with source and terminal).} The $K$-partite planted clique detection problem $K\text{-PC}(N, K, 1/2)$ is a special case of the secret leakage planted clique detection problem $\text{PC}_{\mathcal{D}}(N, K, 1/2)$. Here the vertex set of $G$ has two special vertices: source and terminal, and the remaining vertices are evenly partition into $K$ parts of size $(N - 2) / K$. The distribution $\mathcal{D}$ always picks source, terminal and uniformly picks one element at random in each part.  
\end{definition}

Like the well-known planted clique conjecture, the $K$-Partite PC problem $K\text{-PC}(N, K, 1/2)$ is believed to satisfy the following hardness conjecture. 

\begin{conjecture} \label{conj:KPC-hardness}
\textbf{$K$-Partite PC Hardness Conjecture}, restatement of \cite{brennan2020reducibility}.
Suppose that $\{\mathcal{A}_N\}$ is a sequence of randomized polynomial time algorithms $\mathcal{A}_N: \mathcal{G}_N \to \{0,1\}$ and $K_N$ is a sequence of positive integers satisfying that $\limsup_{N \rightarrow \infty} \log_N K_N < 1/2$ with $\mathcal{G}_N$ the set of graphs with $N$ nodes. Then if $G$ is an instance of $K\text{-PC}(N, K_N, 1/2)$, it holds that $\liminf_{N \rightarrow \infty} \left( \mathbb{P}_{H_0}[\mathcal{A}_N(G) = 1] + \mathbb{P}_{H_1}[\mathcal{A}_N(G) = 0] \right) \geq 1. $
\end{conjecture}

\begin{definition}\label{def:qual-est}
\textbf{Qualified Estimator.} 
A qualified estimator $\widehat{\bm{v}}(n, d_n, k_n, \lambda_n, \epsilon)$ for path-sparse PCA is a sequence of functions $\EST_n: \mathbb{R}^{d_n \times n} \to \mathbb{R}^{d_n}$ mapping $\{\bm{x}_i\}_{i = 1}^n \mapsto \widehat{\bm{v}}$ such that if the set of samples $\{\bm{x}_i\}_{i = 1}^n$ are drawn i.i.d. from $\mathcal{D}(\lambda_n, \bm{v}_*)$ for some $\bm{v}_* \in \mathcal{S}^{d_n - 1} \cap \mathcal{P}^{k_n}$ then $\liminf_{n \to \infty} \Pr \left\{ \| \widehat{\bm{v}} - \bm{v}_* \|_2 < \frac{1}{4} \right\} \geq \frac{1}{2} + \epsilon$ for some fixed $0 < \epsilon < 1/2$. 
\end{definition}

From this point onward, we do not make $\epsilon$ explicit when referring to a qualified estimator. It suffices for the reader to think of it as a small positive constant that does not depend on $n$.
Geometrically, a qualified estimator $\widehat{\bm{v}}$ exhibits proximity to the ground truth $\bm{v}_* \in \mathcal{S}^{d_n - 1} \cap \mathcal{P}^{k_n}$ with probability at least $1/2 + \epsilon$ as $n \rightarrow \infty$. Note that Definition~\ref{def:qual-est} does not require explicit control on the behavior of $\widehat{\bm{v}}$ for a general vector $\bm{v}_* \notin \mathcal{S}^{d_n - 1} \cap \mathcal{P}^{k_n}$.



It is also worth noting (using Corollary~\ref{coro:PS-PCA-PPM} and Corollary~\ref{coro:initial-PS-PCA} and the corresponding algorithms) 
that our end-to-end  estimator for path-sparse PCA is a polynomial-time computable qualified estimator provided $n \geq C k^2 \log(d /k)$ and $\lambda = \Omega(1)$. 

\begin{proposition} \label{prop:reduction-PathPCA}
There exists a universal constant $c > 0$ such that the following holds. Let $1/2 \leq \beta <1$ and $0 < \epsilon < 1/2$ be fixed. Here, we use integer $\subindex$ as our index parameter. Suppose the sequence of parameters $\{(k_{\subindex}, d_{\subindex}, \lambda_{\subindex}, \tau_{\subindex})\}_{ \subindex \in \mathbb{N}}$ is in the parameter regime 
\[k_{\subindex} = \lceil \subindex^{\beta} \rceil, \quad d_{\subindex} = \subindex, \quad \lambda_{\subindex} = \frac{k_{\subindex}^2}{\tau_{\subindex} \cdot \subindex } \cdot \frac{(\log 2)^2}{4 (6 \log (\subindex) + 2 \log 2)},
\] 
where $\tau_{\subindex}$ is an arbitrarily slowly growing function of $\subindex$. If the K-Partite PC hardness conjecture (Conjecture~\ref{conj:KPC-hardness}) holds, then there is no qualified estimator $\widehat{\bm{v}}(n_{\subindex}, d_{\subindex}, k_{\subindex}, \lambda_{\subindex}, \epsilon)$ running in time polynomial in $d_{\subindex}$ when the sample size $n_{\subindex}$ satisfies $n_{\subindex} \leq c \left( \frac{k_{\subindex}^2}{2 \tau_{\subindex} \log k_{\subindex}} \right).$ 
\end{proposition}
%
The proof of Proposition~\ref{prop:reduction-PathPCA} is given in Section~\ref{app:reduction-PathPCA} of the supplementary material.
%
In particular, when the eigengap satisfies\footnote{This can be ensured for dimension $d_j = j$ growing such that $\frac{k_j}{\tau_j d_j \log d_j} = \Theta(1)$.} $\lambda = \Theta(1)$, it shows that $n = \widetilde{\Omega} \left(k^2\right)$ is necessary for computationally efficient estimation. 

%% file: example-tree-SPCA.tex
\subsection{Tree-sparse PCA} \label{sec:TS-PCA}

\subsubsection{Fundamental limits for Tree-Sparse PCA} \label{sec:limits-TS-PCA}

Recall the notation $\mathcal{T}^k$ as the set of all rooted binary subtrees in the underlying \emph{complete binary tree} from Section~\ref{sec:TS-PCA-intro}. We write $\bm{v} \in \mathcal{T}^k$ if the support set of $\bm{v}$ satisfies $\mathsf{supp}(\bm{v}) \in \mathcal{T}^k$. Let
\begin{align}
    \widehat{\bm{v}}_{\mathsf{TS}} := \argmax_{\bm{v}} ~ \bm{v}^{\top} \widehat{\bm{\Sigma}} \bm{v} ~~\text{s.t.}~~ \bm{v} \in \mathcal{S}^{d - 1} \cap \mathcal{T}^k \label{eq:TS-PCA-est}
\end{align} 
denote the estimator obtained from exhaustive search. 

\begin{corollary}\label{coro:TS-PCA-fund-limits}
There exists a pair of positive constants $(c, C)$ such that the following holds. \\
\noindent (a) Without loss of generality, suppose $\langle \bm{v}_*, \widehat{\bm{v}}_{\mathsf{TS}} \rangle \geq 0$. Then for any $c_1 > 0$ and $\bm{v}_* \in \mathcal{S}^{d - 1} \cap \mathcal{T}^k$, we have
\[  \big\|\widehat{\bm{v}}_{\mathsf{TS}} - \bm{v}_*\big\|_2 \leq  C \left( \frac{1 + \lambda}{\lambda}\right) \cdot \sqrt{\frac{(3 + \ln 2 + c_1)k}{n}}
\]
with probability at least $1 - 2\exp(- c_1k)$. \\
\noindent (b) We have the minimax lower bound 
\begin{align*}
&\inf_{\widehat{\bm{v}}} \; \sup_{\bm{v}_* \in \mathcal{S}^{d - 1} \cap \mathcal{T}^k} \mathbb{E}\left[ \left\| \widehat{\bm{v}} \widehat{\bm{v}}^{\top} - \bm{v}_* \bm{v}_*^{\top}  \right\|_F \right] \geq c \cdot  
    \min \Bigg\{\frac{1}{4\sqrt{\log k}}, ~~ \frac{1}{4}\sqrt{\frac{1 + \lambda}{8 \lambda^2}} \sqrt{\frac{k/\log k}{n}} \Bigg\}.
\end{align*}
Here, the infimum is taken over all measurable functions of the observations $\{ \bm{x}_i \}_{i = 1}^n$ drawn i.i.d. from the distribution $\mathcal{D}(\lambda; \bm{v}_*)$.
\end{corollary}
Corollary~\ref{coro:TS-PCA-fund-limits} is proved in Section~\ref{proof-coro-TS-PCA-fund-limits}. The term
%
$\sqrt{k/n}$ arises from evaluating the cardinality of the set  $\mathcal{T}^k$ in tree-sparse PCA. In particular, 
we have $|\mathcal{T}^k| \leq (2e)^k / (k + 1)$  \citep{baraniuk2010model}, and taking logarithms results in a logarithmic factor gain over vanilla sparse PCA. Corollary~\ref{coro:TS-PCA-fund-limits}(b) provides a minimax lower bound of $\Omega(\sqrt{k / (n \log k)})$ for tree-sparse PCA, which has a logarithm gap $\sqrt{1/\log k}$ compared with the upper bound in Corollary~\ref{coro:TS-PCA-fund-limits}(a). This gap is small for small $k$, but we conjecture that it can be eliminated.

\begin{figure}[!h]
\centering
    \begin{subfigure}
        \centering
        \includegraphics[width=0.31\textwidth]{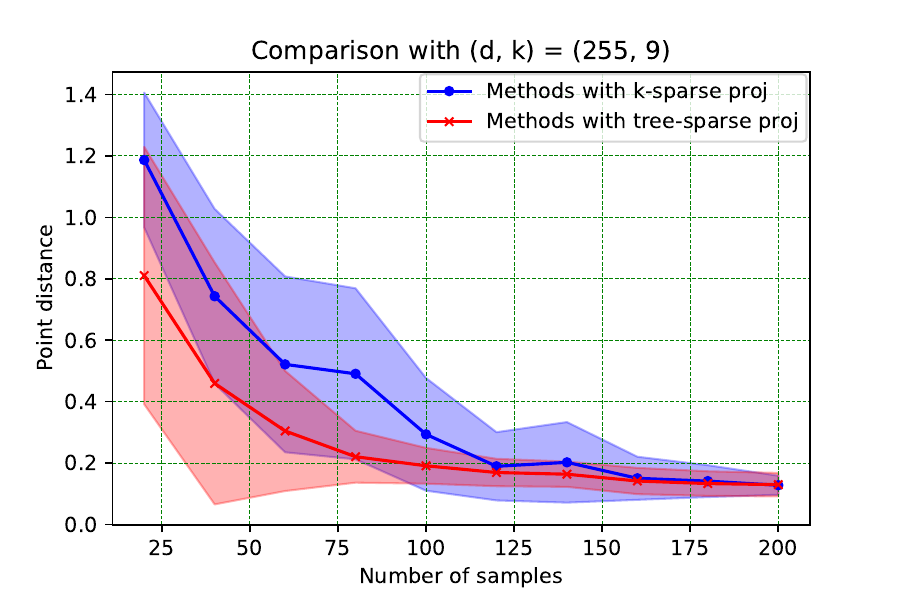}
    \end{subfigure}
    \hfill
    \begin{subfigure}
        \centering
        \includegraphics[width=0.31\textwidth]{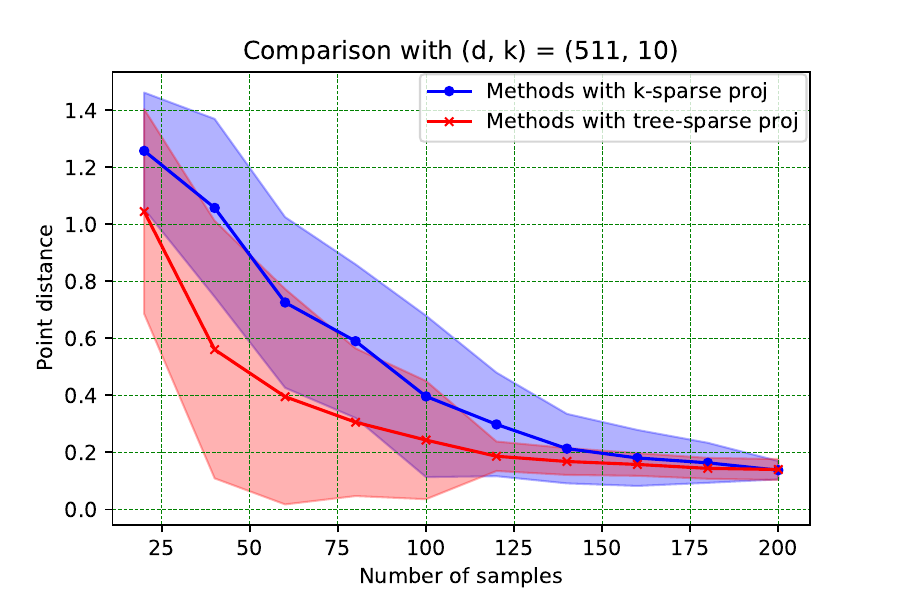}
    \end{subfigure}
    \hfill 
    \begin{subfigure}
        \centering
        \includegraphics[width=0.31\textwidth]{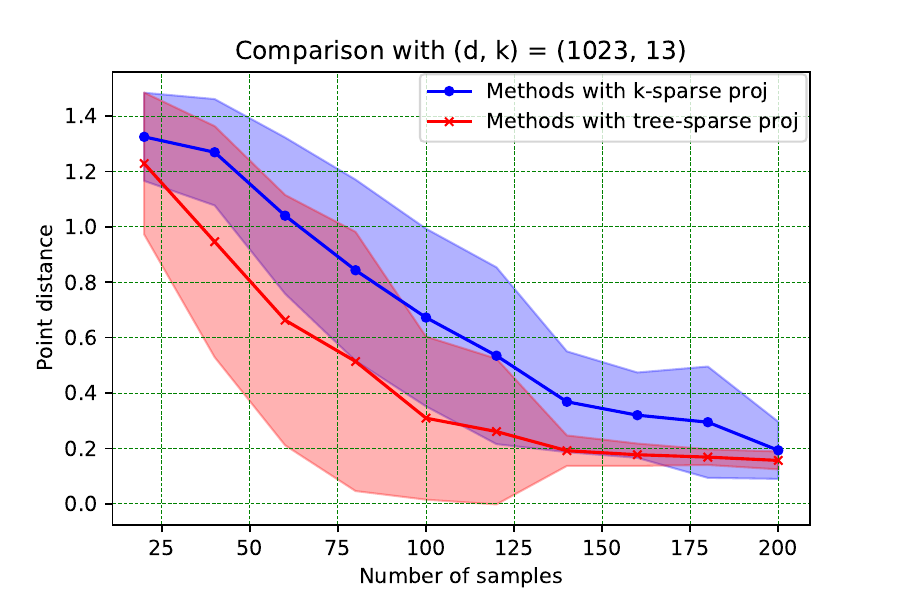}
    \end{subfigure}

    \begin{subfigure}
        \centering
        \includegraphics[width=0.31\textwidth]{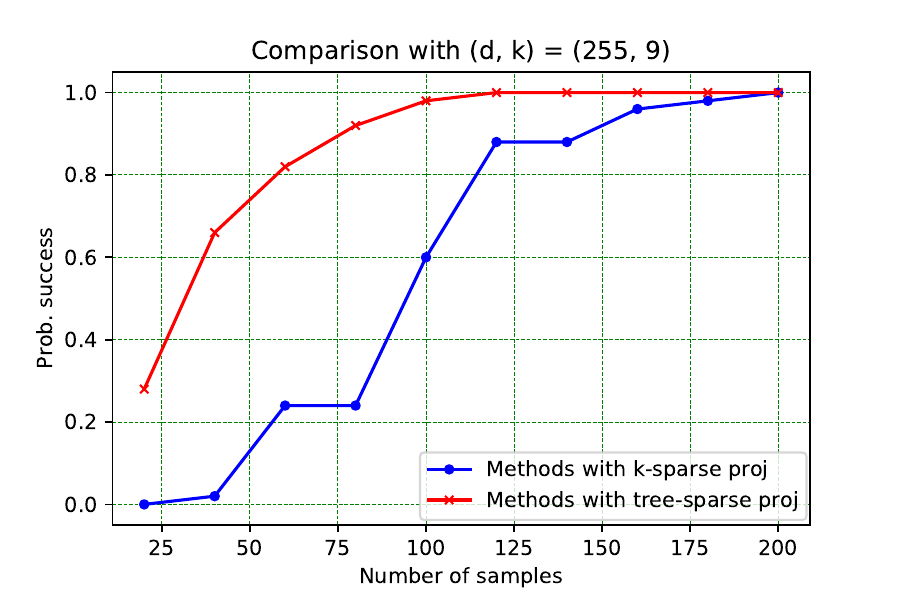}
    \end{subfigure}
    \hfill
    \begin{subfigure}
        \centering
        \includegraphics[width=0.31\textwidth]{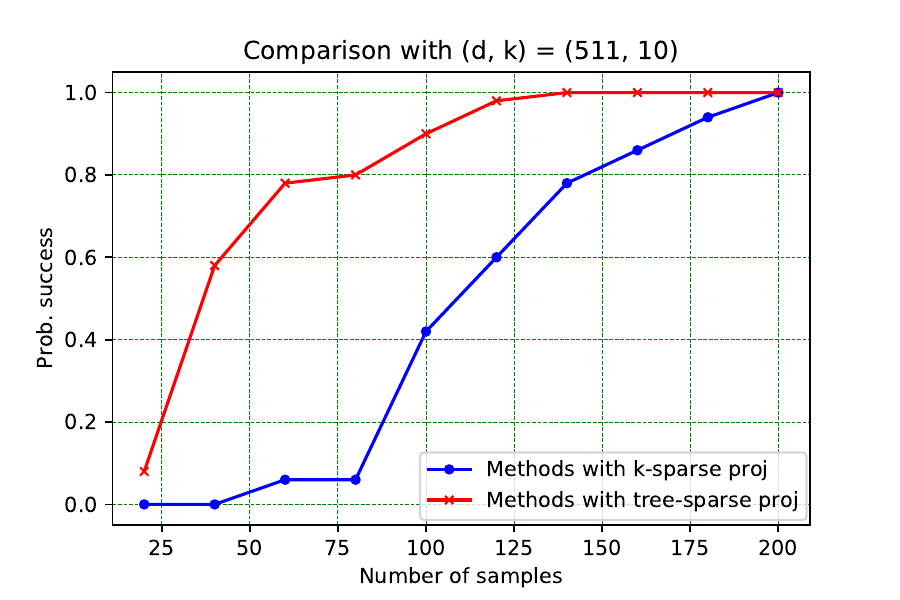}
    \end{subfigure}
    \hfill
    \begin{subfigure}
        \centering
        \includegraphics[width=0.31\textwidth]{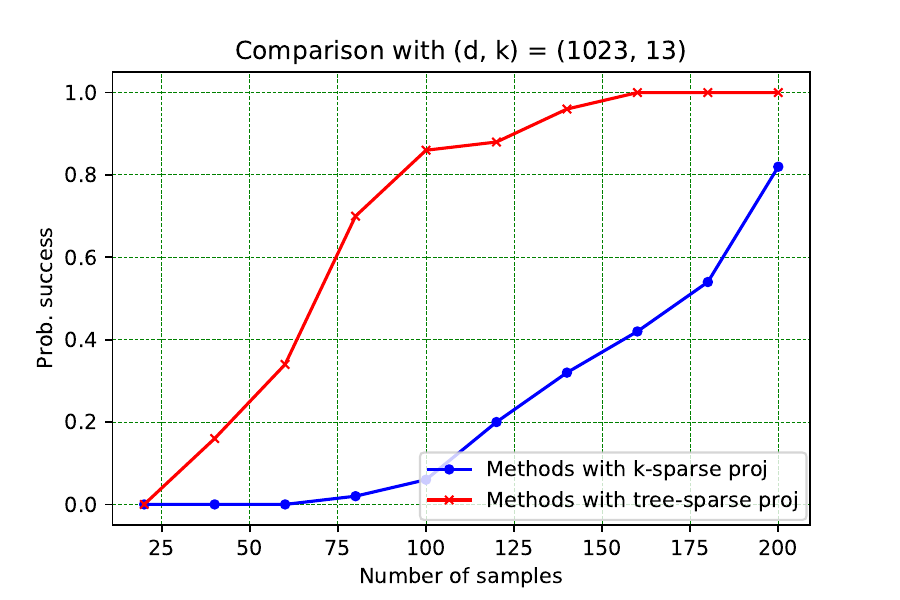}
\end{subfigure}

\caption{
Given the sample dimension $d = 2^L - 1$, sparsity $k$, and eigengap $\lambda$, we choose a particular tree sparsity support set $T_* \in \mathcal{T}^k$ and set the ground truth vector $\bm{v}_*$ as $ [\bm{v}_*]_i = \pm \frac{1}{\sqrt{k}}$ if $i \in T_*$ and $[\bm{v}_*]_i = 0$ if $i \notin T_*$. Given a tuple of $(\lambda,d,k,n)$, for each trial, we generate samples from the distribution $\mathcal{D}(\lambda, \bm{v}_*)$ based on the Wishart model in Section~\ref{sec:setting-background}, and we run Algorithm~\ref{alg:initialization} for initialization, and Algorithm~\ref{alg:PPM} with general $k$-sparse projection or with tree-sparse projection for local refinement. Each trial is repeated 50 times independently. We set $\lambda = 3$ and choose $(d,k) = (255,9),(511,10),(1023,13)$. For each choice of $(d,k)$, we simulate for each $n = \{20,40,\dots,200\}$.\\\\
In the first row, we plot the $\ell_2$ distance $\|\bm{v}_T - \bm{v}_*\|_2$ versus the number of samples $n$. The two curves in each panel correspond to the averaged values over 50 independent trials of the proposed methods with general $k$-sparse projection or with tree-sparse projection; the shaded parts represent the empirical standard deviations over 50 trials. As we can observe, using tree-sparse projection achieves smaller estimation error (for a given, small sample size) than using general $k$-sparse projection. \\\\
In the second row, we further plot of the success probability of support recovery of the methods using general $k$-sparse projection or using tree-sparse projection verse the number of samples $n$. The support of $\bm{v}_{*}$ is considered as successfully recovered if $\mathsf{supp}(\bm{v}_T) = T_{*}$. The success probability is then computed as the ratio of the number of trials that successfully recover the support over 50 independent trials. For a fixed small sample size, we observe that using tree-sparse projection achieves higher success probability of support recovery compared with using the vanilla $k$-sparse projection.}
\label{figure-tree-sparse} 
\end{figure}

\begin{remark} \label{remark:logd-saving-TSPCA}
Compared with the fundamental limits for vanilla sparse PCA, the upper bounds for tree-sparse PCA in Corollary~\ref{coro:TS-PCA-fund-limits} save a factor $\log d$, which parallels the 
model-based compressed sensing literature. The saving could be significant in practice when $d$ is large (see Figures~\ref{figure-tree-sparse} to follow)---indeed, this is one of the successes behind model-based compressive sensing. 
\end{remark}

\subsubsection{Local convergence and initialization} \label{sec:initialization-TS-PCA}


\paragraph{Exact projection oracle} We use the projection method proposed in \cite{cartis2013exact} as our tractable exact projection oracle $\Pi_{\mathcal{T}^k}$ for tree sparse PCA. This oracle has running time $O(kd)$. 
With our projection oracle in hand, we can now state our corollaries for the projected power method for tree sparse PCA.


\begin{corollary} \label{coro:TS-PCA-PPM}
 Suppose in Algorithm~\ref{alg:PPM} that the initialization $\bm{v}_{0} \in \mathcal{T}^{k} \cap \mathcal{S}^{d-1}$ satisfies $\langle \bm{v}_0,\bm{v}_* \rangle \geq 1/2$. There exists a tuple of universal positive constants $(c,C_1,C_2, C_3)$ such that for $\lambda \geq C_1$, $n\geq C_2k$ and all $t \geq 1$, the iterate $\bm{v}_t$ from Algorithm~\ref{alg:PPM-project} satisfies 
 \[ 
 \| \bm{v}_{t} - \bm{v}_{*} \|_2 \leq \frac{1}{2^t} \cdot \| \bm{v}_{0} - \bm{v}_{*} \|_2 + C_3 \sqrt{\frac{k}{n}},
 \]
 with probability at least $1-\exp(- ck)$. 
\end{corollary}
Corollary~\ref{coro:TS-PCA-PPM} is proved in Section~\ref{proof-coro-TS-PCA-PPM} from Theorem~\ref{thm:convergence}. 
We can also use the exact projection oracle $\Pi_{\mathcal{T}^k}$ to obtain the following corollary for our initialization method.

\begin{corollary} \label{coro:initial-TS-PCA}
Assume $k^2 \leq d / e$. There exists a pair of universal positive constants $(C, C')$ such that if  $n\geq \max\{C\log d,k^{2}\}$ and $n \geq C'\max \big\{1,\lambda^{-2}\big\} \log(d/k^2) k^2,$ then Algorithm~\ref{alg:initialization} returns an initial vector $\bm{v}_0 \in \mathcal{S}^{d - 1} \cap \mathcal{T}^k$ satisfying $\langle \bm{v}_0, \bm{v}_* \rangle \geq 1/2$ with probability $1 - C'\exp(- \min\{\sqrt{d}, n\}/C')$. 
\end{corollary} 
Like Corollary~\ref{coro:initial-PS-PCA}, it is straightforward to see that Corollary~\ref{coro:initial-TS-PCA} follows from Theorem~\ref{thm:initialization-method} by specifying $c_0 = 1/2$.

Corollary~\ref{coro:initial-TS-PCA} shows that provided $n = \Omega(k^{2})$, the output $\bm{v}_0 \in \mathcal{S}^{d - 1} \cap \mathcal{T}^k$ satisfies the initialization condition required for the subsequent projected power method to succeed. Putting these two results together, we have produced an end-to-end and computationally efficient algorithm that produces a statistically efficient solution provided $n = \Omega(k^{2})$. The next section is concerned with the question of whether the condition $n = \Omega(k^{2})$ is necessary for polynomial-time algorithms. 

\subsubsection{SDP Hardness for Tree Sparse PCA.} \label{sec:examples-SDP-hard}

To understand the aforementioned gap in sample size, we now provide a computational lower bound for a class of SDP solutions to tree-sparse PCA, showing that they require on the order of $k^2$ samples.

To make things formal, we consider the following subclass of tree sparse PCA problems: every entry of the $k$ tree-sparse ground truth unit vector $\bm{v}_*$ only takes one of the values $\{0, \pm k^{-1/2}\}$. With knowledge of this side information in addition to tree sparsity, the natural choice of exhaustive estimator is given by the maximizer of the following optimization problem:
\begin{align}\label{re-opt-tree-sparse}
	\begin{array}{rllll}
		\max_{\bm{v}} & \bm{v}^{\top} \widehat{\bm{\Sigma}} \bm{v} \\
		\text{s.t.} & \|\bm{v}\|_2^2 = 1,\;\|\bm{v}\|_0 = k \\
		& \bm{v}(i)^2 \leq \bm{v}(\lfloor i / 2 \rfloor)^2\;\text{ for all } 2\leq i \leq d. 
	\end{array}	
\end{align}
The natural semidefinite programming (SDP) relaxation of the program~\eqref{re-opt-tree-sparse} is then given by
\begin{align}\label{SDP-tree-sparse-PCA}
	\begin{array}{rllll}
		\mathsf{SDP}(\widehat{\bm{\Sigma}}) = \max_{\bm{M} \in \mathbb{R}^{d \times d}} & \sum_{i=1}^{d} \sum_{j=1}^{d} \widehat{\bm{\Sigma}}_{ij} \bm{M}_{ij} \\
		\text{s.t.} & \sum_{i=1}^{d} \bm{M}_{ii}^{2} = 1 \\
		& \sum_{i=1}^{d}\sum_{j=1}^{d} |\bm{M}_{ij}| \leq k \\
		& \bm{M} \succeq \bm{0}_{d \times d} \\
		& \bm{M}_{ii} \leq \bm{M}_{\lfloor i / 2 \rfloor\lfloor i / 2 \rfloor} \text{ for all }\; 2\leq i\leq d.
	\end{array}
\end{align}

It is well-known that for vanilla sparse PCA, the SDP attains the best-known sample complexity among all polynomial time algorithms. Proving a lower bound for this class of algorithms is thus powerful---when this subclass of low-degree estimators fails at the indicated threshold, it suggests a natural hardness result.

\begin{proposition}\label{prop:TS-SDP-hard}
    Suppose data $\bm{X}$ are drawn from the distribution $\mathcal{D}(\lambda; \bm{v}_*)$ with ground truth $\bm{v}_*$ given by a $k$ tree-sparse unit vector with every entry of taking one of the values in the set $\{0, \pm k^{-1/2}\}$.
    There exists a tuple of universal positive constants $(c,c_1,C,C_1)$ such that for $c_1d \leq n \leq C_1 d,\; n \leq ck^{2}$ and $1 \leq \lambda \leq \frac{d}{Cn}$, the optimal solution $\bm{M}_*$ of the SDP relaxation~\eqref{SDP-tree-sparse-PCA} satisfies $\big\| \bm{M}_* - \bm{v}_{*}\bm{v}_{*}^{\top}\big\|_{2} \geq \frac{1}{5}$ with probability at least $1- \tilde{c} d^{-\tilde{c}}$ for some constant $\tilde{c} \geq 1$.
\end{proposition}

In words, Proposition~\ref{prop:TS-SDP-hard} shows that unless the number of samples satisfies $n \geq C' k^2$ for some positive constant $C' \geq c$, the optimal solution $\bm{M}_*$ of the SDP relaxation~\eqref{SDP-tree-sparse-PCA} fails to estimate the ground truth consistently, even with the side information that its entries take only one of three values. The proof of Proposition~\ref{prop:TS-SDP-hard} can be found in Section~\ref{proof-prop-TS-SDP-hard}, and builds on the techniques proposed in [Section 4, \cite{ma2015sum}].

%% file: discussion.tex
\section{Discussion}
We studied the local convergence properties of the projected power method in a general class of structured PCA problems. We also established the fundamental limits of estimation in this family of problems, and studied a general family of initialization methods.
Our work generalizes these statistical and algorithmic results from vanilla sparse PCA to this more general class of models. 
We specialized our results to two commonly used notions of structure---given by tree and path sparsity---showing end-to-end estimation algorithms accompanied by evidence of computational hardness.

Let us close with some potential questions for future investigation. The first is to generalize these 
results to other forms of structured PCA
\citep{yi2020non}. Another natural direction is consider more than a single principal component. Progress has been made towards establishing general fundamental limits in these settings~\citep{cai2021optimal}; there are also natural analogs for the projected power method in these settings and it would be interesting to analyze it under a general structural assumption along with statistical and computational limits.

%% file: appendix.tex

\section{Appendix} 

Before proceeding to proofs of our main results, we present some calculations that were alluded to in the main text.

\subsection{Time-Consuming Case in Section~\ref{sec:setting-background}} \label{app:time-consuming-case}

Consider the following time-consuming case. 

\begin{example} \label{exmp:time-consuming-case}
\textbf{Time-Consuming Case.}
Given $\mathcal{L} = \{L_1, \ldots, L_{d - 1}, L_d\}$ with $L_i = \text{span}(\phi_i, \phi_{i + 1})$ for $i = 1, \ldots, d - 1$ and $L_d = \text{span}(\phi_1, \phi_d)$. Each linear subspace $L_i$ is known by given two linearly independent but not necessarily orthonormal vectors, say $\bm{u}^{(i)}_1, \bm{u}^{(i)}_2$, in $L_i$. As a result, for a given linear subspace $L$, we do not know the index $i \in [d]$ of this linear subspace $L$ based on the given vectors $\bm{u}^{(\cdot)}_1, \bm{u}^{(\cdot)}_2 \in L$. Hence the corresponding two bases that spans this known linear subspace $L$ is unknown to us.
\end{example}

From the Example~\ref{exmp:time-consuming-case}, if two linear subspaces $L, L'$ have a non-zero intersection, i.e., $L \cap L' \neq \{\bm{0}\}$, then the base $\phi = L \cap L' \in \mathcal{B}$ is uniquely determined, and so as the rest two bases in $L, L'$ respectively. Thus computing $\bm{\Phi}$ from $\mathcal{L} := \{L_1, \ldots, L_{d - 1}, L_d\}$ is equivalent to find out all bases $\phi \in \mathcal{B}$ via intersection verification. Since we do not know the index corresponding to each linear subspace, to compute one base in $\mathcal{B}$, what we can do is to verify the intersection of two randomly chosen linear subspaces. The detailed procedures of computing the unknown orthonormal basis $\mathcal{B}$ are presented in the randomized algorithm~\ref{alg:IV}. 

\begin{algorithm}
\caption{Intersection Verification for $\mathcal{B}$ or $\bm{\Phi}$}
\label{alg:IV}
\textbf{Input.} $\mathcal{L}$, each linear subspace $L \in \mathcal{L}$ is represented by $\text{dim}(L)$ independent vectors in $L$.
\begin{algorithmic}[1]
\State \textbf{Initialize} $\mathcal{L}^{(0)} := \emptyset, \mathcal{B}^{(0)} := \emptyset, t = 0$.
\While{$|\mathcal{B}^{(t)}| < d$} \hfill \textbf{outer while-loop} \label{alg:IV-outer}
\State Pick a linear subspace $L^{(t)} \in \mathcal{L} \backslash \mathcal{L}^{(t)}$.
\While{True} \hfill \textbf{inner while-loop} \label{alg:IV-inner} 
\State Select $\tilde{L}^{(t)} \in \mathcal{L} \backslash \{L^{(t)}\}$ uniformly at random without replacement. \label{alg:IV-selection}
\If{$L^{(t)} \cap \tilde{L}^{(t)} \neq \{\bm{0}\}$}
\State Compute three bases for $L^{(t)}, \tilde{L}^{(t)}$. 
\State Update $\mathcal{B}^{(t + 1)}$ via adding the above three bases. \label{alg:IV-basis}
\State Update $\mathcal{L}^{(t + 1)} := \mathcal{L}^{(t)} \cup \{L^{(t)}, \tilde{L}^{(t)}\}$.
\State \textbf{Break} inner while-loop.
\EndIf
\EndWhile
\EndWhile
\end{algorithmic}
\textbf{Output.} $\mathcal{B}^{(t + 1)}$ or $\bm{\Phi}$ with columns all bases in $\mathcal{B}^{(t + 1)}$.  
\end{algorithm} 

\begin{proposition} \label{prop:ERT-IV}
\textbf{Expected Running Time of Algorithm~\ref{alg:IV}.} 
Under the setting of $\mathcal{L}$ presented in Example~\ref{exmp:time-consuming-case}, the expected running time of Algorithm~\ref{alg:IV} is of order $O(d^3)$. 
\end{proposition}

The proof of the Proposition~\ref{prop:ERT-IV} is presented later in this Section. As a conclusion, finding all bases takes more than $d^2 / 9$ intersection verifications in expectation. Each intersection verification requires $O(d)$ time. Then the expected running time of computing $\bm{\Phi}$ is $O(d^3)$. 
In contrast, computing the exact projection of $\bm{v}$ onto $\mathcal{M}$ takes $O(d^2)$ running time\footnote{Projecting onto a 1D linear subspace takes $O(d)$ time, and there are $d$ linear subspaces in total.}. Therefore, the above analysis illustrates that extracting $\bm{\Phi}$ takes way more time than just implementing the projection, which further explains why $\bm{\Phi}$ is not necessary to recover the true PC $\bm{v}_*$. \\

Moreover, under the general setting of $\mathcal{L} = \{L_1, \ldots, L_M\}$, given a set of independent and not necessarily orthonormal vectors $\bm{u}_1^{(m)}, \ldots, \bm{u}_{\text{dim}(L_m)}^{(m)}$ of each linear subspace $L_m$ with $m \in [M]$, it is unclear whether and how long one could find the orthonormal basis $\bm{\Phi}$ from $\mathcal{L}$ via solving the following variant of dictionary learning problem~\eqref{eq:dictionary}, 
\begin{align}
    \begin{array}{rlll}
        \min_{\bm{\Phi}, \bm{R}} & \left\| [\bm{U}^{(1)} ~|~ \cdots ~|~ \bm{U}^{(M)}] - \bm{\Phi} [\bm{R}^{(1)} ~|~ \cdots ~|~ \bm{R}^{(M)}] \right\|_F^2 \\
        \text{s.t.} & \bm{\Phi} \bm{\Phi}^{\top} = \bm{I}_d, ~~ \| \bm{R}^{(m)} \|_0 \leq \text{dim}(L_m) ~~ \forall ~ m \in M
    \end{array}, \label{eq:dictionary}
\end{align}
where, for all $m \in [M]$, $\bm{U}^{(m)}$ denotes the matrix with columns $\bm{u}_1^{(m)}, \ldots, \bm{u}_{\text{dim}(L_m)}^{(m)}$ and $\| \bm{R}^{(m)} \|_0 \leq \text{dim}(L_m)$ denotes that the number of non-zero rows of $\bm{R}^{(m)}$ is at most $\text{dim}(L_m)$.



\begin{proof}
\textbf{Proof of Proposition~\ref{prop:ERT-IV}.} 
First, based on the setting of each $L_i$ for $i = 1, \ldots, d$, $L_i$ has non-zero intersection with $L_{i - 1}$ and $L_{i + 1}$. Thus the expected number of selections (i.e., inner while-loop~\eqref{alg:IV-inner}) for step~\eqref{alg:IV-selection} of Algorithm~\ref{alg:IV} satisfies 
\begin{align*}
    \mathbb{E}[\textup{number of selections}] = & ~ 1 \cdot \frac{2}{d - 1} + 2 \cdot \frac{d - 3}{d - 1} \frac{2}{d - 2} + 3 \cdot \frac{d - 3}{d - 1} \frac{d - 4}{d - 2} \frac{2}{d - 3} + \cdots \\
    = & ~ \sum_{i = 1}^{d - 2} i \cdot \frac{2 (d - i - 1)}{(d - 1)(d - 2)} \\
    = & ~ \frac{d}{3}. 
\end{align*}
Every time we find $L^{(t)} \cap \tilde{L}^{(t)} \neq \{\bm{0}\}$, in step~\eqref{alg:IV-basis} of Algorithm~\ref{alg:IV}, we can add three more new bases to $\mathcal{B}^{(t + 1)}$ if $\tilde{L}^{(t)} \notin \mathcal{L}^{(t)}$, and one more new base to $\mathcal{B}^{(t + 1)}$ if $\tilde{L}^{(t)} \in \mathcal{L}^{(t)}$. Therefore, the number of outer while-loop~\eqref{alg:IV-outer} of Algorithm~\ref{alg:IV} satisfies 
\begin{align*}
    & ~ \text{number of outer while-loop} = \text{selection with 3 more bases} + \text{selection with 1 more bases}.
\end{align*}
Moreover, the stopping criteria of outer while-loop~\eqref{alg:IV-outer} of Algorithm~\ref{alg:IV} ensures that the number of outer while-loop~\eqref{alg:IV-outer} is greater than or equal to $d/3$, where the equality holds when we can add three more new bases at every inner while-loop~\eqref{alg:IV-inner} of Algorithm~\ref{alg:IV}. Therefore, in expectation, the total number of selections of Algorithm~\ref{alg:IV} satisfies 
\begin{align*}
    \mathbb{E}[\text{total number of selections}] 
    = & ~ \text{number of outer while-loop} \times \mathbb{E}[\textup{number of selections}]\\
    \geq & ~ d^2 / 9.
\end{align*}
Since we do an intersection verification for each selection, and an intersection verification takes $O(d)$ running time, then the expected total running time of computing $\bm{\Phi}$ is $O(d^3)$. 
\end{proof}




\input{fundamental-limits-proof}

\input{local-convergence-proof}

\subsection{Proofs of Theorem~\ref{thm:initialization-method}} \label{app:initialization}
We first restate the following two existing results in \cite{deshpande2016sparse} that will be used in Theorem~\ref{thm:initialization-method}. 
\begin{proposition} \label{prop:cov-threshold-2}
\textbf{Restated, [Theorem 2, \cite{deshpande2016sparse}].} There exists a numerical constant $C_3 > 0$ such that the following holds. Assume $n > \max\{ C \log d, k^2 \}$ and $k^2 \leq d / e$. Set $\tau$ according to equation~\eqref{initia-threshold-tau}. Then with probability $1 - o(1)$ 
\begin{align*}
    \|\widehat{\bm{v}}_{\textup{soft}} - \bm{v}_*\|_2^2 \leq \frac{C_3}{\lambda^2} \frac{k^2 \max\{\lambda^2, 1\}}{n} \log (d / k^2) =: 2 \cdot \textup{dis}_{\textup{soft}}(n). 
\end{align*}
\end{proposition}

We note that it follows from \textbf{Remark 6.6} of \cite{deshpande2016sparse} that the probability $1-o(1)$ is of the form $1-C'\exp(-\min\{\sqrt{d},n\}/C')$ for every $n$ large enough. Based on the above proposition, we now present the proof of Theorem~\ref{thm:initialization-method}.

\paragraph{Proof of Theorem~\ref{thm:initialization-method}.}
Consider the following two orthogonal decompositions of $\widehat{\bm{v}}_{\textup{soft}}$ in Algorithm~\ref{alg:initialization}, 
\begin{align*}
    & ~ \widehat{\bm{v}}_{\textup{soft}} = \alpha_{\textup{soft}} \bm{v}_* + \beta_{\textup{soft}} \bm{v}_{\bot} \\
    & ~ \alpha_{\textup{soft}}^2 + \beta_{\textup{soft}}^2 = 1, ~ \langle \bm{v}_*, ~\bm{v}_{\bot} \rangle = 0, ~ \|\bm{v}_*\|_2 = \|\bm{v}_{\bot}\|_2 = 1, \\
    & ~ \widehat{\bm{v}}_{\textup{soft}} = \alpha_{\textup{soft}}' \bm{v}_0 +  \beta_{\textup{soft}}' \bm{v}_{\bot}' \\
    & ~ (\alpha_{\textup{soft}}')^2 + (\beta_{\textup{soft}}')^2 = 1, ~ \langle \bm{v}_0, ~\bm{v}_{\bot}' \rangle = 0, ~ \|\bm{v}_0\|_2 = \|\bm{v}_{\bot}'\|_2 = 1.
\end{align*}
Based on Proposition~\ref{prop:cov-threshold-2}, the first orthogonal decomposition decomposes $\widehat{\bm{v}}_{\textup{soft}}$ along the direction of $\bm{v}_*$ such that $\alpha_{\textup{soft}} \in [1 - \textup{dis}_{\textup{soft}}(n), ~ 1]$. In contrast, the second orthogonal decomposition decomposes $\widehat{\bm{v}}_{\textup{soft}}$ along the direction of $\bm{v}_0$. Since $\bm{v}_* \in \mathcal{S}^{d - 1} \cap \mathcal{M}$ and $\bm{v}_0 = \Pi_{\mathcal{S}^{d - 1} \cap \mathcal{M}}(\widehat{\bm{v}}_{\textup{soft}}) \in \mathcal{S}^{d - 1} \cap \mathcal{M}$, then we have $\alpha_{\textup{soft}}' \geq \alpha_{\textup{soft}}$. Using $\|\widehat{\bm{v}}_{\textup{soft}}\|_2 = 1$, we obtain that
\begin{align*}
    1 = & ~ \langle \widehat{\bm{v}}_{\textup{soft}}, ~  \widehat{\bm{v}}_{\textup{soft}} \rangle \\
    = & ~ \langle \alpha_{\textup{soft}} \bm{v}_* + \beta_{\textup{soft}} \bm{v}_{\bot}, ~  \alpha_{\textup{soft}}' \bm{v}_0 +  \beta_{\textup{soft}}' \bm{v}_{\bot}' \rangle \\
    = & ~ \alpha_{\textup{soft}} \alpha_{\textup{soft}}' \langle \bm{v}_*, ~ \bm{v}_0 \rangle + \alpha_{\textup{soft}} \beta_{\textup{soft}}' \langle \bm{v}_*, ~ \bm{v}_{\bot}' \rangle + \beta_{\textup{soft}} \alpha_{\textup{soft}}' \langle \bm{v}_{\bot}, ~ \bm{v}_{0}' \rangle + \beta_{\textup{soft}} \beta_{\textup{soft}}' \langle \bm{v}_{\bot}, ~ \bm{v}_{\bot}' \rangle. 
\end{align*}
Continuing, we obtain that
\begin{align*}
    \langle \bm{v}_0, ~ \bm{v}_* \rangle 
    = & ~ \frac{1}{\alpha_{\textup{soft}} \alpha_{\textup{soft}}'} \big( 1 - \alpha_{\textup{soft}} \beta_{\textup{soft}}' \langle \bm{v}_*, ~ \bm{v}_{\bot}' \rangle \\
    & ~ - \beta_{\textup{soft}} \alpha_{\textup{soft}}' \langle \bm{v}_{\bot}, ~ \bm{v}_{0}' \rangle - \beta_{\textup{soft}} \beta_{\textup{soft}}' \langle \bm{v}_{\bot}, ~ \bm{v}_{\bot}' \rangle \big) \\
    = & ~ \frac{1}{\alpha_{\textup{soft}} \alpha_{\textup{soft}}'} - \frac{\beta_{\textup{soft}}'}{\alpha_{\textup{soft}}'} \langle \bm{v}_*, ~ \bm{v}_{\bot}' \rangle - \frac{\beta_{\textup{soft}}}{\alpha_{\textup{soft}}} \langle \bm{v}_{\bot}, ~ \bm{v}_{0}' \rangle \\
    & ~  - \frac{\beta_{\textup{soft}} \beta_{\textup{soft}}'}{\alpha_{\textup{soft}} \alpha_{\textup{soft}}'} \langle \bm{v}_{\bot}, ~ \bm{v}_{\bot}' \rangle \\
    \geq & ~ 1 - \frac{\beta_{\textup{soft}}'}{\alpha_{\textup{soft}}'} - \frac{\beta_{\textup{soft}}}{\alpha_{\textup{soft}}} - \frac{\beta_{\textup{soft}} \beta_{\textup{soft}}'}{\alpha_{\textup{soft}} \alpha_{\textup{soft}}'}. 
\end{align*}
It is sufficient to have
\begin{align*}
    1 - \frac{\beta_{\textup{soft}}'}{\alpha_{\textup{soft}}'} - \frac{\beta_{\textup{soft}}}{\alpha_{\textup{soft}}} - \frac{\beta_{\textup{soft}} \beta_{\textup{soft}}'}{\alpha_{\textup{soft}} \alpha_{\textup{soft}}'} \geq & ~ 1 - 3 \frac{\beta_{\textup{soft}}}{\alpha_{\textup{soft}}} \\
    \geq & ~ 1 - 3 \frac{\sqrt{2 \textup{dis}_{\textup{soft}}(n) - \textup{dis}_{\textup{soft}}^2(n)}}{1 - \textup{dis}_{\textup{soft}}(n)} \\
    \geq & ~ c_{0}.
\end{align*}
Therefore, by setting 
\begin{align*}
    n \geq \frac{18 C_4 \max\{\lambda^2, ~ 1\} k^2}{2(1 - c_0)^2 \lambda^2} \log(d/k^2),  
\end{align*} 
and applying Proposition~\ref{prop:cov-threshold-2}, the above inequality holds, thus the desired result follows immmediately.

\input{corollaries-proof}

%% file: fundamental-limits-proof.tex
\subsection{Proof of Theorem~\ref{thm:fund-limits}}
\subsubsection{Proof of Theorem~\ref{thm:fund-limits} part (a)}\label{app:fund-limits-up}
We first state the following auxilary lemma.
\begin{lemma} \label{lemma:curvature-of-matrix}
[Lemma 3.2.1, \cite{vu2012minimax}] Given any $\bm{u} \in \mathcal{S}^{d - 1}$. If $\bm{\Sigma} \succeq \bm{0}_{d \times d}$ has a unique largest eigenvalue $\lambda_1$ with a corresponding eigenvector $\bm{v}_*$, then 
\begin{align}
    \frac{1}{2} (\lambda_1 - \lambda_2) \|\bm{v}_*\bm{v}_*^{\top} - \bm{u} \bm{u}^{\top}\|_F^2 \leq \langle \bm{\Sigma}, \bm{v}_*\bm{v}_*^{\top} - \bm{u} \bm{u}^{\top} \rangle.  \notag
\end{align}
\end{lemma}
We defer the proof of Lemma~\ref{lemma:curvature-of-matrix} to the end. Now we are poised to prove part (a) of Theorem~\ref{thm:fund-limits}.
\begin{proof}
Note that $\bm{v}_*, \ESest \in \mathcal{M} \cap \mathcal{S}^{d - 1}$. Applying Lemma~\ref{lemma:curvature-of-matrix} yields  
\begin{align*}
    \frac{1}{2} (1 + \lambda - 1) \|\bm{v}_*\bm{v}_*^{\top} - \ESest \ESest^{\top}\|_F^2 
    \leq & ~ \langle \bm{\Sigma}, \bm{v}_*\bm{v}_*^{\top} - \ESest \ESest^{\top} \rangle\\
    = & ~\langle \widehat{\bm{\Sigma}}, \bm{v}_*\bm{v}_*^{\top} \rangle - \langle \bm{\Sigma}, \ESest \ESest^{\top} \rangle - \langle \widehat{\bm{\Sigma}} - \bm{\Sigma}, \bm{v}_*\bm{v}_*^{\top} \rangle \\
    \leq & ~ \langle \widehat{\bm{\Sigma}}, \ESest \ESest^{\top} \rangle - \langle \bm{\Sigma}, \ESest \ESest^{\top} \rangle - \langle \widehat{\bm{\Sigma}} - \bm{\Sigma}, \bm{v}_*\bm{v}_*^{\top} \rangle \\
    = & ~ \langle \widehat{\bm{\Sigma}} - \bm{\Sigma}, \ESest \ESest^{\top} - \bm{v}_*\bm{v}_*^{\top} \rangle  \\
    = & ~ \langle \bm{W}, \ESest \ESest^{\top} - \bm{v}_*\bm{v}_*^{\top} \rangle.
\end{align*}
Let $\widehat{L} \in \mathcal{L}$ such that contains $\ESest \in \widehat{L} $ and  recall that $\bm{v}_* \in L_* \in \mathcal{L}$. Let $\widehat{F} := \mathsf{conv}(\widehat{L} \cup L_*) \subseteq \mathbb{R}^d$. Let $\bm{P}_{\widehat{F}}$ be the projection matrix with respect to the linear subspace $\widehat{F}$. Continuing, we obtain that
\begin{align*}
    \left| \langle \bm{W}, \ESest \ESest^{\top} - \bm{v}_*\bm{v}_*^{\top} \rangle \right| \leq &  \left| \langle \bm{W}, \bm{P}_{\widehat{F}}( \ESest \ESest^{\top} - \bm{v}_*\bm{v}_*^{\top}) \bm{P}_{\widehat{F}} \rangle \right| \\
    = & \left| \langle \bm{W}_{\widehat{F}},  \ESest \ESest^{\top} - \bm{v}_*\bm{v}_*^{\top} \rangle \right| \\
    \overset{(1)}{\leq} &  \|\bm{W}_{\widehat{F}}\|_{\text{sp}} \cdot \|\bm{v}_*\bm{v}_*^{\top} - \ESest \ESest^{\top}\|_{S_1} \\
    \overset{(2)}{\leq} &  \|\bm{W}_{\widehat{F}}\|_{\text{sp}} \cdot \sqrt{2} \|\bm{v}_*\bm{v}_*^{\top} - \ESest \ESest^{\top}\|_{F}  \notag\\
    = & \sqrt{2} \rho(\bm{W}, \widehat{F}) \cdot  \|\bm{v}_*\bm{v}_*^{\top} - \ESest \ESest^{\top}\|_{F},  \notag
\end{align*}
where $\|\cdot\|_{\text{sp}}$ denotes the spectral norm, i.e., the largest singular value, $\|\cdot\|_{S_1}$ denotes the Schatten-1-norm, i.e., the sum of singular values, step $(1)$ follows from Von Neumann’s trace inequality and step $(2)$ follows since $\|\bm{X}\|_{S_1} \leq \sqrt{\text{rank}(\bm{X})} \|\bm{X}\|_{F}$ for any symmetric matrix $\bm{X} \in \mathbb{R}^{d \times d}$. Now putting the two pieces together yields 
\begin{align*}
    \|\bm{v}_*\bm{v}_*^{\top} - \ESest \ESest^{\top}\|_F \leq \frac{2\sqrt{2}}{\lambda} \rho(\bm{W}, \widehat{F}).
\end{align*}
Without loss of generality, we assume $\langle \bm{v}_*, \ESest \rangle \geq 0$. Note that
\begin{align*}
	\|\bm{v}_*\bm{v}_*^{\top} - \ESest \ESest^{\top}\|_F^2 = & ~ 2 - 2 \langle \bm{v}_*, \ESest \rangle^2 \\
	 = & ~ (1 + \langle \bm{v}_*, \ESest \rangle) \|\bm{v}_* - \ESest\|_2^2 \\
	 \geq & ~ \|\bm{v}_* - \ESest\|_2^2.
\end{align*}
 Consequently, we obtain
 \[
 	\|\bm{v}_* - \ESest\|_2 \leq  \|\bm{v}_*\bm{v}_*^{\top} - \ESest \ESest^{\top}\|_F \leq \frac{2\sqrt{2}}{\lambda} \rho(\bm{W}, \widehat{F}).
 \]
 This concludes the proof.
\end{proof}

\paragraph{Proof of Lemma~\ref{lemma:curvature-of-matrix}}
The proof follows the same steps as presented in Lemma 3.2.1 in \cite{vu2012minimax}. We obtain that
\begin{align*}
    \langle \bm{\Sigma}, \bm{v}_*\bm{v}_*^{\top} - \bm{u} \bm{u}^{\top} \rangle 
    = & ~ \text{tr}(\bm{\Sigma}\bm{v}_*\bm{v}_*^{\top}) - \text{tr}(\bm{\Sigma}\bm{u} \bm{u}^{\top}) \notag\\
    = & ~ \text{tr}(\bm{\Sigma}(\bm{I}_d - \bm{u}\bm{u}^{\top}) \bm{v}_*\bm{v}_*^{\top}) - \text{tr}(\bm{\Sigma}\bm{u} \bm{u}^{\top}(\bm{I}_d - \bm{v}_*\bm{v}_*^{\top})).
\end{align*}
Using $\bm{\Sigma}\bm{v}_* = \lambda_{1} \bm{v}_{*}$, we obtain that
\begin{align*}
    \text{tr}(\bm{\Sigma}(\bm{I}_d - \bm{u}\bm{u}^{\top}) \bm{v}_*\bm{v}_*^{\top}) = & \text{tr}(\bm{v}_*\bm{v}_*^{\top} \bm{\Sigma}(\bm{I}_d - \bm{u}\bm{u}^{\top}) \bm{v}_*\bm{v}_*^{\top})  \\
    = & \lambda_1 \text{tr}(\bm{v}_*\bm{v}_*^{\top} (\bm{I}_d - \bm{u}\bm{u}^{\top}) \bm{v}_*\bm{v}_*^{\top})  \\
    = & \lambda_1 \text{tr}(\bm{v}_*\bm{v}_*^{\top} (\bm{I}_d - \bm{u}\bm{u}^{\top})^2 \bm{v}_*\bm{v}_*^{\top}) \\
    = & \lambda_1 \big\| \bm{v}_*\bm{v}_*^{\top} (\bm{I}_d - \bm{u}\bm{u}^{\top}) \big\|_F^2. 
\end{align*}
Using $(\bm{I}_d - \bm{v}_*\bm{v}_*^{\top}) \bm{\Sigma} \bm{v}_* = \lambda_1(\bm{I}_d - \bm{v}_*\bm{v}_*^{\top}) \bm{v}_* = \bm{0}$, we obtain that 
\begin{align*}
    & ~ \text{tr}(\bm{\Sigma}\bm{u} \bm{u}^{\top}(\bm{I}_d - \bm{v}_*\bm{v}_*^{\top})) \\
    = & ~ \bm{u}^{\top}(\bm{I}_d - \bm{v}_*\bm{v}_*^{\top})\bm{\Sigma} (\bm{I}_d - \bm{v}_*\bm{v}_*^{\top}) \bm{u} + \bm{u}^{\top}(\bm{I}_d - \bm{v}_*\bm{v}_*^{\top})\bm{\Sigma} \bm{v}_*\bm{v}_*^{\top} \bm{u}\\
    = & ~ \bm{u}^{\top}(\bm{I}_d - \bm{v}_*\bm{v}_*^{\top}) \bm{\Sigma}(\bm{I}_d - \bm{v}_*\bm{v}_*^{\top})\bm{u} \\
    \leq & ~ \lambda_2 \cdot \bm{u}^{\top}(\bm{I}_d - \bm{v}_*\bm{v}_*^{\top})^2 \bm{u}\\
    = & ~ \lambda_2 \cdot \big\| \bm{u}\bm{u}^{\top}(\bm{I}_d - \bm{v}_*\bm{v}_*^{\top})\big\|_F^2 \\
    = & ~ \lambda_2 \cdot \big\| \bm{v}_*\bm{v}_*^{\top} (\bm{I}_d - \bm{u}\bm{u}^{\top}) \big\|_F^2. 
\end{align*}
Putting the pieces together yields
\begin{align*}
    \langle \bm{\Sigma}, \bm{v}_*\bm{v}_*^{\top} - \bm{u} \bm{u}^{\top} \rangle \geq & ~  (\lambda_1 - \lambda_2) \big\| \bm{v}_*\bm{v}_*^{\top} (\bm{I}_d - \bm{u}\bm{u}^{\top}) \big\|_F^2\\ 
    = & ~ \frac{1}{2} (\lambda_1 - \lambda_2) \|\bm{v}_*\bm{v}_*^{\top} - \bm{u} \bm{u}^{\top}\|_F^2.
\end{align*}
This concludes the proof.

\subsubsection{Proof of Theorem~\ref{thm:fund-limits} part (b)}\label{app:fund-limits-lb}
We require the following version of Fano Method, which is also stated in [\cite{yu1997assouad}, Lemma 3].
\begin{proposition} \label{prop:g-fano-method}
\textbf{Generalized Fano Method.} Let $\mathcal{V}_{\epsilon} := \{ \bm{v}_1, \ldots, \bm{v}_{|\mathcal{V}_{\epsilon}|} \} \subseteq \mathcal{V}$ be a finite set. Each $\bm{v}_i$ for $i \in [|\mathcal{V}_{\epsilon}|]$ yields a probability measure $\mathcal{D}^{(n)}(\lambda; \bm{v}_i)$ on the $n$ samples. Let $\textrm{d}: \mathcal{V} \times \mathcal{V} \mapsto \mathbb{R}_+$ be a pseudometric on $\mathcal{V}$, $\mathrm{KL}: \mathcal{D} \times \mathcal{D} \mapsto \mathbb{R}_+$ be the Kullback-Leibler divergence,  and suppose that for all $i \neq j \in [|\mathcal{V}_{\epsilon}|]$, 
\begin{align}
    & \textrm{d} (\bm{v}_i, \bm{v}_j) \geq \alpha ~ \text{and} ~ \mathrm{KL}(\mathcal{D}^{(n)}(\lambda; \bm{v}_i) ~\|~ \mathcal{D}^{(n)}(\lambda; \bm{v}_j)) \leq \beta. \label{cond:g-fano-method}
\end{align}
Then any estimator $\widehat{\bm{v}}$ satisfies 
\begin{align}
    \max_{\bm{v}_i \in \mathcal{V}_{\epsilon}} \mathbb{E}_{\mathcal{D}^{(n)}(\lambda; \bm{v}_i)} \left[\textrm{d}(\widehat{\bm{v}}, \bm{v}_i) \right] \geq \frac{\alpha}{2} \left( 1 - \frac{\beta + \log 2}{\log |\mathcal{V}_{\epsilon}|} \right). \notag
\end{align}
\end{proposition}
We state one auxilary lemma and defer the proof to the end.
\begin{lemma} \label{lemma:local-packing}
\textbf{Local Packing.} For any $\xi \in [3/4, 1)$, for any $\epsilon \in (0,1)$, there exists a set $\mathcal{V}_{\epsilon} \subseteq \mathcal{S}^{d - 1} \cap \mathcal{M}$ such that $\sqrt{2(1 - \xi)}\epsilon \leq \left\| \bm{v}_i - \bm{v}_j \right\|_2 \leq \sqrt{2} \epsilon$ for all $\bm{v}_i \neq \bm{v}_j \in \mathcal{V}_{\epsilon}$, and 
\begin{align}
    \log |\mathcal{V}_{\epsilon}| \geq \log \frac{|\mathcal{Z}_*|}{\max_{\bm{z} \in \mathcal{Z}_{*}}\big|\mathcal{N}_H(\bm{z}; 2 (1 - \xi) k)\big| }.
\end{align}
\end{lemma}
Now we present the proof of part (b) of Theorem~\ref{thm:fund-limits} using Proposition~\ref{prop:g-fano-method} and Lemma~\ref{lemma:local-packing}.
\begin{proof}
For any $\xi \in [3/4, 1)$ and any $\epsilon \in (0,1)$, let $\mathcal{V}_{\epsilon} \subseteq \mathcal{S}^{d - 1} \cap \mathcal{M}$ which satisfies the desired properties mentioned in Lemma~\ref{lemma:local-packing}. Using Lemma~\ref{lemma:local-packing}, we obtain that for any $\bm{v}_i \neq \bm{v}_j \in \mathcal{V}_{\epsilon}$
\begin{align*}
    \|\bm{v}_i\bm{v}_i^{\top} - \bm{v}_j\bm{v}_j^{\top}\|_F^2 \geq \|\bm{v}_i - \bm{v}_j\|_2^2 > 2(1 - \xi) \epsilon^2.
\end{align*}
Note that
\begin{align*}
    \mathrm{KL}(\mathcal{D}(\lambda; \bm{v}_i) ~\|~ \mathcal{D}(\lambda; \bm{v}_j)) 
    = & ~ \frac{\lambda}{2(1 + \lambda)} \big[ (1 + \lambda) \text{tr}\left( (\bm{I}_d - \bm{v}_j \bm{v}_j^{\top}) \bm{v}_i \bm{v}_i^{\top}  \right) \\
    & ~ - \text{tr}\left( \bm{v}_j \bm{v}_j^{\top} (\bm{I}_d - \bm{v}_i \bm{v}_i^{\top})  \right) \big]  \\
    = & ~\frac{\lambda^2}{4(1 + \lambda)} \|\bm{v}_i\bm{v}_i^{\top} - \bm{v}_j\bm{v}_j^{\top}\|_F^2  \\
    \leq & ~ \frac{\lambda^2}{(1 + \lambda)}\|\bm{v}_i - \bm{v}_j\|_2^2. 
\end{align*}
Consequently, for $\bm{v}_i,\bm{v}_j$ such that $\|\bm{v}_i - \bm{v}_j\|_2^2 \leq 2 \epsilon^2$, we obtain that
\begin{align*}
    \mathrm{KL}(\mathcal{D}^{(n)}(\lambda; \bm{v}_i) ~\|~ \mathcal{D}^{(n)}(\lambda; \bm{v}_j)) \leq  \frac{\lambda^2 n 2 \epsilon^2}{(1 + \lambda)}.
\end{align*}
Now, applying Proposition~\ref{prop:g-fano-method} yields 
\begin{align}\label{ineq-fano-method}
    & ~ \max_{\bm{v}_i \in \mathcal{V}_{\epsilon}} \mathbb{E}_{\mathcal{D}^{(n)}(\lambda; \bm{v}_i)}  \big\|\widehat{\bm{v}}\widehat{\bm{v}}^{\top} - \bm{v}_i\bm{v}_i^{\top}\big\|_F \geq \frac{\sqrt{2(1 - \xi)}\epsilon}{2} \left(1 - \frac{\frac{\lambda^2 n 2 \epsilon^2}{(1 + \lambda)} + \log 2}{\log |\mathcal{V}_{\epsilon}|} \right).
\end{align}
Let
\begin{align*}
    & ~ \epsilon^2 = \min \bigg\{ 1, ~ \frac{1 + \lambda}{8 \lambda^2} \cdot \frac{ \log \left(|\mathcal{Z}_*|\right) - \log \big( \max_{\bm{z} \in \mathcal{Z}_{*}}|\mathcal{N}_H(\bm{z}; 2(1 - \xi) k)| \big)}{n} \bigg\}.
\end{align*}
Consequently, we obtain that
\begin{align}\label{bounds1-fano-method}
    \frac{\lambda^2 n 2 \epsilon^2}{(1 + \lambda)\log |\mathcal{V}_{\epsilon}|} \leq \frac{1}{4}.
\end{align}
Continuing, using Lemma~\ref{lemma:local-packing} yields that, for $\xi \in [3/4, 1)$ such that Assumption~\ref{assump:minimax-assumption} holds,
\begin{align*}
    \log (|\mathcal{V}_{\epsilon}|) \geq \log \frac{|\mathcal{Z}_*|}{\max_{\bm{z} \in \mathcal{Z}_{*}}\big|\mathcal{N}_H(\bm{z}; 2 (1 - \xi) k)\big| } \geq 4\log 2.
\end{align*}
Consequently, we obtain that
\begin{align}\label{bounds2-fano-method}
    \frac{\log 2}{\log |\mathcal{V}_{\epsilon}|} \leq \frac{1}{4}.
\end{align}
Substituing inequalities~\eqref{bounds1-fano-method}~\eqref{bounds2-fano-method} into inequality~\eqref{ineq-fano-method} yields 
\begin{align*}
    & ~\max_{\bm{v}_i \in \mathcal{V}_{\epsilon}} \mathbb{E}_{\mathcal{D}^{(n)}(\lambda; \bm{v}_i)} \left[ \|\widehat{\bm{v}}\widehat{\bm{v}}^{\top} - \bm{v}_i\bm{v}_i^{\top}\|_F \right]  \notag\\
    \geq & ~ \frac{\sqrt{2(1 - \xi)}}{4} \min\bigg\{1, ~ \sqrt{\frac{1 + \lambda}{8 \lambda^2}} \sqrt{\frac{ \log \big(|\mathcal{Z}_*|\big) - \log \big( \max_{\bm{z} \in \mathcal{Z}_{*}}\big|\mathcal{N}_H(\bm{x}; 2(1 - \xi) k)\big| \big)}{n}} \bigg\},
\end{align*}
which concludes the proof.
\end{proof}

\paragraph{Proof of Lemma~\ref{lemma:local-packing}}
We first claim that, for any $\xi \in [3/4, 1)$, there exists a set $\mathcal{Z}_{\xi} \subseteq \mathcal{Z}_*$ such that $2 (1 - \xi) k < \delta_H(\bm{z}, \bm{z}') \leq 2(k-1)$ for all $\bm{z} \neq \bm{z}' \in \mathcal{Z}_{\xi}$ and 
\begin{align}\label{claim-proof-local-pack}
    \log |\mathcal{Z}_{\xi}| \geq \log \big(|\mathcal{Z}_*|\big) - \log \big( \max_{\bm{z} \in \mathcal{Z}_*}\big|\mathcal{N}_H(\bm{z}; 2 (1 - \xi) k)\big| \big).
\end{align}
We defer the proof of Claim~\eqref{claim-proof-local-pack} to the end. Now we use Claim~\eqref{claim-proof-local-pack} to prove Lemma~\ref{lemma:local-packing}. Recall $\mathcal{Z}_{*}$ is defined in Eq~\eqref{definition-Z-star} and $i_*$ is defined in Eq~\eqref{definition-i-star}. For any $\epsilon \in (0,1)$, define the bijective mapping $\psi : \mathcal{Z}_{\xi} \mapsto \mathbb{R}^d$ defined as 
\begin{align*}
    [\psi(\bm{z})]_i = \left\{ 
    \begin{array}{lll}
        \bm{z}(i_*) \cdot \sqrt{1 - \epsilon^2} & \text{if } i = i_*  \\
        \bm{z}(i) \cdot \epsilon / \sqrt{k - 1} & \text{if } i \neq i_*
    \end{array}
    \right. .
\end{align*}
By the definitions of $\mathcal{Z}_{\xi}$ and $i_*$, we obtain that $\bm{z}({i_*}) = 1$ when $\bm{z} \in \mathcal{Z}_{\xi}$. Let $\bm{\Phi} := [\phi_1 ~|~ \cdots ~|~ \phi_d] \in \mathbb{R}^{d \times d}$ be the unitary matrix generated from the orthonormal basis $\mathcal{B}$ stated in Definition~\ref{cond:linear-structure}. Now we show that the set 
\begin{align*}
    \mathcal{V}_{\epsilon} := \left\{ \bm{\Phi} \psi(\bm{z}) ~\left|~ \bm{z} \in \mathcal{Z}_{\xi} \right.\right\}
\end{align*}
has the desired properties mentioned in Lemma~\ref{lemma:local-packing}. 
\begin{itemize}
    \item First, we verify that $\mathcal{V}_{\epsilon} \subseteq \mathcal{S}^{d - 1} \cap \mathcal{M}$. Note that for any $\bm{z} \in \mathcal{Z}_{\xi} \subseteq \mathcal{Z}_{*}$, we have $\bm{z} = \bm{z}_{m}$ for some $m\in [M]$, where $\bm{z}_{m}$ is defined in Eq~\eqref{definition-z-m}. Consequently,
    \begin{align*}
        \mathsf{supp}(\psi(\bm{z})) = \mathsf{supp}(\bm{z}) = \mathsf{supp}(\bm{z}_m).
    \end{align*}
    It immediately follows that
    \begin{align*}
        \bm{\Phi} \psi(\bm{z}) \in \text{span}(\mathcal{B}_{m}) \subseteq \mathcal{M}.
    \end{align*}
    Moreover,
    \begin{align*}
        \|\bm{\Phi} \psi(\bm{z})\|_2^2 = & ~ \psi(\bm{z})^{\top} \bm{\Phi}^{\top} \bm{\Phi} \psi(\bm{z}) = \|\psi(\bm{z})\|_2^2 \\
        = & ~ 1 - \epsilon^2 + \sum_{i \neq i_*} \bm{z}(i)^2 \frac{\epsilon^2}{k - 1} = 1.
    \end{align*} 
    \item Second, for all pairs of points $\bm{\Phi}\psi(\bm{z}), \bm{\Phi}\psi(\bm{z}' \in \mathcal{V}_{\epsilon}$ such that $\bm{z} \neq \bm{z}' \in \mathcal{Z}_{\xi}$, we obtain that 
    \begin{align*}
        & ~ \left\| \bm{\Phi}\psi(\bm{z}) - \bm{\Phi}\psi(\bm{z}') \right\|_2^2 = \left\| \psi(\bm{z}) - \psi(\bm{z}') \right\|_2^2 \\
        = & ~ \delta_H(\bm{z}, \bm{z}') \frac{\epsilon^2}{k - 1} \leq 2(k - 1) \frac{\epsilon^2}{k - 1} = 2 \epsilon^2,
    \end{align*}
    where we use $\bm{z}(i_*) = \bm{z}'(i_*) = 1$ and $|\mathsf{supp}(\bm{z})| = |\mathsf{supp}(\bm{z}')| = k$ for any $\bm{z}, \bm{z}' \in \mathcal{Z}_{\xi}$. Similarly, we have
    \begin{align*}
        & ~ \left\| \bm{\Phi}\psi(\bm{z}) - \bm{\Phi}\psi(\bm{z}') \right\|_2^2 = \left\| \psi(\bm{z}) - \psi(\bm{z}') \right\|_2^2 \\
        = & ~ \delta_H(\bm{z}, \bm{z}') \frac{\epsilon^2}{k - 1} \geq 2(1 - \xi) k \frac{\epsilon^2}{k - 1} > 2(1 - \xi)\epsilon^2, 
    \end{align*}
    where the last step holds due to the definition of $\mathcal{Z}_{\xi}$. 
    \item Finally, by the definition of $\mathcal{V}_{\epsilon}$, we have $|\mathcal{V}_{\epsilon}| = |\mathcal{Z}_{\xi}|$. 
\end{itemize}
This concludes the proof of Lemma~\ref{lemma:local-packing}. It remains to show claim~\eqref{claim-proof-local-pack}.

\paragraph{Proof of Claim~\eqref{claim-proof-local-pack}} 
By assumption~\ref{assump:minimax-assumption}, for any $\bm{z} \in \mathcal{Z}_*$, it satisfies $|\mathsf{supp}(\bm{z})| = k$. Consequently, we obtain that for $\bm{z} \neq \bm{z}' \in \mathcal{Z}_{*}$, we have
\begin{align*}
    \delta_{H}(\bm{z}, \bm{z}') = 2 \big( k - |\{i \in [d] ~|~ \bm{z}(i) = \bm{z}'(i) = 1\}| \big) \leq 2(k-1),
\end{align*}
where in the last step we use $\bm{z}(i_*) = \bm{z}'(i_*) = 1$. For any $\xi \in [3/4, 1)$, consider a \emph{maximal} set $\mathcal{Z}_{\xi} \subseteq \mathcal{Z}_{*}$ such that $\delta_H(\bm{z}, \bm{z}') > 2 (1 - \xi) k$ for all $\bm{z} \neq \bm{z}' \in \mathcal{Z}_{\xi}$. Here the maximal means the set $\mathcal{Z}_{\xi}$ cannot be augmented by adding any other points in $\mathcal{Z}_{*} \backslash \mathcal{Z}_{\xi}$. Consequently, we obtain that
\[
    \mathcal{Z}_{*} \subseteq \cup_{\bm{z} \in \mathcal{Z}_{\xi}} \;\mathcal{N}_H(\bm{z}; 2(1 - \xi)k).
\]
Otherwise, there exists a $\tilde{\bm{z}} \in \mathcal{Z}_{*} \backslash \mathcal{Z}_{\xi}$ such that $\delta_H(\bm{z}, \tilde{\bm{z}}) > 2(1 - \xi) k$ for all $\bm{z} \in \mathcal{Z}_{\xi}$, which contradicts the maximality of $\mathcal{Z}_{\xi}$. Therefore,
\begin{align*}
    |\mathcal{Z}_*| \leq & ~ \sum_{\bm{z} \in \mathcal{Z}_{\xi}} |\mathcal{N}_H(\bm{z}; 2 (1 - \xi) k)| \leq |\mathcal{Z}_{\xi}| \cdot \max_{\bm{z} \in \mathcal{Z}_{*}}|\mathcal{N}_H(\bm{z}; 2 (1 - \xi) k)|. 
\end{align*}
Consequently,
\begin{align*}
    |\mathcal{Z}_{\xi}| \geq & ~  \frac{|\mathcal{Z}_*|}{\max_{\bm{z} \in \mathcal{Z}_{*}}|\mathcal{N}_H(\bm{z}; 2 (1 - \xi) k)|}.
\end{align*}
This completes the proof.

%% file: local-convergence-proof.tex
\subsection{Proof of Theorem~\ref{thm:convergence}}\label{app:PPM-EM}
We start by introducing some notation. Let $L^{(t)} \in \mathcal{L}$ denote the linear subspace that contains the iterate $\bm{v}_{t}$ in Algorithm~\ref{alg:PPM}. 
Define the linear subspace $F^{(t)} = \mathsf{conv}(L^{(t)} \cup L^{(t+1)} \cup L_{*})$ as the convex hull of $L^{(t)},L^{(t+1)}$ and $L_{*}$. Let
\[
    \widehat{\bm{v}}_{F^{(t)}} = \argmax_{\bm{v} \in F^{(t)} \cap \mathcal{S}^{d-1}} \bm{v}^{\top} \widehat{\bm{\Sigma}} \bm{v}.
\] 
Now we state three lemmas and defer their proofs to subsections~\ref{proof-lemma-prop:F-est},~\ref{proof-lemma-equivalence-of-v} and~\ref{proof-lemma-dist-vF-v'}. 

\begin{lemma} \label{prop:F-est}
Under the setting of linear structure condition~\ref{cond:linear-structure}, for any linear subspace $F$ such that $\bm{v}_* \in L_* \subseteq F$, let 
\[
    \widehat{\bm{v}}_F = \argmax_{\bm{v} \in F \cap \mathcal{S}^{d-1}} \bm{v}^{\top} \bm{\widehat{\Sigma}} \bm{v}^{\top}.
\]
If the eigen-gap $\lambda$ satisfies $\lambda > 2 \rho(\bm{W}, F)$, then 
\begin{align*}
    \|\widehat{\bm{v}}_F - \bm{v}_*\|_2 \leq \frac{\rho(\bm{W}, F)}{\lambda - 2\rho(\bm{W}, F) }.
\end{align*}
\end{lemma}

\begin{lemma}\label{equivalence-of-v}
Let $\{\bm{v}_{t+1}\}_{t=0}^{T}$ be iterates in Algoritm~\ref{alg:PPM}. Then for each $0\leq t \leq T$, we have that
\[
    \bm{v}_{t+1} = \argmin_{\bm{v} \in \mathcal{M} \cap \mathcal{S}^{d-1}} \bigg\| \bm{v} - \frac{\widehat{\bm{\Sigma}}_{F^{(t)}} \bm{v}_{t} }{\| \widehat{\bm{\Sigma}}_{F^{(t)}} \bm{v}_{t} \|_{2}} \bigg\|_{2}, 
\]
i.e., replacing $\widehat{\bm{\Sigma}} \bm{v}_{t}$ by $\widehat{\bm{\Sigma}}_{F^{(t)}} \bm{v}_{t}$ equivalently without changing the left-hand-side result $\bm{v}_{t+1} $. 
\end{lemma}
\begin{lemma}\label{distance-vF-v'}
Let $\lambda_{1}^{F^{(t)}},\lambda_{2}^{F^{(t)}}$ be the first and the second eigenvalues of $\widehat{\bm{\Sigma}}_{F^{(t)}}$. Let 
\begin{align*}
 & ~ \widehat{\kappa}_{F^{(t)}} = \lambda_{2}^{F^{(t)}}/\lambda_{1}^{F^{(t)}},\quad \alpha_{t} = \langle \widehat{\bm{v}}_{F^{(t)}} , \bm{v}_{*} \rangle, \quad \bm{v}_{t+1}' = \frac{\widehat{\bm{\Sigma}}_{F^{(t)}} \bm{v}_{t} }{\| \widehat{\bm{\Sigma}}_{F^{(t)}} \bm{v}_{t} \|_{2}}.
\end{align*}
Then we have 
\[
    \left\| \widehat{\bm{v}}_{F^{(t)}} - \bm{v}_{t+1}' \right\|_{2} \leq \frac{\widehat{\kappa}_{F^{(t)}}}{\alpha_{t}} \cdot \left\| \bm{v}_{t} - \bm{v}_{*} \right\|_{2} + 
    \frac{\widehat{\kappa}_{F^{(t)}}}{\alpha_{t}} \cdot \frac{\rho(\bm{W}, F^{(t)})}{ \lambda - 2\rho(\bm{W}, F^{(t)})}.
\]
\end{lemma}
We are now ready to prove Theorem~\ref{thm:convergence} using Lemma~\ref{prop:F-est}, Lemma~\ref{equivalence-of-v} and Lemma~\ref{distance-vF-v'}.
\begin{proof}
We claim that if $\bm{v}_{t} \in \mathbb{G}(\lambda)$ then

\begin{align}\label{claim-proof-convergence}
	\| \bm{v}_{t+1} - \bm{v}_{*} \|_{2} \leq \frac{1}{2} \cdot \| \bm{v}_{t} - \bm{v}_{*} \|_{2} + \frac{2.5\rho(\bm{W},F^{*})}{\lambda - 2\rho(\bm{W},F^{*})}.
\end{align}
We defer the proof of inequality~\eqref{claim-proof-convergence} to the end. Now we prove Theorem~\ref{thm:convergence} by induction and using inequality~\eqref{claim-proof-convergence}.\\
\textbf{Base case:} $t=0$. Since $\bm{v}_{0} \in \mathbb{G}(\lambda)$ by assumption, then inequality~\eqref{ineq-thm-convergence} follows immediately from inequality~\eqref{claim-proof-convergence}. \\
\textbf{Induction:} Suppose inequality~\eqref{ineq-thm-convergence} holds for $0 \leq t \leq k$. To reduce the notational burden, we use the shorhand
\[
    \epsilon_{*} = \frac{2.5\rho(\bm{W},F^{*})}{\lambda - 2\rho(\bm{W},F^{*})}.
\]
Then by applying inequality~\eqref{ineq-thm-convergence} recursively, we obtain that
\begin{align}\label{recursive-ineq}
    \|\bm{v}_{k+1} - \bm{v}_{*} \|_{2} \leq \frac{1}{2^{k+1}} \cdot \| \bm{v}_{0} - \bm{v}_{*} \|_{2} +  \sum_{t=0}^{k}\frac{1}{2^{t}}\cdot \epsilon_{*}.
\end{align}
We then consider two cases.\\
\textbf{Case 1:} $\| \bm{v}_{0} - \bm{v}_{*} \|_{2} \leq 2\epsilon_{*}$. Then we obtain that 
\[
\|\bm{v}_{k+1} - \bm{v}_{*} \|_{2} \leq \big( \sum_{t=0}^{k}\frac{1}{2^{t}} + \frac{1}{2^{k}} \big) \epsilon_{*} \leq 2\epsilon_{*}.
\]
Using the fact that $\bm{v}_{k+1},\bm{v}_{*}$ are unit norm vectors, we obtain that 
\[
    \langle\bm{v}_{k+1}, \bm{v}_{*} \rangle \geq  1 - 2\epsilon_{*}^{2} \geq 1-2\epsilon_{*} \geq 1 - \frac{5\rho(\bm{W},F^{*})}{\lambda - 2\rho(\bm{W},F^{*})},
\]
where in the last inequality we use $\epsilon_{*}<1$ by the inequality~\eqref{eigen-gap-condition}. Consequently, by the assumption that $\lambda$ is large enough so that inequality~\eqref{eigen-gap-condition} holds, we obtain that $\bm{v}_{k+1} \in \mathbb{G}(\lambda)$. Then the induction step follows immediately from claim~\eqref{claim-proof-convergence}.\\
\textbf{Case 2:} $\| \bm{v}_{0} - \bm{v}_{*} \|_{2} > 2\epsilon_{*}$. In this case, we obtain that
\[
     \sum_{t=0}^{k}\frac{1}{2^{t}} \cdot \epsilon_{*} \leq 2\cdot(1-2^{-(k+1)}) \cdot \epsilon_{*} \leq (1-2^{-(k+1)}) \cdot \| \bm{v}_{0} - \bm{v}_{*} \|_{2}.
\]
Combining with inequality~\eqref{recursive-ineq}, we obtain that
\begin{align*}
    \|\bm{v}_{k+1} - \bm{v}_{*} \|_{2} 
    \leq & ~ \|\bm{v}_{0} - \bm{v}_{*} \|_{2} + \sum_{t=0}^{k}\frac{1}{2^{t}} \cdot \epsilon_{*} - (1-2^{-(k+1)}) \cdot \| \bm{v}_{0} - \bm{v}_{*} \|_{2} \\
    \leq & ~ \|\bm{v}_{0} - \bm{v}_{*} \|_{2}.
\end{align*}
Consequently, we obtain $\langle\bm{v}_{k+1}, \bm{v}_{*} \rangle \geq \langle\bm{v}_{0}, \bm{v}_{*} \rangle$. Then $\bm{v}_{k+1} \in \mathbb{G}(\lambda)$. Then the induction step follows immediately from inequality~\eqref{claim-proof-convergence}.
We now apply Lemmas~\ref{prop:F-est},~\ref{equivalence-of-v} and~\ref{distance-vF-v'} to prove the claim~\eqref{claim-proof-convergence}.
\paragraph{Proof of inequality~\eqref{claim-proof-convergence}}
Let $\widehat{\kappa}_{F^{(t)}},\alpha_{t}$ and $\bm{v}_{t+1}'$ be defined as in Lemma~\ref{distance-vF-v'}. Using the triangular inequality, we obtain that
\begin{align*}
\|\bm{v}_{t+1} - \bm{v}_{*} \|_{2} 
\leq & ~ \|\bm{v}_{t+1} - \bm{v}_{t+1}' \|_{2}  + \|\bm{v}_{t+1}' - \bm{v}_{*} \|_{2} \\ 
\overset{(1)}{\leq} & ~ 2\|\bm{v}_{t+1}' - \bm{v}_{*} \|_{2} \leq 2\|\bm{v}_{t+1}' - \widehat{\bm{v}}_{F^{(t)}} \|_{2} + 2\|\widehat{\bm{v}}_{F^{(t)}} - \bm{v}_{*} \|_{2},
\end{align*}
where in step $(1)$ we use $\|\bm{v}_{t+1} - \bm{v}_{t+1}' \|_{2} \leq \|\bm{v}_{t+1}' - \bm{v}_{*} \|_{2}$ following directly from Lemma~\ref{equivalence-of-v}. Since $L_{*} \subset F^{(t)}$, using Lemma~\ref{prop:F-est}, we obtain that
\begin{align}\label{part1-proof-claim-thm-convergence}
\|\widehat{\bm{v}}_{F^{(t)}} - \bm{v}_{*} \|_{2} \leq \frac{\rho(\bm{W},F^{(t)})}{\lambda - 2\rho(\bm{W},F^{(t)})} \leq \frac{\rho(\bm{W},F^{*})}{\lambda - 2\rho(\bm{W},F^{*})}.
\end{align}
Combining inequality~\eqref{part1-proof-claim-thm-convergence} with Lemma~\ref{distance-vF-v'}, we obtain that
\begin{align}\label{ineq-one-step-away}
    \|\bm{v}_{t+1} - \bm{v}_{*} \|_{2} \leq & ~ \frac{2\widehat{\kappa}_{F^{(t)}}}{\alpha_{t}} \cdot \left\| \bm{v}_{t} - \bm{v}_{*} \right\|_{2} + 
    \big(\frac{2\widehat{\kappa}_{F^{(t)}}}{\alpha_{t}} + 2\big) \cdot \frac{\rho(\bm{W}, F^{(t)})}{ \lambda - 2\rho(\bm{W}, F^{(t)})}. 
\end{align}
Note that
\begin{align*}
\lambda_{1}^{F^{(t)}} \geq & ~ \bm{v}_{*}^{\top} \widehat{\bm{\Sigma}}_{F^{(t)}} \bm{v}_{*} \\
= & ~ \bm{v}_{*}^{\top} \bm{\Sigma} \bm{v}_{*} + \bm{v}_{*}^{\top} \bm{W} \bm{v}_{*} \geq \lambda + 1 - \rho(\bm{W},F^{(t)}) \quad \text{and}\\
\lambda_{2}^{F^{(t)}} \leq & ~ \lambda_{2}(\bm{\Sigma}_{F^{(t)}}) + \rho(\bm{W},F^{(t)}) \leq \lambda_{2}(\bm{\Sigma}) + \rho(\bm{W},F^{(t)}) \\
= & ~ 1 + \rho(\bm{W},F^{(t)}).
\end{align*}
Moreover, we note that
\begin{align*}
\alpha_{t} = \langle \widehat{\bm{v}}_{F^{(t)}} , \bm{v}_{*} \rangle &= \langle \bm{v}_{t}, \bm{v}_{*} \rangle + \langle \widehat{\bm{v}}_{F^{(t)}} - \bm{v}_{t}, \bm{v}_{*} \rangle \\
&\geq \langle \bm{v}_{t}, \bm{v}_{*} \rangle - \| \widehat{\bm{v}}_{F^{(t)}} - \bm{v}_{t} \|_{2} \\
&\geq \frac{4(1+\rho(\bm{W},F^{*}))}{\lambda+1 - \rho(\bm{W},F^{*})},
\end{align*}
where the last step follows from $\bm{v}_{t} \in \mathbb{G}(\lambda)$ and inequality~\eqref{part1-proof-claim-thm-convergence}.
Putting the pieces together yields
\begin{align}\label{coeff-upper-bound}
    \frac{\widehat{\kappa}_{F^{(t)}}}{\alpha_{t}} = \frac{\lambda_{2}^{F^{(t)}}}{\lambda_{1}^{F^{(t)}} \alpha_{t}} \leq \frac{1}{4}.
\end{align}
Substituting inequality~\eqref{coeff-upper-bound} into inequality~\eqref{ineq-one-step-away} yields the desired result and concludes the proof.
\end{proof}

\subsubsection{Proof of Lemma~\ref{prop:F-est}}\label{proof-lemma-prop:F-est}
\begin{proof}
Note that we have the decomposition
\begin{align*}
	& ~ \widehat{\bm{v}}_F = \alpha \bm{v}_* + \beta \bm{v}_{\bot} \quad \text{such that} \quad \alpha^2 + \beta^2 = 1, \|\bm{v}_*\|_2^2 = \|\bm{v}_{\bot}\|_2^2 = 1, \langle \bm{v}_*, \bm{v}_{\bot}\rangle = 0.
\end{align*}
 By definition, $\widehat{\bm{v}}_F$ is the eigenvector with respect to the largest eigenvalue $\lambda_{\max}(\widehat{\bm{\Sigma}}_F)$ of $\widehat{\bm{\Sigma}}_F$. We obtain that
\begin{align*}
    \bm{v}_{\bot}^{\top} \widehat{\bm{\Sigma}}_F \widehat{\bm{v}}_F &= \lambda_{\max}(\widehat{\bm{\Sigma}}_F) \bm{v}_{\bot}^{\top} \widehat{\bm{v}}_F = \lambda_{\max}(\widehat{\bm{\Sigma}}_F) \cdot \beta \quad \text{and} \\
    \bm{v}_{\bot}^{\top} \widehat{\bm{\Sigma}}_F \widehat{\bm{v}}_F &= \bm{v}_{\bot}^{\top} \widehat{\bm{\Sigma}}_F (\alpha \bm{v}_* + \beta \bm{v}_{\bot}) = \alpha \cdot \bm{v}_{\bot}^{\top} \widehat{\bm{\Sigma}}_F \bm{v}_* + \beta \cdot \bm{v}_{\bot}^{\top} \widehat{\bm{\Sigma}}_F \bm{v}_{\bot}.
\end{align*} 
Consequently, we obtain that
\begin{align}\label{ineq-beta-lemma-F-est}
     |\beta| = & ~ |\alpha| \frac{|\bm{v}_{\bot}^{\top} \widehat{\bm{\Sigma}}_F \bm{v}_*|}{\lambda_{\max}(\widehat{\bm{\Sigma}}_F) - \bm{v}_{\bot}^{\top} \widehat{\bm{\Sigma}}_F \bm{v}_{\bot}} \\
     \overset{(1)}{=} & ~ |\alpha| \frac{| \bm{v}_{\bot}^{\top} \bm{W}_F \bm{v}_*|}{\lambda_{\max}(\widehat{\bm{\Sigma}}_F) - \bm{v}_{\bot}^{\top} \widehat{\bm{\Sigma}}_F \bm{v}_{\bot}} \notag \\
     \leq & ~ |\alpha| \frac{\rho(\bm{W}, F)}{\lambda - 2\rho(\bm{W}, F) }, \notag
\end{align}
where in step $(1)$ we use 
\begin{align*}
	\bm{v}_{\bot}^{\top} \widehat{\bm{\Sigma}}_F \bm{v}_* = & ~ \bm{v}_{\bot}^{\top} \bm{\Sigma}_F \bm{v}_* + \bm{v}_{\bot}^{\top} \bm{W}_F \bm{v}_* \\
	= & ~ (\lambda + 1) \bm{v}_{\bot}^{\top}\bm{v}_* + \bm{v}_{\bot}^{\top} \bm{W}_F \bm{v}_* \\
	= & ~ \bm{v}_{\bot}^{\top} \bm{W}_F \bm{v}_*, 
\end{align*}
and in the last step we use
\begin{align*}
    & ~ \lambda_{\max}(\widehat{\bm{\Sigma}}_F) - \bm{v}_{\bot}^{\top} \widehat{\bm{\Sigma}}_F \bm{v}_{\bot} \\
    \geq & ~ \lambda_{\max}(\widehat{\bm{\Sigma}}_F) - \lambda_2 (\widehat{\bm{\Sigma}}_F) \notag\\
    = & ~ \lambda_{\max}(\bm{\Sigma}_F + \bm{W}_F) - \lambda_2(\bm{\Sigma}_F + \bm{W}_F) \notag\\
    \geq & ~ \lambda_{\max}(\bm{\Sigma}_F) - \lambda_{\max}(\bm{W}_F) - \lambda_2(\bm{\Sigma}_F) - \lambda_{\max}(\bm{W}_F) \notag\\
    = & ~ \lambda_{\max}(\bm{\Sigma}_F) - \lambda_2(\bm{\Sigma}_F) - 2 \rho(\bm{W}, F) \notag\\
    \geq & ~ \lambda - 2 \rho(\bm{W}, F).
\end{align*}
Applying inequality~\eqref{ineq-beta-lemma-F-est} yields
\begin{align*}
    \alpha^2 \left( 1 + \frac{\rho^2(\bm{W}, F)}{(\lambda - 2\rho(\bm{W}, F))^2} \right) \geq \alpha^2 + \beta^2 = 1.
\end{align*}
Consequently we obtain that
\[
    \alpha \geq \frac{1}{\sqrt{ 1 + \frac{\rho^2(\bm{W}, F)}{(\lambda - 2\rho(\bm{W}, F))^2} }} \geq 1 - \frac{\rho^2(\bm{W}, F)}{2(\lambda - 2\rho(\bm{W}, F))^2},
\]
where in the last step we use $1/\sqrt{1+x} \geq 1-x/2$ for $x>0$. Continuing, we obtain that
\begin{align*}
   & ~ \|\widehat{\bm{v}}_F - \bm{v}_*\|_2^2 = 2 - 2 \alpha \leq \frac{\rho^2(\bm{W}, F)}{(\lambda - 2\rho(\bm{W}, F))^2} \\
   \quad \Rightarrow & ~ \|\widehat{\bm{v}}_F - \bm{v}_*\|_2 \leq \frac{\rho(\bm{W}, F)}{\lambda - 2\rho(\bm{W}, F) },
\end{align*}
which completes the proof. 
\end{proof}

\subsubsection{Proof of Lemma~\ref{equivalence-of-v}}\label{proof-lemma-equivalence-of-v}
\begin{proof}
Note that
\begin{align}\label{piece1-proof-lemma-equi}
 \bm{\theta} = \argmin_{\bm{v} \in \mathcal{M} \cap \mathcal{S}^{d-1}} \bigg\| \bm{v} - \frac{\widehat{\bm{\Sigma}}_{F^{(t)}} \bm{v}_{t} }{\| \widehat{\bm{\Sigma}}_{F^{(t)}} \bm{v}_{t} \|_{2}} \bigg\|_{2} = \argmax_{\bm{v} \in \mathcal{M} \cap \mathcal{S}^{d-1}} \langle \bm{v}, \widehat{\bm{\Sigma}}_{F^{(t)}} \bm{v}_{t} \rangle.
\end{align}
Let $L' \in \mathcal{L}$ be the linear subspace such that $\bm{\theta} \in L'$. By definition~\ref{cond:linear-structure}, there are $\mathcal{B}_{L'},\mathcal{B}_{F^{(t)}} \subseteq \mathcal{B} = \{\phi_1,\cdots,\phi_{d}\}$ such that $L' = \mathsf{span}(\mathcal{B}_{L'})$ and $F^{(t)} = \mathsf{span}(\mathcal{B}_{F^{(t)}})$. Hence we obtain that $\bm{P}_{L'} \bm{P}_{F^{(t)}} = \bm{P}_{L' \cap F^{(t)}}$, where $L' \cap F^{(t)} = \mathsf{span}(\mathcal{B}_{L'} \cap \mathcal{B}_{F^{(t)}})$. Consequently, we obtain that
we obtain that
\[
    \bm{\theta} = \frac{\bm{P}_{L'} \widehat{\bm{\Sigma}}_{F^{(t)}} \bm{v}_{t} }{ \| \bm{P}_{L'} \widehat{\bm{\Sigma}}_{F^{(t)}} \bm{v}_{t} \|_{2}} = \frac{\bm{P}_{L'} \bm{P}_{F^{(t)}} \widehat{\bm{\Sigma}} \bm{v}_{t} }{ \| \bm{P}_{L'} \widehat{\bm{\Sigma}}_{F^{(t)}} \bm{v}_{t} \|_{2}} = \frac{\bm{P}_{L' \cap F^{(t)}}  \widehat{\bm{\Sigma}} \bm{v}_{t} }{ \| \bm{P}_{L'} \widehat{\bm{\Sigma}}_{F^{(t)}} \bm{v}_{t} \|_{2}}.
\] 
It immediately shows that $\bm{\theta} \in F^{(t)}$. On the other hand, it follows from Algorithm~\ref{alg:PPM} that 
\begin{align*}
    \bm{v}_{t+1}^{\mathcal{M}} &= \argmin_{\bm{v} \in \mathcal{M}} \bigg\|\bm{v} - \frac{\widehat{\bm{\Sigma}} \bm{v}_{t}}{\|\widehat{\bm{\Sigma}} \bm{v}_{t}\|_{2}}\bigg\|_{2}^{2} \\
    &= \argmin_{\bm{v} \in \mathcal{M}} \|\bm{v}\|_{2}^{2} - 2\langle \bm{v}/\|\bm{v}\|_{2}, \widehat{\bm{\Sigma}} \bm{v}_{t} \rangle \cdot \frac{\|\bm{v}\|_{2}}{\|\widehat{\bm{\Sigma}} \bm{v}_{t}\|_{2}}  + 1.
\end{align*}
Note that the minimization over $\|\bm{v}\|_{2}$ and $\bm{v}/\|\bm{v}\|_{2}$ are independent. Consequently, we obtain that
\begin{align}\label{piece2-proof-lemma-equi}
     \bm{v}_{t+1} = \frac{\bm{v}_{t+1}^{\mathcal{M}}}{\|\bm{v}_{t+1}^{\mathcal{M}}\|_{2}} = \argmax_{\bm{v} \in \mathcal{M} \cap \mathcal{S}^{d-1}} \langle \bm{v}, \widehat{\bm{\Sigma}} \bm{v}_{t} \rangle.
\end{align}
Putting the pieces together yields that
\begin{align*}
    \langle \bm{\theta}, \widehat{\bm{\Sigma}}_{F^{(t)}} \bm{v}_{t} \rangle \overset{(1)}{=} & ~ \langle \bm{\theta}, \widehat{\bm{\Sigma}} \bm{v}_{t} \rangle \overset{(2)}{\leq} \langle \bm{v}_{t+1}, \widehat{\bm{\Sigma}} \bm{v}_{t} \rangle \\
    \overset{(3)}{=} & ~ \langle \bm{v}_{t+1}, \widehat{\bm{\Sigma}}_{F^{(t)}} \bm{v}_{t} \rangle \overset{(4)}{\leq} \langle \bm{\theta}, \widehat{\bm{\Sigma}}_{F^{(t)}} \bm{v}_{t} \rangle,
\end{align*}
where step $(1)$ and $(3)$ follow from $\bm{\theta},\bm{v}_{t+1} \in F^{(t)}$, step $(2)$ follows from inequality~\eqref{piece2-proof-lemma-equi} and step $(4)$ follows from inequality~\eqref{piece1-proof-lemma-equi}. Consequently, we obtain that
\[
    \langle \bm{v}_{t+1}, \widehat{\bm{\Sigma}}_{F^{(t)}} \bm{v}_{t} \rangle = \langle \bm{\theta}, \widehat{\bm{\Sigma}}_{F^{(t)}} \bm{v}_{t} \rangle.
\]
Since the optimal solution in the last step of inequality~\eqref{piece1-proof-lemma-equi} is unique, we conclude that $\bm{v}_{t+1} = \bm{\theta}$.
\end{proof}

\subsubsection{Proof of Lemma~\ref{distance-vF-v'}}\label{proof-lemma-dist-vF-v'}
\begin{proof}
By definition, we obtain that
\begin{align*}
    \left\| \widehat{\bm{v}}_{F^{(t)}} - \bm{v}_{t + 1}' \right\|_2^2 = & ~ 2 - 2 \langle \widehat{\bm{v}}_{F^{(t)}}, \bm{v}_{t + 1}' \rangle \\
    = & ~ 2 - 2 \left\langle \widehat{\bm{v}}_{F^{(t)}},  \frac{\widehat{\bm{\Sigma}}_{F^{(t)}} \bm{v}_t}{\|\widehat{\bm{\Sigma}}_{F^{(t)}} \bm{v}_t\|_2} \right\rangle. 
\end{align*} 
Note that we have the decomposition
\begin{align*}
	& ~ \bm{v}_{t} = \alpha_{t} \cdot \widehat{\bm{v}}_{F^{(t)}} + \beta_{t} \cdot \widehat{\bm{v}}_{\perp}, \\
    \text{where} & ~ \langle \widehat{\bm{v}}_{\perp}, \widehat{\bm{v}}_{F^{(t)}} \rangle = 0,\quad \|\widehat{\bm{v}}_{\perp}\|_{2} = 1\quad \text{and} \quad \beta_{t} = \langle \bm{v}_{t}, \widehat{\bm{v}}_{\perp} \rangle. 
\end{align*}
Using the decomposition in the display above, we have 
\begin{align*}
    \left\langle \widehat{\bm{v}}_{F^{(t)}},  \frac{\widehat{\bm{\Sigma}}_{F^{(t)}} \bm{v}_t}{\|\widehat{\bm{\Sigma}}_{F^{(t)}} \bm{v}_t\|_2} \right\rangle = & ~ \frac{\widehat{\bm{v}}_{F^{(t)}}^{\top} \widehat{\bm{\Sigma}}_{F^{(t)}} (\alpha_t \widehat{\bm{v}}_{F^{(t)}} + \beta_t \widehat{\bm{v}}_{\perp}) }{\|\widehat{\bm{\Sigma}}_{F^{(t)}} (\alpha_t \widehat{\bm{v}}_{F^{(t)}} + \beta_t \widehat{\bm{v}}_{\perp}) \|_2} \\
    = & ~ \frac{\alpha_t \lambda_{1}^{F^{(t)}}}{\sqrt{ \alpha_t^2 (\lambda_{1}^{F^{(t)}})^2 + \beta_t^2 \|\widehat{\bm{\Sigma}}_{F^{(t)}} \bm{v}_{\perp}\|_2^2 }},
\end{align*}
where in the last step we use $\langle \widehat{\bm{v}}_{F^{(t)}}, \widehat{\bm{\Sigma}}_{F^{(t)}} \widehat{\bm{v}}_{\perp} \rangle = 0$. Continuing, we obtain that
\begin{align*}
    \frac{\alpha_t \lambda_{1}^{F^{(t)}}}{ \sqrt{ \alpha_t^2 (\lambda_{1}^{F^{(t)}})^2 + \beta_t^2 \|\widehat{\bm{\Sigma}}_{F^{(t)}} \bm{v}_{\perp}\|_2^2 } }  
    \geq & ~ \frac{\alpha_t \lambda_{1}^{F^{(t)}}}{ \sqrt{ \alpha_t^2 (\lambda_{1}^{F^{(t)}})^2 + \beta_t^2 (\lambda_{2}^{F^{(t)}})^2 } }  \\
    = & ~ \frac{1}{ \sqrt{ 1 + \frac{\beta_t^2}{\alpha_t^2} \widehat{\kappa}_{F^{(t)}}^2 } } \\
    \geq & ~ 1 - \frac{\beta_t^2}{\alpha_t^2} \frac{\widehat{\kappa}_{F^{(t)}}^2}{2},
\end{align*}
where we use $\|\widehat{\bm{\Sigma}}_{F^{(t)}} \bm{v}_{\perp}\|_2 \leq \lambda_{2}^{F^{(t)}}$ in the first step and we use $\frac{1}{\sqrt{1 + x}} \geq 1 - \frac{x}{2}$ for any $x \geq 0$ in the last step. Putting the three pieces together yields
\begin{align*}
    \left\| \widehat{\bm{v}}_{F^{(t)}} - \bm{v}_{t + 1}' \right\|_2^2 \leq & ~ \frac{\beta_t^2 \widehat{\kappa}_{F^{(t)}}^2 }{\alpha_t^2} = \frac{\widehat{\kappa}_{F^{(t)}}^2 (1 + \alpha_t)}{2 \alpha_t^2} \left\| \widehat{\bm{v}}_{F^{(t)}} - \bm{v}_{t} \right\|_2^2 \\
    \leq & ~ \frac{\widehat{\kappa}_{F^{(t)}}^2}{\alpha_t^2} \left\| \widehat{\bm{v}}_{F^{(t)}} - \bm{v}_{t} \right\|_2^2,
\end{align*}
where the final inequality holds since $1 + \alpha_t \leq 2$. Hence we get 
\begin{align*}
    \left\| \widehat{\bm{v}}_{F^{(t)}} - \bm{v}_{t + 1}' \right\|_2 \leq & ~ \frac{\widehat{\kappa}_{F^{(t)}}}{\alpha_t} \left\| \widehat{\bm{v}}_{F^{(t)}} - \bm{v}_{t} \right\|_2 \\
    \leq & ~ \frac{\widehat{\kappa}_{F^{(t)}}}{\alpha_t} \left\| \widehat{\bm{v}}_{F^{(t)}} - \bm{v}_* \right\|_2  + \frac{\widehat{\kappa}_{F^{(t)}}}{\alpha_t} \left\| \bm{v}_* - \bm{v}_{t} \right\|_2 \\
    \leq & ~ \frac{\widehat{\kappa}_{F^{(t)}}}{\alpha_t} \frac{\rho(\bm{W}, F^{(t)})}{\lambda - 2\rho(\bm{W}, F^{(t)})} + \frac{\widehat{\kappa}_{F^{(t)}}}{\alpha_t} \left\| \bm{v}_* - \bm{v}_{t} \right\|_2, 
\end{align*}
where the final inequality follows from Lemma~\ref{prop:F-est}. 
\end{proof}

%% file: corollaries-proof.tex
\subsection{Initialization Algorithms Proposed in Section~\ref{sec:specific-examples}} \label{app:initial-examples}

\begin{algorithm}
\caption{Exact Projection Oracle -- Path Sparse PCA}
\label{alg:projection-PSPCA}
\textbf{Input:} A $(d,k)$-layered graph $G$, a vector $\bm{v} \in \mathbb{R}^d$.
\begin{algorithmic}[1]
\For{$\ell = 1, \ldots, k$}
\State Pick $S_{\ell}$ the index set of the $\ell$-th layer in $G$. 
\State Compute path sparsity vector $\bm{v}^{\tt PS}$ as follows: for its sub-vector $\bm{v}^{\tt PS}_{S_{\ell}}$, set
	\begin{align*}
		~ [\bm{v}^{\tt PS}_{S_{\ell}}]_i := \left\{
		\begin{array}{lll}
			~ [\bm{v}_{S_{\ell}}]_i & \text{ if component $i$ has} \\
			& \text{ the largest absolute value,} \\
			& \text{ breaking ties lexicographically} \\
			~ 0 & \text{ otherwise}
		\end{array}
		\right. .
	\end{align*} 
\EndFor 
\State Normalize $\bm{v}^{\tt PS} := \bm{v}^{\tt PS} / \|\bm{v}^{\tt PS}\|_2$.  
\end{algorithmic}
\textbf{Output:} $\bm{v}^{\tt PS}$.  
\end{algorithm}

%
%
%

\subsection{Proof of Corollaries in Section~\ref{sec:specific-examples}}
We first state an auxilary Lemma that will be used later. 
\begin{lemma}\label{lemma-rho-probability-bound}
 Let $\bm{W}$ be the noise matrix defined in equation~\eqref{eq:noise-def} and $F$ be a fixed linear subspace of dimension $k$. There exists a universal constant $C>0$ such that 
 \begin{align*}
 \Pr\bigg\{ \rho\big(\bm{W},F\big) \geq C(\lambda+1) \sqrt{\frac{k+u}{n}} \bigg\} \leq 2\exp(-u).
 \end{align*} 
\end{lemma}
The proof of Lemma~\ref{lemma-rho-probability-bound} can be found at the end of this section.

\subsubsection{Proof of Corollary~\ref{coro:PS-PCA-fund-limits}} \label{proof-coro-PS-PCA-fund-limits}
\paragraph{Proof of part (a)}
Let $\widehat{P},P_{*}\in \mathcal{P}^k$ be the index sets such that $\mathsf{supp}(\widehat{\bm{v}}_{\mathsf{PS}}) \subseteq \widehat{P}$ and $\mathsf{supp}(\bm{v}_*) \subseteq P_*$. Moreover, let 
\begin{align*}
	& ~ \widehat{L} = \mathsf{span}\big(\{\bm{e}_i\}_{i \in \widehat{P}} \big),\; L_* = \mathsf{span}(\{\bm{e}_i\}_{i \in P_*}), \\
    & ~ \widehat{F} = \mathsf{conv}\big(\widehat{L} \cup L_*\big), \\
    & ~ \mathcal{L} = \bigg\{ \mathsf{span}\big(\{\bm{e}_i\}_{i \in P_1 \cup P_2}\big) \;|\; P_1 \neq P_2 \in \mathcal{P}^{k} \bigg\},
\end{align*}
where $\bm{e}_i$ is the standard base vector with the $i$-th entry being $1$. It follows from part (a) of Theorem~\ref{thm:fund-limits} that
\[
    \big\|\widehat{\bm{v}}_{\mathsf{PS}} - \bm{v}_* \big\|_2 \leq \frac{2\sqrt{2}}{\lambda} \cdot \rho\big(\bm{W}, \widehat{F} \big) \leq \frac{2\sqrt{2}}{\lambda} \cdot \max_{F \in \mathcal{L}}\rho\big(\bm{W}, F \big).
\]
Applying Lemma~\ref{lemma-rho-probability-bound} and union bound, we obtain that there exists a universal constant $C$ for which 
\begin{align*}
    & ~ \Pr\bigg\{\max_{F \in \mathcal{L}}\rho\big(\bm{W}, F \big) \geq C(\lambda+1) \sqrt{\frac{2k+u}{n}} \bigg\}\leq \big|\mathcal{L}\big| \cdot  2\exp(-u) \leq  (d/k)^{2k} \cdot 2\exp(-u),
\end{align*}
where in the last step we use $\big|\mathcal{L}\big| \leq \big|\mathcal{P}^{k} \big|^{2} \leq (d/k)^{2k}$. By setting $u = 2k \cdot \ln(d/k) - \ln(\tau/2)$ and putting the pieces together, we obtain that with probability at least $1-\tau$,
\begin{align*}
    \big\|\widehat{\bm{v}}_{\mathsf{PS}} - \bm{v}_* \big\|_2 &\lesssim \frac{1+\lambda}{\lambda} \cdot \sqrt{\frac{(2+2\ln(d/k))k - \ln(\tau/2)}{n}} \\ &\leq 
    \frac{1+\lambda}{\lambda} \cdot \sqrt{\frac{3(\ln d - \ln k)k - \ln(\tau/2)}{n}}.
\end{align*}
Choosing $\tau = 2\exp(-ck)$ yields the desired result.

\paragraph{Proof of part (b)}
Denote $\mathcal{P}^{k} = \{P_1,...,P_M\}$, where $M = \big|\mathcal{P}^{k} \big|$. For each $m\in [M]$, let $\bm{z}_m \in \mathbb{R}^{d}$ such that
\[
    \bm{z}_{m}(i) = 1 \text{ if }i\in P_m \quad \text{and} \quad \bm{z}_{m}(i) = 0 \text{ if }i\notin P_m.
\]
Note that $1 \in P_m$ for each $m\in M$ since $1$ is index corresponding to the sourse vertex $v_s$. It follows from the definition of $i_*$ and $\mathcal{Z}_*$ in equations~\eqref{definition-i-star}~\eqref{definition-Z-star} that 
\[
i_* = 1 \quad \text{and} \quad \mathcal{Z}_{*} = \{\bm{z}_m\}_{m=1}^{M}.
\]
We next verify Assumption~\ref{assump:minimax-assumption}. Note that $M = (d-2)^{k}/k^{k}$ and it follows from [Lemma 7.4, \cite{asteris2015stay}] that
\begin{align*}
    \big|\mathcal{N}_H(\bm{z}; k/2)\big| \leq \binom{k}{3k/4} d^{k/4}  \leq 2^{2 k} d^{k/4}.
\end{align*}
Consequently, we obtain that
\[
    \frac{|\mathcal{Z}_*|}{\big|\mathcal{N}_H(\bm{z}; k/2)\big|}\geq \frac{(d-2)^k}{k^{k}4^kd^{k/4}} = \frac{d^{3k/4}}{(4k)^{k}} \cdot (1-2/d)^{k} \geq 2^{4},
\]
where the last step follows from the assumption that $d \geq 16k^{2}$ and $k\geq 4$. So we conclude that Assumption~\ref{assump:minimax-assumption} holds by setting $\xi = 3/4$. Continuing, we obtain that
\begin{align*}
    \log \bigg( \frac{|\mathcal{Z}_*|}{ \big|\mathcal{N}_H(\bm{z}; k/2)\big|} \bigg) \geq & ~ \log \bigg( \frac{(d-2)^k}{k^{k}4^kd^{k/4}} \bigg) \\
    \gtrsim & ~ k \big( \log(d)/2 - \log(4k) \big).
\end{align*}
Now, applying part (b) of Theorem~\ref{thm:fund-limits} (by setting $\xi = 3/4$) yiels the desired result.

\subsubsection{Proof of Corollary~\ref{coro:PS-PCA-PPM}}\label{proof-coro-PS-PCA-PPM}
Denote $\mathcal{P}^{k} = \{P_1,...,P_M\}$, where $M = |\mathcal{P}^{k}|$. For each $m \in M$, let $L_m = \mathsf{span}\big(\{\bm{e}_i\}_{i\in P_m}\big)$. Let
\begin{align*}
	& ~ F^{*} = \argmax_{F} \rho(\bm{W},F) \\
	\text{s.t.} ~ & ~ F = \mathsf{conv}(L_{m_1} \cup L_{m_2} \cup L_{m_3}),\;\forall\;m_1 \neq m_2 \neq m_3 \in [M].
\end{align*}
Note that the number of subspaces $F$ such that $F = \mathsf{conv}(L_{m_1} \cup L_{m_2} \cup L_{m_3})$ where $m_1 \neq m_2 \neq m_3 \in [M]$ is bounded by $\big|\mathcal{P}^{k}\big|^{3}$. Applying Lemma~\ref{lemma-rho-probability-bound} and using union bound yield
\begin{align*}
    & ~ \Pr\bigg\{ \rho\big(\bm{W},F^{*}\big) \geq C(\lambda+1) \sqrt{\frac{3k+u}{n}} \bigg\} \leq \big|\mathcal{P}^{k}\big|^{3} \cdot 2\exp(-u) \leq \frac{d^{3k}}{k^{3k}}\cdot 2\exp(-u),
\end{align*}
where the last step follows since $\big|\mathcal{P}^{k}\big| \leq \frac{d^{k}}{k^k}$. By setting $u = (1+\ln2 + 3\ln(d/k))k$, we obtain that with probability at least $1-\exp(-k)$,
\begin{align}\label{bound-error-ps-pca}
    \rho\big(\bm{W},F^{*}\big) \leq & ~ C' (\lambda + 1)\sqrt{ \frac{k(1+\ln d - \ln k)}{n} } \\
    \lesssim & ~  (\lambda + 1)\sqrt{ \frac{k(2\ln d - \ln k)}{n} }. \notag
\end{align}
Letting $n \geq C_{2}k\ln d$ for a large enough constant $C_{2}$, we ontain that
\begin{align*}
    \frac{4}{\lambda + 1 -  \rho\big(\bm{W},F^{*}\big)} + \frac{5 \rho\big(\bm{W},F^{*}\big)}{\lambda - 2 \rho\big(\bm{W},F^{*}\big)}
    \overset{(1)}{\leq} & ~  \frac{8}{\lambda} + \frac{10\rho\big(\bm{W},F^{*}\big)}{\lambda} \\
    \leq & ~ \frac{8}{\lambda} + 20C' \sqrt{\frac{k\ln d}{n}} \leq 0.5,
\end{align*} 
where in step $(1)$ we use $\rho\big(\bm{W},F^{*}\big) \leq \lambda/4$ and in last step we let $\lambda$ be large enough. Consequently, for $\bm{v}_0 \in \mathcal{P}^{k} \cap \mathcal{S}^{d-1}$ and $\langle \bm{v}_0, \bm{v}_*\rangle \geq 0.5$, we obtain that $\bm{v}_0  \in \mathbb{G}(\lambda)$. Similarly, condition~\eqref{eigen-gap-condition} can be verified. Applying Theorem~\ref{thm:convergence} yield
\begin{align*}
	\|\bm{v}_{t+1} - \bm{v}_*\|_2 \leq & ~ \frac{1}{2^{t}} \cdot \|\bm{v}_{0} - \bm{v}_*\|_2 + \frac{6\rho\big(\bm{W},F^{*}\big)}{ \lambda - 2\rho\big(\bm{W},F^{*}\big)} \\
    \leq & ~ \frac{1}{2^{t}} \cdot \|\bm{v}_{0} - \bm{v}_*\|_2 + C_3 \sqrt{ \frac{k(2\ln d - \ln k)}{n} }, 
\end{align*}
where the last inequality holds with probability at least $1-\exp(-k)$ by inequaity~\eqref{bound-error-ps-pca}.

\subsubsection{Proof of Corollary~\ref{coro:TS-PCA-fund-limits}} \label{proof-coro-TS-PCA-fund-limits}
\paragraph{Proof of part (a)}
Let $\widehat{T},T_{*}\in \mathcal{T}^k$ be the index sets such that $\mathsf{supp}(\widehat{\bm{v}}_{\mathsf{TS}}) \subseteq \widehat{T}$ and $\mathsf{supp}(\bm{v}_*) \subseteq T_*$. Moreover, let 
\begin{align*}
	& ~ \widehat{L} = \mathsf{span}\big(\{\bm{e}_i\}_{i \in \widehat{T}} \big),\; L_* = \mathsf{span}(\{\bm{e}_i\}_{i \in T_*}), \\
	& ~ \widehat{F} = \mathsf{conv}\big(\widehat{L} \cup L_*\big)\;\text{and}\; \mathcal{L}^{2k} = \bigg\{ \mathsf{span}\big(\{\bm{e}_i\}_{i \in T}\big) \;|\; T \in \mathcal{T}^{2k} \bigg\},
\end{align*}
where $\bm{e}_i$ is the standard base vector with the $i$-th entry being $1$. It follows from part (a) of Theorem~\ref{thm:fund-limits} that
\[
    \big\|\widehat{\bm{v}}_{\mathsf{TS}} - \bm{v}_* \big\|_2 \leq \frac{2\sqrt{2}}{\lambda} \cdot \rho\big(\bm{W}, \widehat{F} \big) \leq \frac{2\sqrt{2}}{\lambda} \cdot \max_{F \in \mathcal{L}^{2k}}\rho\big(\bm{W}, F \big).
\]
Applying Lemma~\ref{lemma-rho-probability-bound} and union bound, we obtain that there exists a universal constant $C$ for which 
\begin{align*}
    & ~ \Pr\bigg\{\max_{F \in \mathcal{L}^{2k}}\rho\big(\bm{W}, F \big) \geq C(\lambda+1) \sqrt{\frac{k+u}{n}} \bigg\} \leq \big|\mathcal{L}^{2k}\big| \cdot  2\exp(-u) \leq  \frac{(2e)^{2k}}{2k + 1} \cdot 2\exp(-u),
\end{align*}
where in the last step we use $\big|\mathcal{L}^{2k}\big| = \big|\mathcal{T}^{2k} \big| \leq \frac{(2e)^{2k}}{2k + 1}$. By setting $u = 2(1+\ln2)k - \ln(\tau/2)$ and putting the pieces together, we obtain that with probability at least $1-\tau$,
\[
    \big\|\widehat{\bm{v}}_{\mathsf{TS}} - \bm{v}_* \big\|_2 \lesssim \frac{1+\lambda}{\lambda} \cdot \sqrt{\frac{(3+2\ln2)k - \ln(\tau/2)}{n}}.
\]
Choosing $\tau = 2\exp(-ck)$ yields the desired result.
\paragraph{Proof of part (b)}
Denote $\mathcal{T}^{k} = \{T_1,...,T_M\}$, where $M = \big|\mathcal{T}^{k} \big|$. For each $m\in [M]$, let $\bm{z}_m \in \mathbb{R}^{d}$ such that
\[
    \bm{z}_{m}(i) = 1 \text{ if }i\in T_m \quad \text{and} \quad \bm{z}_{m}(i) = 0 \text{ if }i\notin T_m.
\]
Note that $1 \in T_m$ for each $m\in M$. It follows from the definition of $i_*$ and $\mathcal{Z}_*$ in equations~\eqref{definition-i-star}~\eqref{definition-Z-star} that 
\[
i_* = 1 \quad \text{and} \quad \mathcal{Z}_{*} = \{\bm{z}_m\}_{m=1}^{M}.
\]
Let us first lower bound the size of $\mathcal{Z}_{*}$. Note that, for a fixed $k$, when $\mathsf{CBT}$ is of height greater than or equal to $k$, any nodes with height $\geq k$ will never be selected. That is to say, it is sufficient to consider $d \leq 2^{k} - 1$. Based on the proof of [Proposition 1, \cite{baraniuk2010model}], when $k \geq \log_2 d$, the total number of subtrees with size $k$ in $\mathsf{CBT}$ can be represented as
\begin{align*}
	M = |\mathcal{T}^k| = & ~ \sum_{h = \lfloor \log_2 k \rfloor + 1}^{\log_2 d} t_{k, h}. 
\end{align*}
Here $t_{k, h}$ denotes the number of binary subtrees with size $k$ and height $h$, which can be lower bounded by 
\begin{align*}
	& ~ t_{k, h} \\
	\geq & ~ \frac{4^{k + 1.5}}{h^4} \sum_{m \geq 1} \left[ \frac{2k}{h^2} (2\pi m)^4 - 3 (2 \pi m)^2 \right] \exp\left( - \frac{k(2 \pi m)^2}{h^2}\right) \\
	\geq & ~ \frac{4^{k + 1.5}}{h^4} \sum_{m \geq h / (\sqrt{2 \pi^2 k})} \left[ \frac{2k}{h^2} (2\pi m)^4 - 3 (2 \pi m)^2 \right] \exp\left( - \frac{k(2 \pi m)^2}{h^2}\right) \\
	\geq & ~ \frac{4^{k + 1.5}}{h^4} \sum_{m \geq h / (\sqrt{2 \pi^2 k})} (2 \pi m)^2 \exp\left( - \frac{k(2 \pi m)^2}{h^2}\right) \\
	\geq & ~ \frac{4^{k + 1.5}}{h^4} \int_{m \geq h / (\sqrt{2 \pi^2 k})} (2 \pi m)^2 \exp\left( - \frac{k(2 \pi m)^2}{h^2}\right) \mathrm{d}m
\end{align*}
where the third inequality holds since $\frac{2k}{h^2} (2 \pi m)^4  \geq 4 (2 \pi m)^2$ if $m \geq \frac{h}{\sqrt{2 \pi^2 k}}$, and the forth inequality (the infinity summation is greater than or equal to the integration) holds since the function $f(m) := (2 \pi m)^2 \exp\left( - \frac{k(2 \pi m)^2}{h^2}\right)$ becomes monotone decreasing when $m \geq \frac{h}{\sqrt{4 \pi^2 k}}$. Therefore, by integration, we have 
\begin{align*}
	t_{k, h} 
	\geq & ~ \frac{4^{k + 1.5}}{h^4} \left[ \frac{\sqrt{\pi} h^3}{4 k^{3/2}}\text{erfc}\left( \frac{1}{\sqrt{2 \pi^2}} \right) + \frac{h^3}{\sqrt{8 \pi^2} k^{3/2}} \exp\left( - \frac{1}{2 \pi^2} \right) \right] \\
	= & ~ \frac{4^{k + 1.5}}{k^{3/2}} \cdot \underbrace{ \left[ \frac{\sqrt{\pi}}{4}\text{erfc}\left( \frac{1}{\sqrt{2 \pi^2}} \right) + \frac{1}{\sqrt{8 \pi^2}} \exp\left( - \frac{1}{2 \pi^2} \right) \right]}_{=: c_1 > 0} \cdot \frac{1}{h},
\end{align*}
where $\text{erfc}(x) := 1 - \frac{2}{\sqrt{\pi}} \int_{0}^{x} \exp(- t^2) \mathrm{d}t \in (0, 1)$ denotes the complementary error function. Inserting the above result into the formulation for total number of subtrees gives
\begin{align*}
	M = \sum_{h = \lfloor \log_2 k \rfloor + 1}^{\log_2 d} t_{k, h} \geq & ~ \frac{4^{k + 1.5}}{k^{3/2}} \cdot \sum_{h = \lfloor \log_2 k \rfloor + 1}^{\log_2 d} \frac{c_1}{h} \\
	\geq & ~ \frac{4^{k + 1.5}}{k^{3/2}} \cdot c_1 \cdot \ln \left( \frac{\log_2 d}{\lfloor \log_2 k \rfloor + 1} \right). 
\end{align*}

By picking $i_* = 1$, since all the subtree are rooted at the root node $r_{\mathsf{CBT}}$ with index $1$, then we have $M = |\mathcal{Z}_{*}|$. Given any $\xi \in (3/4, 1)$ and any characteristic vector $\bm{z} \in \mathcal{Z}$, the size of neighborhood $\mathcal{N}_H(\bm{z}; 2(1 - \xi)k)$ can be upper bounded by
\begin{align*}
	|\mathcal{N}_H(\bm{z}; 2(1 - \xi)k)| = & ~ \sum_{i = 0}^{(1 - \xi)k} |\{\bm{z}' ~|~ \delta_H(\bm{z}, \bm{z}') = i\}| \\
	\leq & ~ \sum_{i = 0}^{(1 - \xi)k} \binom{k}{i} k^i \leq 2^{2 (1 - \xi) k \cdot \log_2 k},
\end{align*}
where the first inequality holds since $|\{\bm{z}' ~|~ \delta_H(\bm{z}, \bm{z}') = i\}|$ can be upper bounded by the product between the upper bound of all sub-subtrees in $T(\bm{z})$ with size $k - i$ and root node $r_{\mathsf{CBT}}$ (i.e., $\binom{k}{i}$) and the upper bound of all possible choices for expanding a subtree with size $k - i$ to size $k$ (i.e., $k^i$). When $2(1 - \xi) k \leq k / \log_2 k$, we have 
\begin{align*}
	|\mathcal{N}_H(\bm{z}; 2(1 - \xi)k)| \leq 2^{2 \log_2 k (1 - \xi) k} \leq 2^{k}
\end{align*}
and therefore, the Assumption~\ref{assump:minimax-assumption}.2 
\begin{align*}
	\frac{|\mathcal{Z}_{*}|}{|\mathcal{N}_H(\bm{z}; 2(1 - \xi)k)|} \geq \frac{c_1 4^{k + 1.5}}{k^{3/2} \cdot 2^{k}} \cdot \ln \left( \frac{\log_2 d}{\lfloor \log_2 k \rfloor + 1} \right) \geq 16
\end{align*} 
holds with $k$ sufficiently greater than some constant. To achieve the above result, by setting $\xi = 1 - \frac{1}{2 \log_2 k}$, using results in part (b) of Theorem~\ref{thm:fund-limits}, we have
\begin{align*}
    & ~ \inf_{\widehat{\bm{v}}} \; \sup_{\bm{v}_* \in \mathcal{S}^{d - 1} \cap \mathcal{M}} \mathbb{E} \left[ \left\| \widehat{\bm{v}} \widehat{\bm{v}}^{\top} - \bm{v}_* \bm{v}_*^{\top}  \right\|_F \right]  \notag\\
    \geq & ~ \frac{\sqrt{2(1 - \xi)}}{4} \min\Bigg\{1, ~ \sqrt{\frac{1 + \lambda}{8 \lambda^2}} \sqrt{\frac{ \log \left(|\mathcal{Z}_{*}|\right) - \log \big( \max_{\bm{z} \in \mathcal{Z}_{*}}|\mathcal{N}_H(\bm{z}; 2(1 - \xi) k)| \big)}{n}} \Bigg\} \\
    = & ~ \frac{1}{4\sqrt{\log k}} \min \Bigg\{1,  ~ \sqrt{\frac{1 + \lambda}{8 \lambda^2}} \sqrt{\frac{k + 3 + \log c_1 - \frac{3}{2} \log k - \log \ln \left( \frac{\log_2 d}{\lfloor \log_2 k \rfloor + 1} \right) }{n}} \Bigg\} 
\end{align*}
As mentioned before, it is sufficient to consider $d \leq 2^{k} - 1$, then the term $\ln \left( \frac{\log_2 d}{\lfloor \log_2 k \rfloor + 1} \right) \leq \ln k$, and thus
\begin{align*}
	\inf_{\widehat{\bm{v}}} \; \sup_{\bm{v}_* \in \mathcal{S}^{d - 1} \cap \mathcal{M}} \mathbb{E} \left[ \left\| \widehat{\bm{v}} \widehat{\bm{v}}^{\top} - \bm{v}_* \bm{v}_*^{\top}  \right\|_F \right]
	\gtrsim & ~ \min \Bigg\{ \frac{1}{4 \sqrt{\log k}}, ~~ \frac{1}{4}\sqrt{\frac{1 + \lambda}{8 \lambda^2}} \sqrt{\frac{k/\log k}{n}} \Bigg\}. 
\end{align*}

\subsubsection{Proof of Corollary~\ref{coro:TS-PCA-PPM}}\label{proof-coro-TS-PCA-PPM}
Denote $\mathcal{T}^{k} = \{T_1,...,T_M\}$, where $M = |\mathcal{T}^{k}|$. For each $m \in M$, let $L_m = \mathsf{span}\big(\{\bm{e}_i\}_{i\in T_m}\big)$. Let
\begin{align*}
	& ~ F^{*} = \argmax_{F} \rho(\bm{W},F) \\
	\text{s.t.} & ~ F = \mathsf{conv}(L_{m_1} \cup L_{m_2} \cup L_{m_3}),\;\forall\;m_1 \neq m_2 \neq m_3 \in [M]. 
\end{align*}
Note that the number of subspaces $F$ such that $F = \mathsf{conv}(L_{m_1} \cup L_{m_2} \cup L_{m_3})$ where $m_1 \neq m_2 \neq m_3 \in [M]$ is bounded by $\big|\mathcal{T}^{3k}\big|$. Applying Lemma~\ref{lemma-rho-probability-bound} and using union bound yield
\begin{align*}
    \Pr\bigg\{ \rho\big(\bm{W},F^{*}\big) \geq C(\lambda+1) \sqrt{\frac{3k+u}{n}} \bigg\}
    \leq & ~ \big|\mathcal{T}^{3k}\big| \cdot 2\exp(-u) \leq \frac{(2e)^{3k}}{3k + 1}\cdot 2\exp(-u),
\end{align*}
where the last step follows since $\big|\mathcal{T}^{3k}\big| \leq \frac{(2e)^{3k}}{3k + 1}$. By setting $u = 4(1+\ln2)k$, we obtain that with probability at least $1-\exp(-k)$,
\begin{align}\label{bound-error-ts-pca}
    \rho\big(\bm{W},F^{*}\big) \leq C' (\lambda + 1)\sqrt{ \frac{k}{n} } \leq 2C' \lambda \sqrt{\frac{k}{n}},
\end{align}
where the last step follows for $\lambda \geq 1$. For $n \geq (50C')^2k$, we ontain that
\begin{align*}
    \frac{4}{\lambda + 1 -  \rho\big(\bm{W},F^{*}\big)} + \frac{5 \rho\big(\bm{W},F^{*}\big)}{\lambda - 2 \rho\big(\bm{W},F^{*}\big)} 
    \overset{(1)}{\leq} & ~ \frac{8}{\lambda} + \frac{10\rho\big(\bm{W},F^{*}\big)}{\lambda} \leq \frac{8}{\lambda} + 20C' \sqrt{\frac{k}{n}} \leq 0.5,
\end{align*} 
where in step $(1)$ we use $\rho\big(\bm{W},F^{*}\big) \leq \lambda/4$ and in last step we let $\lambda$ be large enough. Consequently, for $\bm{v}_0 \in \mathcal{T}^{k} \cap \mathcal{S}^{d-1}$ and $\langle \bm{v}_0, \bm{v}_*\rangle \geq 0.5$, we obtain that $\bm{v}_0  \in \mathbb{G}(\lambda)$. Similarly, condition~\eqref{eigen-gap-condition} can be verified. Applying Theorem~\ref{thm:convergence} yield
\begin{align*}
	\|\bm{v}_{t+1} - \bm{v}_*\|_2 \leq & ~ \frac{1}{2^{t}} \cdot \|\bm{v}_{0} - \bm{v}_*\|_2 + \frac{6\rho\big(\bm{W},F^{*}\big)}{ \lambda - 2\rho\big(\bm{W},F^{*}\big)} \\
	\leq & ~ \frac{1}{2^{t}} \cdot \|\bm{v}_{0} - \bm{v}_*\|_2 + C_3 \sqrt{\frac{k}{n}},
\end{align*}
where the last inequality holds with probability at least $1-\exp(-k)$ by inequaity~\eqref{bound-error-ts-pca}.

\subsubsection{Proof of Lemma~\ref{lemma-rho-probability-bound}}

\paragraph{Proof of Lemma~\ref{lemma-rho-probability-bound}:} Without loss of generality, let $\{\phi_1,...,\phi_k\}$ be an orthonormal basis of $F$. Let $\Phi = [\phi_1\;|\cdots|\phi_k] \in \mathbb{R}^{d \times k}$. By definition, we obtain that
\begin{align*}
    \rho\big(\bm{W},F\big) = & ~ \max_{\bm{v}\in \mathcal{S}^{d-1} \cap F} \bm{v}^{\top} \bm{W} \bm{v} \\
    = & ~ \max_{\bm{u} \in \mathcal{S}^{k-1}} \bm{u}^{\top} \Phi^{\top} \bm{W} \Phi \bm{u} = \big\|\Phi^{\top} \bm{W} \Phi\big\|.
\end{align*}
Continuing, we obtain that
\begin{align*}
\big\|\Phi^{\top} \bm{W} \Phi\big\| = \bigg\| \frac{1}{n}\sum_{i=1}^{n} (\Phi^{\top}\bm{x}_{i}) (\Phi^{\top}\bm{x}_{i})^{\top} - \Phi^{\top} \bm{\Sigma} \Phi \bigg\|.
\end{align*}
Note that $\Phi^{\top}\bm{x}_{i} \sim \mathcal{N}(\bm{0},\Phi^{\top} \bm{\Sigma} \Phi)$. Applying the result in [Exercise 4.7.3 (Tail bound), \cite{vershynin2018high}] yields that
\begin{align*}
    \Pr\bigg\{ \big\|\Phi^{\top} \bm{W} \Phi\big\| \leq C \|\Phi^{\top} \bm{\Sigma} \Phi\| \cdot  \sqrt{\frac{k+u}{n}} \bigg\}  \geq 1 - 2\exp(-u).
\end{align*}
Note that $\|\Phi^{\top} \bm{\Sigma} \Phi\| \leq \lambda+1$. Putting the pieces together yields the desired result.

\subsubsection{Proof of inequality~\eqref{eq:SPCA-known}}\label{min-max-pca-proof}
In the setting of sparse PCA, the number of possible linear subspaces is $M = \binom{d}{k}$. By definitions of $\mathcal{Z}_*$ and $i_*$, we obtain that
\begin{align*}
	|\mathcal{Z}_*| = & ~ \sum_{m=1}^{M} \bm{z}_m(i_*) \geq \frac{\sum_{i=1}^{d} \sum_{m=1}^{M} \bm{z}_m(i) }{d} \\
	= & ~ \frac{ \sum_{m=1}^{M} \sum_{i=1}^{d} \bm{z}_m(i) }{d} = \frac{Mk}{d} \geq \frac{M}{d}.
\end{align*}
The neighborhood $\mathcal{N}_H(\bm{z}; k/2)$ with a common support index $i_*$ satisfies 
\begin{align*}
    \big|\mathcal{N}_H(\bm{z}; k/2) \big| = & ~ \sum_{i = 0}^{k/2} \binom{k - 1}{i} \binom{d - k}{i} \leq [k(d - k)]^{k/2}.
\end{align*}
Continuing, we obtain that
\begin{align*}
    & ~ \frac{|\mathcal{Z}_*|}{\max_{\bm{z} \in \mathcal{Z}_*}|\mathcal{N}_H(\bm{z}; k/2)|}\\
    \geq & ~ \frac{M}{d \cdot \max_{\bm{z} \in \mathcal{Z}_*}|\mathcal{N}_H(\bm{z}; k/2)|} \\
    \geq & ~ \frac{\binom{d}{k}}{d \cdot [k(d - k)]^{k/2}} \geq \frac{d^k}{k^k} \frac{1}{k^{k/2}d^{k/2}d}  = \bigg( \frac{d}{k^{3}}\bigg)^{k/2} \cdot \frac{1}{d}.
\end{align*}
Consequently,
\[
    \log \bigg( \frac{|\mathcal{Z}_*|}{\max_{\bm{z} \in \mathcal{Z}_*}|\mathcal{N}_H(\bm{z}; k/2)|} \bigg) \gtrsim k \log d.
\]
Applying Theorem~\ref{thm:fund-limits}(b) by setting $\xi = 3/4$ yields the desired result.


\subsection{Proof of Propositions in Section~\ref{sec:specific-examples}}

\subsubsection{Proof of Proposition~\ref{prop:reduction-PathPCA}} \label{app:reduction-PathPCA}

\begin{proof}
For the proof, one should think of the estimator and parameters as indexed by the natural number $n$, although we drop this explicit dependence for clarity of exposition.
We begin by introducing formal definition for the detection problem for path sparse PCA. Recall the definition of a qualified estimator (Definition~\ref{def:qual-est}).

\begin{definition}
\textbf{Detection problem for path sparse PCA.} Suppose $\bm{v}_*$ is $\bm{0}_d$ with probability $1/2$ and an arbitrary vector in the set $\mathcal{S}^{d - 1} \cap \mathcal{P}^k$ with probability $1/2$. The detection problem for path sparse PCA -- $\text{D-PSPCA}(n, k, d, \lambda)$ is defined as the resulting hypothesis testing problem 
\begin{align*}
	H_0: \bm{X} \sim \mathcal{D}(0; \bm{0}_d)^{\otimes n} ~~~~\text{and}~~~~  H_1: \bm{X} \sim \mathcal{D}(\lambda; \bm{v}_*)^{\otimes n}.
\end{align*}
\end{definition}


We show our average-case hardness result by contradiction, i.e., we assume there exists a randomized polynomial-time qualified estimator. We then transform it into a good detector for the path-sparse PCA problem and eventually the secret leakage planted clique problem. The proof of Proposition~\ref{prop:reduction-PathPCA} can be separated into four parts: 

\begin{enumerate}
	\item \textbf{Constructing a randomized polynomial time algorithm for detection based on an estimation algorithm for path sparse PCA.} 
	
	Suppose $\EST_n: (\mathbb{R}^{d_n})^n \to \mathbb{R}^{d_n}$ is a sequence of randomized polynomial time functions for the assumed qualified estimator. 
	

	Given $\EST_n$, we construct a detector $\DET_n: \EST \times (\mathbb{R}^{d_n})^n \times \mathbb{R} \to \{0,1\}$ of $\text{D-PSPCA}(n, k, d, \lambda)$, which maps a tuple of three components including an estimator function $\EST_n$, an instance $\bm{X}$ and a known eigengap $\lambda$ to  a detection algorithm that outputs one of the two hypotheses; this is presented in Algorithm~\ref{alg:D-from-E}. 
	

\begin{algorithm}
\caption{Detector $\DET_n$ of path sparse PCA}
\label{alg:D-from-E}
\textbf{Input:} A function $\EST_n$, an instance $\bm{X}$ drawn from $\text{D-PSPCA}(n,k,d,\lambda)$, and an eigengap $\lambda$.  
\begin{algorithmic}[1]
\State Compute the estimation $\widehat{\bm{v}} := \EST_n(\bm{X})$ of path sparse PCA. \label{alg:D-from-E-1}
\State Normalize the estimation 
	\begin{align*}
		\tilde{\bm{v}} := \left\{
		\begin{array}{lll}
			\widehat{\bm{v}} / \|\widehat{\bm{v}}\|_2 & \text{ if } \widehat{\bm{v}} \neq \bm{0}_{d_n} \\
			\text{any point in $\mathcal{S}^{d_n} \cap \mathcal{P}^k$} & \text{ if } \widehat{\bm{v}} = \bm{0}_{d_n}
		\end{array}
		\right. .
	\end{align*} \label{alg:D-from-E-2}
\State Set the sample covariance matrix $\widehat{\bm{\Sigma}} := \frac{1}{n} \bm{X} \bm{X}^{\top}$. \label{alg:D-from-E-3}
\If{$\tilde{\bm{v}}^{\top} \widehat{\bm{\Sigma}} \tilde{\bm{v}} \geq 1 + \lambda/4 - (1 + \lambda) \sqrt{k \ln d / n}$} \label{alg:D-from-E-4}
\State \textbf{Output:} $1$, i.e., $\bm{X}$ is from $H_1$.
\Else 
\State \textbf{Output:} $0$, i.e., $\bm{X}$ is from $H_0$.
\EndIf
\end{algorithmic}
\end{algorithm} 

	The detector $\DET_n$ clearly runs in polynomial time. Next, we show that for any instance $\bm{X}$ of $\text{D-PSPCA}(n, k, d, \lambda)$ and a known eigengap $\lambda$ (satisfying the conditions required in Theorem~\ref{thm:convergence} and Theorem~\ref{thm:initialization-method}), $\DET_n$ satifies
	\begin{align} \label{eq:key-claim}
		\liminf_{n \rightarrow \infty} \left( \mathbb{P}_{H_0}[\DET_n(\bm{X}) = 1] + \mathbb{P}_{H_1}[\DET_n(\bm{X}) = 0] \right) < 1. 
	\end{align}

		
		
		To establish claim~\eqref{eq:key-claim}, note that the ``if criteria'' (Step-\eqref{alg:D-from-E-3} in Algorithm~\ref{alg:D-from-E}) is based on the following two-part calculation. 
 On the one hand, suppose the instance $\bm{X}$ is drawn from $\mathcal{D}(\lambda; \bm{v}_*)^{\otimes n}$ for $\bm{v}_* \in \mathcal{S}^{d - 1} \cap \mathcal{P}^k$. Consider the orthogonal decomposition of $\tilde{\bm{v}}$ given by $\tilde{\bm{v}} = \alpha \bm{v}_* + \beta \bm{v}_{\perp}$ with $\|\bm{v}_*\|_2 = \|\bm{v}_{\perp}\|_2 = 1, ~ \langle \bm{v}_*, \bm{v}_{\perp} \rangle = 0, \alpha^2 + \beta^2 = 1$. Based on the definition of the qualified estimator, the normalized vector $\|\tilde{\bm{v}} - \bm{v}_*\|_2$ satisfies $\| \widehat{\bm{v}} / \|\widehat{\bm{v}} \|_2 - \bm{v}_* \|_2 < \sqrt{2 - 2\sqrt{15/16}} < 1/2$. Consequently, we have $\alpha > 1/2$, and therefore the objective satisfies
		\begin{align*}
			\tilde{\bm{v}}^{\top} \widehat{\bm{\Sigma}} \tilde{\bm{v}} = & ~ (\alpha \bm{v}_* + \beta \bm{v}_{\perp})^{\top} \widehat{\bm{\Sigma}}  (\alpha \bm{v}_* + \beta \bm{v}_{\perp}) \\
			= & ~ \alpha^2 \bm{v}_*^{\top} \bm{\Sigma} \bm{v}_* + \beta^2 \bm{v}_{\perp}^{\top} \bm{\Sigma} \bm{v}_{\perp} + 2 \alpha \beta \bm{v}_{*}^{\top} \bm{\Sigma} \bm{v}_{\perp} + \widehat{\bm{v}}^{\top} \bm{W} \widehat{\bm{v}} \\
			\geq & ~ \alpha^2 (1 + \lambda) + \beta^2 + 0 + \rho(\bm{W}, P^*) \\
			\geq & ~ 1 + \frac{\lambda}{4} - (1 + \lambda) \sqrt{k \ln d / n}, 
		\end{align*}
		where the final inequality holds due to $\rho(\bm{W}, P^*) \leq (1 + \lambda) \sqrt{k \ln d / n}$ with probability $1 - \exp(- ck)$.  
		On the other hand, if the instance $\bm{X}$ is drawn from $\mathcal{D}(0; \bm{0}_d)^{\otimes n}$, then the objective satisfies
		\begin{align*}
			\tilde{\bm{v}}^{\top} \widehat{\bm{\Sigma}} \tilde{\bm{v}} = & ~ \tilde{\bm{v}}^{\top} (\bm{I}_d + \bm{W}) \tilde{\bm{v}} \\
			\leq & ~ 1 + \rho(\bm{W}, P^*) \\
			\leq & ~ 1 + (1 + \lambda) \sqrt{k \ln d / n} \\
			< & ~ 1 + \frac{\lambda}{4} - (1 + \lambda) \sqrt{k \ln d / n},
		\end{align*}
		where the final strict inequality holds in the parameter regime of interest. Assuming that $n$ is sufficiently large and combining the above two cases implies 
		\begin{align*}
			\mathbb{P}_{H_0}[\DET_n(\bm{X}) = 1] \leq & ~ \frac{1}{2} - \epsilon + \exp(- ck), \\
			\mathbb{P}_{H_1}[\DET_n(\bm{X}) = 0] \leq & ~ \frac{1}{2} - \epsilon + \exp(- ck), 
		\end{align*} 
		and therefore, 
		\begin{align*}
			\liminf_{n \rightarrow \infty} \left( \mathbb{P}_{H_0}[\DET_n(\bm{X}) = 1] + \mathbb{P}_{H_1}[\DET_n(\bm{X}) = 0] \right) < 1. 
		\end{align*}
	
	\item \textbf{Reduction from $K$-Partite PC detection problem to path sparse PCA detection problem.} 
		
		
		In this step, we use the average-case reduction method -- SPCA-RECOVERY [Figure 19, \citep{brennan2018reducibility}] to be our reduction method. Recall \cite{brennan2018reducibility} show that SPCA-RECOVERY maps an instance of planted clique problem $G \sim \text{PC}(n,k,1/2)$ to an instance $\bm{X}$ of vanilla sparse PCA detection problem (with ground truth $\bm{v}_*$ takes values in discrete set $\{0, 1 / \sqrt{k}\}$) approximately under total variance distance. Note that SPCA-RECOVERY ensures a preserving property (See Remark~\ref{remark:str-pre} for details). In words, SPCA-RECOVERY maps an instance $G \sim \text{K-PC}(n,k,1/2)$ to an instance $\bm{X} \sim \text{D-PSPCA}(n, k, d, \lambda)$ approximately under total variance distance while maintaining the structure, i.e., mapping rows associated with planted $k$-clique to rows associated with the corresponding path.

		\begin{remark} \label{remark:str-pre}
		\textbf{Structure preserving property of \textup{SPCA-RECOVERY}.} Given any $G \sim K\text{-PC}(n, k, 1/2)$ as an input instance of \textup{SPCA-RECOVERY}, \textup{SPCA-RECOVERY} maps the rows in the adjacency matrix $\bm{A}(G)$ concerning the planted $k$-clique of $G$ to the rows corresponding to the support set of the underlying path structure for the sample matrix (instance) $\bm{X} = [\bm{x}_1 ~|~ \cdots ~|~ \bm{x}_n]^{\top} \sim \text{D-PSPCA}(n, k, d, \lambda)$. 
		\end{remark} 
		

		Armed with these tools, we now present a quantitative analysis of the total variance distance that arises from our reductions. Let us start by recalling the parameter regime with some additional notation as follows. Let $\beta \geq 1/2$. For path sparse PCA detection problem, define the following sequence of parameters 
		\begin{align*}
			& ~ k_{\subindex} = \lceil \subindex^{\beta} \rceil, \quad \rho_{\subindex} = \frac{1}{2}, \quad n_{\subindex} = d_{\subindex} = \subindex, \quad \mu_{\subindex} = \frac{\log 2}{2\sqrt{6 \log \subindex + 2 \log 2}}, \quad \lambda_{\subindex} = \frac{k_{\subindex}^2}{\tau_{\subindex} \subindex} \cdot \frac{(\log 2)^2}{4 (6 \log \subindex + 2 \log 2)},
		\end{align*}
		where $\tau_{\subindex} \rightarrow \infty$ as ${\subindex} \rightarrow \infty$, i.e., an arbitrarily slowly growing function of $\subindex$. In this part 2), to be clear, we use $\subindex \in \mathbb{Z}$ to denote the integer index of the above sequence of parameters, and $n_{\subindex} ~ (= \subindex)$ to denote the sample size corresponding with index $\subindex$. Later in this proof, we do not distinguish the difference between sample size $n, n_{\subindex}$ and index $z$. Let $\varphi_{\subindex} = \text{SPCA-RECOVERY}$ be the reduction method. We use graph $G_{\subindex}$ to denote an instance of $\text{K-PC}(n_{\subindex},k_{\subindex},1/2)$, and use $\bm{X}_{\subindex} = \varphi_{\subindex} (G_{\subindex})$ to be the output of $\text{SPCA-RECOVERY}$ with input $G_{\subindex}$. To be concise, we use $\mathcal{L}(\bm{X}_{\subindex})$ to denote the distribution of a given instance $\bm{X}_{\subindex}$.

		Suppose $G_{\subindex} \sim \mathcal{G}_{\mathcal{D}}(\subindex, k_{\subindex}, 1/2)$, is drawn from the $H_1$ hypothesis of $\text{K-PC}(\subindex,k_{\subindex}, 1/2)$. Let $\bm{v}_*$ denote the unit vector supported on indices with respect to the clique in $G_{\subindex}$ with nonzero entries equal to $1/\sqrt{k_{\subindex}}$. Using\footnote{It is easy to observe that our parameter regime of $(\subindex, \mu_{\subindex}, \rho_{\subindex}, \tau(\subindex))$ satisfies the conditions presented in Lemma 6.7 and Lemma 8.2 in \cite{brennan2018reducibility}.} Lemma 6.7 and Lemma 8.2 in \cite{brennan2018reducibility}, we have
		\begin{align*}
			\text{d}_{\text{TV}} \left( \mathcal{L}(\bm{X}_{\subindex}), \mathcal{D}(\lambda; \bm{v}_*) \right) \leq O\left( \frac{1}{\sqrt{\log {\subindex}}} \right) + \frac{2 (\subindex + 3)}{\tau \subindex - \subindex - 3} \rightarrow 0 
		\end{align*}
		as $\subindex \rightarrow \infty$. On the other hand, if an instance $G_{\subindex} \sim \mathcal{G}_{\mathcal{D}}(\subindex, 1/2)$, is drawn from the $H_0$ hypothesis of $\text{K-PC}(\subindex,k_{\subindex},1/2)$. Still using Lemma 6.7 and Lemma 8.2 in \cite{brennan2018reducibility}, we also have 
		\begin{align*}
			\text{d}_{\text{TV}} \left( \mathcal{L}(\bm{X}_{\subindex}), \mathcal{D}(0; \bm{0}_d) \right) \leq O\left( \frac{1}{\sqrt{\log \subindex}} \right) + \frac{2 (\subindex + 3)}{\tau \subindex - \subindex - 3} \rightarrow 0 
		\end{align*}
		as $\subindex \rightarrow \infty$. Combining the above two cases ensures the reduction from $K$-Partite PC detection problem to path sparse PCA detection problem as we desired.

		
	\item \textbf{Constructing a detection algorithm for $K$-Partite PC based on a detection algorithm for path sparse PCA.}

		Recall that there exists a sequence of detector algorithms $\DET_n$ that solves the detection problem of path-sparse PCA ($\text{D-PSPCA}(n, k, d, \lambda)$) as described above. Observe that under $H_1$ hypothesis,
		\begin{align*}
			& ~ \left|\mathbb{P}_{\bm{X} \sim \mathcal{L}(\bm{X}_n)}[\DET_n(\bm{X}) = 1] - \mathbb{P}_{\bm{X} \sim \mathcal{D}(\lambda; \bm{v}_*)}[\DET_n(\bm{X}) = 1] \right| \\
			\leq & ~ \text{d}_{\text{TV}} \left( \mathcal{L}(\bm{X}_n), \mathcal{D}(\lambda; \bm{v}_*) \right) \rightarrow 0 \quad \text{as } n \rightarrow \infty,  
		\end{align*}
		where the limitation of the final inequality holds due to the analysis in part 2) with $n_{\subindex} = z$. Since 
		\begin{align*}
			& ~ \mathbb{P}_{\bm{X} \sim \mathcal{D}(\lambda; \bm{v}_*)}[\DET_n(\bm{X}) = 1] = \mathbb{P}_{H_1}[\DET_n(\bm{X}) = 1] \geq \frac{1}{2} + \epsilon
		\end{align*} 
		for sufficiently large $n$ by the definition of the detector $\DET_n$, it follows that 
		\begin{align*}
			\mathbb{P}_{\bm{X} \sim \mathcal{L}(\bm{X}_n)}[\DET_n \circ \varphi_n(G_n) = 1] 
			= & ~ \mathbb{P}_{\bm{X} \sim \mathcal{L}(\bm{X}_n)}[\DET_n (\bm{X}_n) = 1] \geq \frac{1}{2} + \frac{\epsilon}{2} 
		\end{align*}
		for sufficiently large $n$. On the other hand, under $H_0$ hypothesis, 
		\begin{align*}
			& ~ \big|\mathbb{P}_{\bm{X} \sim \mathcal{L}(\bm{X}_n)}[\DET_n(\bm{X}) = 0] - \mathbb{P}_{\bm{X} \sim \mathcal{D}(0; \bm{0}_d)}[\DET_n(\bm{X}) = 0] \big| \\
			\leq & ~ \text{d}_{\text{TV}} \left( \mathcal{L}(\bm{X}_n), \mathcal{D}(0; \bm{0}_d) \right) \rightarrow 0 \quad \text{as } n \rightarrow \infty. 
		\end{align*}
		Still 
		\begin{align*}
			& ~ \mathbb{P}_{\bm{X} \sim \mathcal{D}(0; \bm{0}_d)}[\DET_n(\bm{X}) = 0] = \mathbb{P}_{H_0}[\DET_n(\bm{X}) = 0] \geq \frac{1}{2} + \epsilon
		\end{align*} 
		holds for sufficiently large $n$ by the definition of the detector $\DET_n$. Then 
		\begin{align*}
			& ~ \mathbb{P}_{\bm{X} \sim \mathcal{L}(\bm{X}_n)}[\DET_n \circ \varphi_n(G_n) = 0] = \mathbb{P}_{\bm{X} \sim \mathcal{L}(\bm{X}_n)}[\DET_n (\bm{X}_n) = 0] \geq \frac{1}{2} + \frac{\epsilon}{2}. 
		\end{align*}
		Combining the above two results together implies
		\begin{align*}
			\liminf_{n \rightarrow \infty} & ~ \bigg( \mathbb{P}_{G_n \sim \mathcal{G}(n, 1/2)}[\DET_n \circ \varphi_n(G_n) = 1] + \mathbb{P}_{G_n \sim \mathcal{G}(n, k_n, 1/2)}[\DET_n \circ \varphi_n(G_n) = 0] \bigg) \leq 1 - \epsilon, 
		\end{align*}
		i.e., the detection method works.

	\item \textbf{Putting together the pieces.}

		Putting the above three parts together, we have used a polynomial-time qualified estimator to construct a sequence of functions $\DET_n \circ \varphi_n$ that detects K-Partite PC (Definition~\ref{def:KPC}) in polynomial time. This contradicts the K-Partite PC conjecture (Conjecture~\ref{conj:KPC-hardness}). Therefore, no such sequence of qualified estimator functions $\EST_n$ exists when the sequence of parameters $\{(k_n, d_n, \lambda_n, \tau_n)\}_{n \in \mathbb{N}}$ is in the proposed regime.
	which completes the proof of the theorem.
\end{enumerate}

\end{proof}

\subsubsection{Proof of Proposition~\ref{prop:TS-SDP-hard}} \label{proof-prop-TS-SDP-hard}
We first state an auxilary lemma.
\begin{lemma}\label{lemma:concentration-prop-TS-SDP-hard}
Let $\bm{Z} \in \mathbb{R}^{n \times d}$ be a random matrix such that entries of $\bm{Z}$ are independent gaussian random variables with zero mean and unit variance. Let $\bm{Z}^{(i)} \in \mathbb{R}^{n}$ denotes the $i$-th column of $\bm{Z}$ for each $i \in [d]$. Further, let
\[
    \bar{\bm{Z}}^{(i)} = \frac{\sqrt{n} \bm{Z}^{(i)} }{\|\bm{Z}^{(i)}\|_{2}},\;\forall \; i \in [d] \quad \text{and} \quad \bar{\bm{Z}} = \big[ \bar{\bm{Z}}^{(1)}\;|\cdots|\;\bar{\bm{Z}}^{(d)} \big].
\]
Then the following holds with probability at least $1- c d^{- c}$ for some constant $c \geq 1$
\begin{subequations}
\begin{align}
    &\| \bar{\bm{Z}}^{(i)} \|_2 = \sqrt{n},\;\forall \; i \in [d] \label{ineq1-concentration-prop-TS-SDP-hard} \\
    &| \langle \bar{\bm{Z}}^{(i)}, \bar{\bm{Z}}^{(j)} \rangle| \lesssim \sqrt{n},\; \forall \; i \neq j \in [d] \label{ineq2-concentration-prop-TS-SDP-hard}\\
    &\big| \|\bm{Z}^{(i)}\|_{2} - \sqrt{n} \big| \lesssim \log (dn), \;\forall \; i \in [d] \label{ineq3-concentration-prop-TS-SDP-hard} \quad \text{and} \\
    & \| \bar{\bm{Z}}^{\top} \bar{\bm{Z}} \|_{F}^{2} \geq (1-o(1)) \cdot nd^{2}.\label{ineq4-concentration-prop-TS-SDP-hard}
\end{align}
\end{subequations}
\end{lemma}
The inequalities~\eqref{ineq1-concentration-prop-TS-SDP-hard},~\eqref{ineq2-concentration-prop-TS-SDP-hard} and~\eqref{ineq4-concentration-prop-TS-SDP-hard} follows directly from Theorem 7.1 in \cite{ma2015sum}. The inequality~\eqref{ineq3-concentration-prop-TS-SDP-hard} follows from Theorem 3.1.1 in \cite{vershynin2018high} and applying union bound. We are now poised to prove Proposition~\ref{prop:TS-SDP-hard}.
\begin{proof}
    Let $\bm{M}_*$ be the optimal solution of the problem~\eqref{SDP-tree-sparse-PCA}. We will prove Proposition~\ref{prop:TS-SDP-hard} by contradiction. Assume that
    \begin{align}\label{assum1-proof-prop-TS-SDP-hard}
        \big\| \bm{M}_* - \bm{v}_*\bm{v}_{*}^{\top} \big\|_2 \leq \frac{1}{5},
    \end{align}
    where $\|\cdot\|_2$ denotes the spectral norm of matrices. By the triangle inequality, we obtain that
    \[
        \big\| \bm{M}_* \big\|_2 \geq \big \|\bm{v}_* \bm{v}_{*}^{\top} \big \|_2 - \big\| \bm{M}_* - \bm{v}_*\bm{v}_{*}^{\top} \big\|_2 \geq 1 - \frac{1}{5} = \frac{4}{5}.
    \]
    Let $\widehat{\bm{v}}$ be the eigenvector of $\bm{M}_*$ corresponding to the largest eigenvalue and $\|\widehat{\bm{v}}\|_{2}=1$. Then $\bm{M}_* - \frac{4}{5}\widehat{\bm{v}}\widehat{\bm{v}}^{\top}$ is positive semidefinite. We use the decomposition
    \[
        \bm{M}_* - \bm{v}_*\bm{v}_{*}^{\top} = \underbrace{\bm{M}_* - \frac{4}{5}\widehat{\bm{v}}\widehat{\bm{v}}^{\top} }_{\bm{M}_1} + \underbrace{\frac{4}{5} \big( \widehat{\bm{v}}\widehat{\bm{v}}^{\top} - \bm{v}_*\bm{v}_{*}^{\top}\big)}_{\bm{M}_2} - \frac{1}{5}\bm{v}_*\bm{v}_{*}^{\top}.
    \]
    Applying the triangle inequality, we obtain that
    \begin{align*}
        \big \| \bm{M}_2 \big\|_{2} &\leq \big\| \bm{M}_* - \bm{v}_*\bm{v}_{*}^{\top} \big\|_{2} + \big\| \bm{M}_1\big\|_{2} + \big \| \frac{1}{5}\bm{v}_*\bm{v}_{*}^{\top} \big\|_{2}
        \\ &\leq \frac{1}{5} + \mathsf{Tr}\big( \bm{M}_1 \big) + \frac{1}{5} \leq \frac{3}{5},
    \end{align*}
    where the last step follows since $\mathsf{Tr}\big( \bm{M}_1 \big) = \mathsf{Tr}\big( \bm{M}_* \big) - \frac{4}{5}\mathsf{Tr}\big(\widehat{\bm{v}}\widehat{\bm{v}}^{\top}\big) = 1/5$. Consequently, we obtain that
    \[
        \big \| \widehat{\bm{v}}\widehat{\bm{v}}^{\top} - \bm{v}_*\bm{v}_{*}^{\top} \big\|_{2} \leq \frac{3}{4}.
    \]
    It immediately follows that
    \begin{align}\label{lower-bound-alpha-proof-TSSDP}
        (\langle \widehat{\bm{v}}, \bm{v}_*\rangle)^{2} = & ~ 1 - \frac{1}{2}\cdot \big\| \widehat{\bm{v}}\widehat{\bm{v}}^{\top} - \bm{v}_*\bm{v}_{*}^{\top} \big\|_{F}^{2} \\ 
        \geq & ~ 1 - \frac{1}{2}\cdot 2\big\| \widehat{\bm{v}}\widehat{\bm{v}}^{\top} - \bm{v}_*\bm{v}_{*}^{\top} \big\|_{2}^{2} \geq \frac{7}{16}. \notag
    \end{align}
    Continuing, we use the decomposition
    \[
        \widehat{\bm{v}} = \alpha \bm{v}_* + \beta\bm{v}_{\perp},\quad \text{where }  \langle \bm{v}_*,\bm{v}_{\perp} \rangle = 0 \text{ and } \|\bm{v}_*\|_2 = \|\bm{v}_{\perp}\|_2 =1.
    \]
    Note that it follows from inequality~\eqref{lower-bound-alpha-proof-TSSDP} that $\alpha^{2} \geq 7/16$ and $\beta^{2}\leq 9/16$.
    We next claim that if $1\leq \lambda \leq d/Cn$ for a large enough constant $C$, then with probability at least $1- c_1 \exp(- c_1 n)$ for some positive constant $c_1$,
    \begin{align}\label{claim1-proof-prop-TS-SDP-hard}
            \bm{v}_{*}^{\top} \widehat{\bm{\Sigma}} \bm{v}_* \leq \frac{0.01d}{n} \quad \text{and} \quad \big\|  \widehat{\bm{\Sigma}} \big\|_{2}  \leq \frac{1.01d}{n}.
    \end{align}
    We defer the proof of Claim~\eqref{claim1-proof-prop-TS-SDP-hard} to the end. Using Claim~\eqref{claim1-proof-prop-TS-SDP-hard}, we obtain that
    \begin{align*}
         \| \widehat{\bm{v}} \|_{\widehat{\bm{\Sigma}}} \leq & ~ |\alpha| \cdot  \| \bm{v}_* \|_{\widehat{\bm{\Sigma}}} + |\beta| \cdot  \| \bm{v}_{\perp} \|_{\widehat{\bm{\Sigma}}} \\
         \leq & ~ 0.1\sqrt{d/n} + \frac{3}{4} \cdot \big\|  \widehat{\bm{\Sigma}} \big\|_{2}^{1/2} \leq (0.1 + 3\sqrt{1.01}/4) \cdot \sqrt{d/n}.
    \end{align*}
    Putting the pieces together yields 
    \begin{align}\label{upper-bound-optimal-SDP-TS}
    \begin{split}
            \mathsf{SDP}(\widehat{\bm{\Sigma}}) =  \big \langle \bm{M}_*, \widehat{\bm{\Sigma}}\big\rangle 
            = & ~ \big \langle \bm{M}_1, \widehat{\bm{\Sigma}}\big\rangle + \big \langle \frac{4}{5}\widehat{\bm{v}}\widehat{\bm{v}}^{\top} , \widehat{\bm{\Sigma}}\big\rangle \\ 
            \overset{\1}{\leq} & ~ \mathsf{Tr}(\bm{M}_1) \cdot \| \widehat{\bm{\Sigma}} \|_{2} + \frac{4}{5} \cdot \|\widehat{\bm{v}}\|_{\widehat{\bm{\Sigma}}}^{2} \\
            \leq & ~ \frac{1}{5} \cdot \frac{1.01d}{n} + \frac{4}{5} \cdot (0.1 + 3\sqrt{1.01}/4)^{2} \cdot \frac{d}{n} \leq 0.79 \cdot \frac{d}{n},
    \end{split}
    \end{align}
    where step $\1$ follows since $\bm{M}_1$ is positive semidefinite. We claim that with probability at least $1- c_2 (dn)^{-c_2}$ for some positive constant $c_2 \geq 1$, the optimal value $\mathsf{SDP}(\widehat{\bm{\Sigma}})$ in the optimization problem~\eqref{SDP-tree-sparse-PCA} satisfies
    \begin{align}\label{claim2-proof-prop-TS-SDP-hard}
        \mathsf{SDP}(\widehat{\bm{\Sigma}}) \geq \frac{0.99d}{n}.
    \end{align}
    We defer the proof of Claim~\eqref{claim2-proof-prop-TS-SDP-hard} to the end. Note that claim~\eqref{claim2-proof-prop-TS-SDP-hard} contradicts with the inequality~\eqref{upper-bound-optimal-SDP-TS}. Consequently, the assumption~\eqref{assum1-proof-prop-TS-SDP-hard} is wrong and then we conclude that
    \[
        \big\| \bm{M}_* - \bm{v}_*\bm{v}_{*}^{\top} \big\|_2 \geq \frac{1}{5}.
    \]
    \paragraph{Proof of Claim~\eqref{claim1-proof-prop-TS-SDP-hard}:} Recall that
    \[
        \widehat{\bm{\Sigma}} = \frac{1}{n}\sum_{i=1}^{n} \bm{x}_{i}\bm{x}_{i}^{\top}, \quad \text{where } \{\bm{x}_{i}\}_{i=1}^{n} \overset{\mathsf{i.i.d.}}{\sim} \mathcal{N}(\bm{0}, \bm{\Sigma}). 
    \]
    Let $\bm{z}_{i} = \bm{\Sigma}^{-1/2} \bm{x}_{i}$ for each $i \in [d]$. Some straightforward calculation yields
    \begin{align*}
        \bm{v}_{*}^{\top} \widehat{\bm{\Sigma}} \bm{v}_* = & ~ \frac{1}{n} \sum_{i=1}^{n} (\bm{z}_{i}^{\top} \bm{\Sigma}^{1/2} \bm{v}_{*})^{2} \\
        = & ~ \frac{\|\bm{\Sigma}^{1/2} \bm{v}_{*}\|_{2}^{2}}{n} \cdot \sum_{i=1}^{n} \bigg( \frac{\bm{z}_{i}^{\top} \bm{\Sigma}^{1/2} \bm{v}_{*}}{\|\bm{\Sigma}^{1/2} \bm{v}_{*}\|_{2}} \bigg)^{2} \\
        = & ~ \frac{\lambda+1}{n} \cdot \sum_{i=1}^{n} \bigg( \frac{\bm{z}_{i}^{\top} \bm{\Sigma}^{1/2} \bm{v}_{*}}{\|\bm{\Sigma}^{1/2} \bm{v}_{*}\|_{2}} \bigg)^{2}.
    \end{align*}
    Let $\bm{G} \sim \mathcal{N}(\bm{0},\bm{I}_{n\times n})$ be independent of data $\bm{X}$. Note that $\sum_{i=1}^{n} \bigg( \frac{\bm{z}_{i}^{\top} \bm{\Sigma}^{1/2} \bm{v}_{*}}{\|\bm{\Sigma}^{1/2} \bm{v}_{*}\|_{2}} \bigg)^{2}$ and $\|\bm{G}\|_{2}^{2}$ are identically and independently distributed. Using the concentration of $\|\bm{G}\|_{2}^{2}$, we obtain that with probability at least $1- c_1 \exp(- c_1 n)$ for some positive constant $c_1$,
    \[
        \sum_{i=1}^{n} \bigg( \frac{\bm{z}_{i}^{\top} \bm{\Sigma}^{1/2} \bm{v}_{*}}{\|\bm{\Sigma}^{1/2} \bm{v}_{*}\|_{2}} \bigg)^{2} \leq 2n.
    \]
    Consequently, for $1 \leq \lambda \leq d/(Cn)$, we obtain that with probability at least $1- c_1 \exp(- c_1 n)$ for some positive constant $c_1$,
    \[
        \bm{v}_{*}^{\top} \widehat{\bm{\Sigma}} \bm{v}_* \leq 2(\lambda+1) \leq 4\lambda \leq \frac{4d}{Cn} \leq \frac{0.01d}{n}.
    \]
    It follows from Proposition 5.3 of \cite{krauthgamer2015semidefinite} that if $1 \leq \lambda \leq d/(Cn)$ for a large enough $C$ then
    \[
        \| \widehat{\bm{\Sigma}} \|_{2} \leq \frac{1.01d}{n}.
    \]

    \paragraph{Proof of Claim~\eqref{claim2-proof-prop-TS-SDP-hard}:}
    Let $S = \mathsf{supp}(\bm{v}_*)$ be the support set of the ground truth $\bm{v}_*$ and let $T = [d] \setminus S$ be the complement of $S$. Without loss of generality, we assume $S = [k]$. Let $\bm{X}^{(i)}$ denotes the $i$-th column of the data matrix $\bm{X}$ for $i\in [d]$. Note that $\{ \bm{X}^{(i)}\}_{i \in T} \overset{i.i.d.}{\sim} \mathcal{N}(\bm{0},\bm{I}_{n\times n})$. Let
    \[
         \bar{\bm{X}}^{(i)} = \frac{\sqrt{n} \bm{X}^{(i)} }{\|\bm{X}^{(i)}\|_{2}},\;\forall \; i \in [d] \quad \text{and} \quad \bar{\bm{X}} = \big[ \bar{\bm{X}}^{(1)}\;|\cdots|\;\bar{\bm{X}}^{(d)} \big].
    \]
    Moreover, we use $\bm{X}_{T}$ and $\bar{\bm{X}}_{T}$ to denote the submatrices of $\bm{X}$ and $\bar{\bm{X}}$ by restricting to the columns indexed by $T$. We construct $\bm{M} \in \mathbb{R}^{d}$ such that
    \begin{align*}
        \bm{M}_{ij} = \left\{ \begin{array}{llll} \langle \bar{\bm{X}}^{(i)}, \bar{\bm{X}}^{(i)} \rangle/(dn)&,\;\text{if}\; i=j\in [d]\\ 
        \langle \bar{\bm{X}}^{(i)}, \bar{\bm{X}}^{(j)} \rangle/(dn)&,\;\text{if}\; i\neq j \in T \\
        0&,\;\text{else}  \end{array} \right. .
    \end{align*}
    We first verify that $\bm{M}$ is feasible for the optimization problem~\eqref{SDP-tree-sparse-PCA}. Note that
    \begin{align*}
        \sum_{i=1}^{d} \bm{M}_{ii} = & ~ \sum_{i=1}^{d} \langle \bar{\bm{X}}^{(i)}, \bar{\bm{X}}^{(i)} \rangle/(dn) = \sum_{i=1}^{d} \| \bar{\bm{X}}^{(i)} \|_{2}^{2}/(dn) = \sum_{i=1}^{d} n/(dn) =1, \\
        \bm{M}_{ii} = & ~ \bm{M}_{\lfloor \frac{i}{2} \rfloor \lfloor \frac{i}{2} \rfloor},\; \forall\; i \in [d] \quad \text{ and} \\
        \sum_{i,j\in [d]} | \bm{M}_{ij} | = & ~ \sum_{i=1}^{d} |\bm{M}_{ii}| + \sum_{i \neq j\in T} | \bm{M}_{ij} | \\
        = & ~ 1 + \sum_{i \neq j\in T} | \langle \bar{\bm{X}}^{(i)}, \bar{\bm{X}}^{(j)} \rangle|/(dn) \\
        \overset{\1}{\leq} & ~ 1 + Cd^{2} \cdot \sqrt{n} /(dn) = 1 + Cd/\sqrt{n} \leq k,\\
    \end{align*}
    where in step $\1$ we use $\langle \bar{\bm{X}}^{(i)}, \bar{\bm{X}}^{(j)} \rangle \lesssim \sqrt{n}$ by Lemma~\ref{lemma:concentration-prop-TS-SDP-hard} and in the last step we use $d \asymp n$ and $n \leq ck^{2}$. Moreover, $\bm{M}$ is positive semidefinite since
    \[
        \bm{M} = \left[\begin{array}{cc} \frac{1}{dn}\bm{I}_{k \times k} & \bm{O} \\ \bm{0} & \frac{1}{dn} \bar{\bm{X}}_{T}^{\top}\bar{\bm{X}}_{T}  \end{array}\right].
    \]
    Consequently, $\bm{M}$ is a feasible solution of problem~\eqref{SDP-tree-sparse-PCA}. We next turn to obtain a lower bound of the objective value of $\bm{M}$. Note that
    \begin{align*}
        \sum_{i=1}^{d} \sum_{j=1}^{d} \widehat{\bm{\Sigma}}_{ij} \bm{M}_{ij} &= \sum_{i\in S} \widehat{\bm{\Sigma}}_{ii} \bm{M}_{ii} + \sum_{i,j\in T} \widehat{\bm{\Sigma}}_{ij} \bm{M}_{ij} \\
        & = \sum_{i\in S} \widehat{\bm{\Sigma}}_{ii} + \sum_{i,j\in T} \frac{\langle \bm{X}^{(i)}, \bm{X}^{(j)}\rangle}{n} \cdot \frac{\langle \bar{\bm{X}}^{(i)}, \bar{\bm{X}}^{(j)} \rangle}{dn} \\
        & \geq \sum_{i,j\in T} \frac{ \| \bm{X}^{(i)}\|_2 \| \bm{X}^{(j)} \|_2}{n^2} \cdot \frac{ \big( \langle \bar{\bm{X}}^{(i)}, \bar{\bm{X}}^{(j)} \rangle \big)^{2}}{dn} \\
        & \overset{\1}{\geq} \frac{(\sqrt{n} - C\log (dn))^{2}}{dn^{3}} \cdot \sum_{i,j\in T} \big( \langle \bar{\bm{X}}^{(i)}, \bar{\bm{X}}^{(j)} \rangle \big)^{2} \\
        & = \frac{(\sqrt{n} - C\log n)^{2}}{dn^{3}} \cdot \| \bar{\bm{X}}_{T}^{\top} \bar{\bm{X}}_{T} \|_{F}^{2} \\
        & \overset{\2}{\geq} \frac{(\sqrt{n} - C\log (dn))^{2}}{dn^{3}} \cdot (1-o(1)) \cdot  n(d-k)^{2} \\
        & = \frac{d}{n} \cdot \big(1-\frac{C\log (dn)}{\sqrt{n}} \big) \cdot (1-o(1)) \cdot (1-\frac{k}{d})^{2},
    \end{align*}
    where in steps $\1$ and $\2$ we use the result of Lemma~\ref{lemma:concentration-prop-TS-SDP-hard} that
	\begin{align*}
		& ~ \|\bm{X}^{(i)}\|_2 \geq \sqrt{n} - C\log n, \; \forall \; i \in T \\
        \quad \text{and} & ~ \| \bar{\bm{X}}_{T}^{\top} \bar{\bm{X}}_{T} \|_{F}^{2} \geq (1-o(1))n(d-k)^{2}
	\end{align*}
    holds with probability at least $1- c_2 (dn)^{-c_2}$ for some positive constant $c_2 \geq 1$. 
    Consequently,
	\begin{align*}
		\mathsf{SDP}(\widehat{\bm{\Sigma}}) \geq & ~ \sum_{i=1}^{d} \sum_{j=1}^{d} \widehat{\bm{\Sigma}}_{ij} \bm{M}_{ij} \\
		\geq & ~ \frac{d}{n} \cdot \big(1-\frac{C\log n}{\sqrt{n}} \big) \cdot (1-o(1)) \cdot (1-\frac{k}{d})^{2} \\
		\geq & ~ \frac{0.99d}{n}.
	\end{align*}
\end{proof}